\documentclass[twoside,11pt]{article}

\usepackage{jmlr2e}
\usepackage{soul}
\usepackage{graphicx}
\usepackage{amsmath}
\usepackage{booktabs}
\usepackage{algorithm}
\usepackage{algorithmic}

\usepackage{amsfonts}
\usepackage{subfigure}
\usepackage{verbatim}
\newtheorem{assumption}{Assumption}

\newlength\myindent
\jmlrheading{~}{~}{~}{~}{~}{~}{~}

\ShortHeadings{~}{~}
\firstpageno{1}
\date{\today}

\begin{document}

\title{Revisiting SGD with Increasingly Weighted Averaging:\\ Optimization and Generalization Perspectives}

\author{\name Zhishuai Guo \email zhishuai-guo@uiowa.edu\\
        \addr The University of Iowa \\
        \name Yan Yan \email yanyan.tju@gmail.com\\
        \addr The University of Iowa \\
        \name Tianbao Yang \email tianbao-yang@uiowa.edu\\
        \addr  The University of Iowa \\
        }

\maketitle
\begin{abstract}
Stochastic gradient descent (SGD) has been widely studied in the literature from different angles, and is commonly employed for solving many big data machine learning problems. 
However, the averaging technique, which combines all iterative solutions into a single solution, is still under-explored.
While some increasingly weighted averaging schemes have  been considered in the literature, existing works are mostly restricted to strongly convex objective functions and the convergence of optimization error.
It remains unclear how these averaging schemes affect the convergence of {\it both optimization error and  generalization error} (two equally important components of testing error) for {\bf non-strongly convex objectives, including non-convex problems}.
In this paper, we {\it fill the gap} by comprehensively analyzing the increasingly weighted averaging on convex, strongly convex and non-convex objective functions in terms of both optimization error and generalization error.
In particular, we analyze a family of increasingly weighted averaging, where the weight for the solution at iteration $t$ is proportional to $t^{\alpha}$ ($\alpha > 0$).
We show how $\alpha$ affects the optimization error and the generalization error, and exhibit the trade-off caused by $\alpha$.
Experiments have demonstrated this trade-off and the effectiveness of polynomially increased weighted averaging compared with other averaging schemes  for a wide range of problems including deep learning. 
\end{abstract}

\section{Introduction}
In machine learning,  the task of learning a predictive model is usually formulated as the following empirical risk minimization (ERM) problem:
\begin{equation}\label{eqn:erm}
\begin{split}
x_* = \arg\min\limits_{x \in \Omega} F_{\mathcal{S}}(x)  = \frac{1}{n} \sum\limits_{i=1}^{n} f(x; z_i),
\end{split}
\end{equation}
where $f(x; z)$ is a loss function of the model $x$ on a data $z$, $\Omega$ is a closed convex set, and $\mathcal{S} = \{z_1, ..., z_n\}$ denotes a set of $n$ observed data points that are sampled from an underlying unknown distribution $\mathbb{P}_z$ with support on $\mathcal{Z}$.
For a function $f$, $\nabla f(x)$ denotes a subgradient of $f$ at $x$.

To solve the ERM problem, stochastic gradient descent (SGD) is usually employed, which updates the solution according to
\begin{equation}
x_{t + 1} = \Pi_{\Omega}[x_t - \eta_t \nabla f(x_t; z_{i_t})],
\end{equation}
for $t = 1, ..., T$, where $i_t \in \{1, ..., n\}$ is randomly sampled, $\eta_t$ is the step size, and $\Pi_{\Omega} [x]$ is a projection operator, i.e.,  $\Pi_{\Omega} [x] = \arg \min_{v \in \Omega} \|x - v\|^2$, where $\|\cdot\|$ denotes the Euclidean norm.
After all the $T$ iterations, a solution is output as the prediction model. 

SGD has been widely studied in the literature from different angles, and is commonly employed for solving many big data machine learning problems. However, the averaging technique that combines all iterative solutions into a single solution is still under-explored. Most studies simply output the last solution or adopt the uniform averaging that computes a uniformly averaged solution based on all iterative solutions, i.e. $(x_1 + \cdots + x_T)/T$. 
Nonetheless, both approaches could lead to unsatisfactory performance regardless of their benefits.  The last solution could have a faster convergence in terms of optimization error in practice, but is less stable~\cite{bottou2010large,hardt2015train}. Uniform averaging can improve the stability of the output solution, but may slow down the convergence of optimization error in some cases~\cite{hazan2007logarithmic,hazan2014beyond}.  Although some non-uniform averaging has been considered in the literature~\cite{rakhlin2011making,shamir2013stochastic,lacoste2012simpler}, most of their analysis is focused on the effect on the convergence of optimization error. Their impact on the tradeoff between optimization error and the generalization error is unclear. 

In order to understand this tradeoff, we need to consider the following risk minimization that is of utmost  interest: 
\begin{align}
\min_{x\in\Omega} F(x) = E_{z\sim \mathbb P}[f(x; z)].
\end{align}
Regarding an iterative algorithm $\mathcal A$  that outputs solution $x_S$  by solving~(\ref{eqn:erm}), the central concern is its generalization performance, which can be measured by $E_{\mathcal {A,S}}[F(x_S)]$, where the expectation is taken over all randomness in the algorithm $\mathcal A$ and the training dataset $\mathcal S$.  To analyze the generalization performance (referred to as the testing error hereafter) of a random solution $x_{\mathcal S}$, we use the following decomposition of testing error:
\begin{equation*}
\begin{split}
E_{\mathcal{A,S}} [F(x_{\mathcal S})] 
=
E_{\mathcal{S}}[F_{\mathcal{S}}(x_*)] + E_{\mathcal{S}} \underbrace{E_{\mathcal{A}}[F_{\mathcal{S}}(x_{\mathcal S}) - F_{\mathcal{S}}(x_*)]}_{\varepsilon_{\text{opt}}} 
+ \underbrace{E_{\mathcal{A,S}} [F(x_{\mathcal S}) - F_{\mathcal{S}}(x_{\mathcal S})]}_{\varepsilon_{\text{gen}}},
\end{split}
\end{equation*}
where $\varepsilon_{\text{opt}}$ denotes the optimization error of the algorithm $\mathcal A$ and $\varepsilon_{\text{gen}}$ denotes the generalization error of the solution $x_{\mathcal S}$.


Most existing analysis about the non-uniform averaging is restricted to the convergence analysis of  optimization error for strongly convex objectives $F_{\mathcal S}$.  
 This paper aims to fill the gap by comprehensively analyzing a polynomially increased weighted averaging (PIWA) scheme where the weight of the solution of iteration $t$ is proportional to $t^{\alpha}$ ($\alpha>0$). 
We analyze SGD with PIWA for general convex, strongly convex and non-convex objective functions in terms of both optimization error and generalization error.
We prove that SGD with PIWA has the optimal convergence rate in term of optimization error in both general convex and strongly convex cases, i.e., $O(1/\sqrt{T})$ for the convex case, and $O(1/T)$ for the strongly convex case.
For non-convex case, we employ the PIWA  in a stagewise algorithm, which uses SGD with the averaging for solving  a convex subproblem at each stage. We establish  a convergence rate of $O(1/T)$ in terms of the optimization error for a family of weakly-convex functions that satisfies the Polyak-\L ojasiewicz condition. 
Moreover, we analyze the generalization error of SGD with  PIWA following the analysis framework in \cite{hardt2015train} that uses the uniform stability tool. 
We show that SGD with PIWA may have smaller generalization error than SGD using the last solution.
We also show how $\alpha$ affects optimization error and generalization error, and thus exhibits their trade-off caused by $\alpha$. We have also conducted extensive experiments on convex, strongly convex and non-convex functions.
The experimental results demonstrate  the trade-off caused by $\alpha$ and the effectiveness of PIWA compared with other commonly used averaging schemes.

\section{Related work}
There are a lot of works that analyze SGD with uniform averaging \cite{polyak1990new,polyak1992acceleration,zinkevich2003online,bottou2010large,chen2018universal}.
Uniform averaging can help to improve the stability in terms of  generalization \cite{hardt2015train}.
It can also help to get an optimal convergence rate in the convex case \cite{polyak1992acceleration,zinkevich2003online}.
However, as it gives equal weight to every solution, it could actually slow down the convergence in many cases as later solutions are usually more accurate than earlier solutions~\cite{chen2018universal,shamir2013stochastic}.
For example, in strongly convex case, 
the uniformly averaged solution has an $O(\log T/T)$ convergence rate \cite{hazan2007logarithmic,hazan2014beyond,shamir2013stochastic}, which is suboptimal. 

In order to improve the convergence rate for optimization of strongly convex functions, many non-uniform averaging schemes have been proposed~\cite{rakhlin2011making,shamir2013stochastic,lacoste2012simpler}. 
\cite{rakhlin2011making} considered suffix averaging that takes average of the last $\alpha T$ solutions (where $\alpha \in (0, 1)$). However, suffix averaging cannot be updated online and thus are computationally expensive. \cite{shamir2013stochastic} proposed polynomial-decay averaging for minimizing strongly convex functions. 
\cite{lacoste2012simpler} consider a simple polynomially increased weighted averaging, where the weight for the solution of the $t$-th iteration is proportional to $O(t)$. SGD with these averaging schemes  have been shown to achieve the optimal convergence rate of $O(1/T)$ for minimizing a strongly convex objective.

However, these existing works restrict their attention for minimizing a strongly convex objective and only analyze the optimization error.
Thus, the theory for these non-uniform averaging schemes in convex and non-convex cases are lacked and their impact on the generalization error is unclear.
It should be mentioned that exponential moving averaging technique has also been widely used ~\cite{kingma2014adam,zhang2015deep}, which maintains a moving averaging  by $\bar{x}_{t+1} = \alpha \bar{x}_t + (1 - \alpha)x_t$ with $\alpha\in(0,1)$.
However, as the weights for previous solution decay exponentially, its performance is close to last solution.
What is more, we do not know any existing theoretical guarantee on the performance of moving averaging.

Studies on the non-convex case used to analyze a randomly sampled solution \cite{ghadimi2013stochastic,DBLP:conf/ijcai/YanYLLY18,davis2018stochastic}.
Recently, stagewise algorithms enable the use of averaging for a class of non-convex functions, namely weakly convex functions~\cite{chen2018universal,davis2017proximally}.
These stagewise algorithms construct  a convex objective function as a subproblem for each stage.
By solving these subproblems in a stagewise manner, it can guarantee the convergence for the original problem.
In \cite{chen2018universal}, they use uniform averaging, thus it may also suffer from  slow convergence in practice.
In \cite{davis2017proximally}, the weight of the solution of iteration $t$ at each stage is proportional to $t$.
Hence,  it may not achieve the best trade-off between the optimization error and the generalization error.  


In \cite{hardt2015train}, they have shown that uniform averaging can improve the generalization stability.
In \cite{yang2018does}, they have derived the bound of generalization error for stagewise algorithm with uniformly averaging.
The analysis of generalization error of SGD in this work can be considered as extensions of these works \cite{hardt2015train,yang2018does} by analyzing the impact of PIWA on the generalization error. 

\section{Preliminaries}

A function $f(x)$ is G-Lipchitz continuous if $\|\nabla f(x)\| \leq G$, i.e., $\|f(x) - f(y) \| \leq G\|x - y\|$, and  is $L$-smooth if it is differentiable and its gradient is $L$-Lipchitz continuous.

A function $f(x)$ is $\lambda$-strongly convex for $\lambda > 0$, if for all $x, y \in \Omega$,
\begin{equation*}
f(y) \geq f(x) + \langle \nabla f(x), y - x \rangle + \frac{\lambda}{2} \|x - y\|^2
\end{equation*}
A non-convex function $f(x)$ is called $\rho$-weakly convex for $\rho > 0$ if $f(x) + \frac{\rho}{2}\|x\|^2$ is convex. 



For the analysis of the generalization error, we will use the uniform stability tool \cite{bousquet2002stability}. 
The definition of uniform stability is given below.

\begin{definition}
A randomized algorithm $\mathcal{A}$ is called $\epsilon$-uniformly stable if for all datasets $\mathcal{S}, \mathcal{S'} \in \mathcal{Z}^n$ that differ at most one example, the following holds:
\begin{equation*}
\sup\limits_{z} E_{\mathcal{A}}[f(\mathcal{A}(\mathcal{S}); z) - f(\mathcal{A}(\mathcal{S}'); z)] \leq \epsilon.
\end{equation*}
where $\mathcal A(\mathcal S)$ denotes the random solution returned by algorithm $\mathcal A$ based on the dataset $\mathcal S$.
\end{definition}

The relation between uniform stability and generalization error is given in the following lemma.
\begin{lemma}(\cite{bousquet2002stability}) 
If $\mathcal{A}$ is $\epsilon$-uniformly stable, we have $\varepsilon_{\text{gen}} \leq \epsilon$.
\end{lemma}

Therefore, in order to compare the testing error of different randomized algorithms, it suffices to analyze their convergence in terms of optimization error and their uniform  stability.

\section{Main Theoretical Results}
In this section, we will analyze SGD with PIWA  in terms of both optimization and generalization error.
We denote the algorithm by SGD-PIWA and present it in \textbf{Algorithm \ref{alpha}}.
We particularly use $w_t = t^\alpha$ where $\alpha \geq 0$ to control the averaging weights of the solution at the $t$-th iteration.
It should be noticed that SGD with uniform averaging is a special case of this algorithm by taking $\alpha = 0$, and the averaging scheme proposed in \cite{lacoste2012simpler} is also a special case by setting  $\alpha = 1$. The step size $\eta_t$ at the $t$-th iteration will be different for different classes of functions, which will be exhibited later. 

\begin{algorithm}
\caption {SGD-PIWA: $(F_{\mathcal{S}}, x_1, \eta_1, T, \Omega)$}
\begin{algorithmic}
\FOR {$t = 1, ..., T-1$}
\STATE{Compute a step size $\eta_t$}
\STATE{$x_{t+1} =\Pi _{\Omega}[x_t - \eta_t \nabla f(x_t; z_{i_t})]$}
\ENDFOR
\STATE {\textbf{Output} $\bar{x}_T = \frac{1}{\sum\limits_{t = 1}^{T} w_t}\sum\limits_{t = 1}^{T} w_t x_t$}, where $w_t = t^\alpha, \alpha\geq 0$. 
\end{algorithmic}
\label{alpha}
\end{algorithm}

It is notable that the final averaged solution  can be computed {\it online} by updating an averaged sequence $\{\bar x_t\}$:
\begin{equation*}
\begin{split}
\bar{x}_T =  \left(\left(\sum\limits_{t=1}^{T-1} w_t \right) \bar{x}_{T-1} + w_Tx_T \right)\bigg/\left(\sum\limits_{t=1}^{T-1} w_t + w_T\right).
\end{split}
\end{equation*}

Before showing the trade-off of $\alpha$, we would emphasize that
most proof of convergence of optimization error in existing works for averaging schemes uses the following Jensen's inequality to first upper bound the objective gap: 
\begin{equation}\label{eq:PIWA_average}
\begin{split}
F_{\mathcal{S}}(\bar{x}_T) - F_{\mathcal{S}}(x_*) \leq
 \frac{1}{\sum\limits_{t=1}^{T}t^{\alpha}}{\sum\limits_{t=1}^{T} \Big(t^{\alpha}F_{\mathcal{S}}(x_t)}\Big) - F_{\mathcal{S}} (x_*) ,
\end{split}
\end{equation}
and then further upper bound the right hand side (RHS) of (\ref{eq:PIWA_average}). 
In this way, the objective gap, i.e., $F_{\mathcal{S}}(\bar{x}_T) - F_{\mathcal{S}}(x_*)$, is relaxed twice.

However, it may not be precise to only focus on the effect of $\alpha$ on the two-time relaxation of $F_{\mathcal{S}}(\bar{x}_T) - F_{\mathcal{S}}(x_*)$.
Instead, to investigate the benefit of PIWA on the convergence of optimization error, we propose to additionally inspect the first relaxation, which we refer to the RHS of (\ref{eq:PIWA_average}).
Specifically, we present the following lemma to illustrate how $\alpha$ affects the RHS of (\ref{eq:PIWA_average}).

\begin{lemma}\label{lem:1}
Assume $F_{\mathcal{S}}(x_T) \leq F_{\mathcal{S}}(x_{T - 1}) \leq ... \leq F_{\mathcal{S}}(x_2) \leq F_{\mathcal{S}}(x_1)$.
The function
\begin{equation}
\begin{split}
H_T(\alpha) = \frac{1}{\sum\limits_{t=1}^{T}t^{\alpha}}{\sum\limits_{t=1}^{T}\Big(t^{\alpha}F_{\mathcal{S}}(x_t)}\Big)
\end{split}
\end{equation}
is non-increasing in $\alpha$ for $\alpha \geq 0$.
\label{motivation}
\end{lemma}

We can see under the condition that the sequence of solutions yield non-increasing  objective values, a larger $\alpha$ in PIWA will make $H_T(\alpha)$ smaller, 
which indicates that the RHS of (\ref{eq:PIWA_average}) is smaller.
The assumption may not hold in practice, but, to some degree, it explains the effect of $\alpha$ on the first relaxation of the objective gap.

In the subsequent subsections, we provide convergence analysis of PIWA in optimization and generalization error in different conditions,
i.e., general convex, strongly convex and non-convex cases.
We reveal how $\alpha$ affects the upper bound of $H_T(\alpha) - F_{\mathcal S}(x_*)$ and how $\alpha$ causes tradeoff in optimization error.

\subsection{General Convex Case}
\subsubsection{Optimization Error}
In  the general convex case, we need the following assumptions.
\begin{assumption}
~\\
(i) $f(x; z)$ is a convex function in terms of $x$ for any $z \in \mathcal{Z}$; \\
(ii) $E [\|\nabla f(x; z)\|^2] \leq G^2$,  for any $x \in \Omega$;\\
(iii) there exists $D$ such that $\|x - y\| \leq D$  for any $x, y \in \Omega$.
\end{assumption}
Based on Assumption 1, we have the following theorem.
\begin{theorem}
Suppose \textbf{Assumption 1} holds and by setting $\eta_t = \frac{\eta_1}{{t}^{1/2}}$, we have
\begin{equation*}
\begin{split}
\frac{1}{\sum\limits_{t=1}^{T}t^{\alpha}}  E\left[{\sum\limits_{t=1}^{T}t^{\alpha}(F_{\mathcal{S}}(x_t)} - F_{\mathcal{S}} (x_*)) \right] 
\leq \left\{
\begin{aligned}
&\frac{(\alpha + 1)D^2}{2\eta_1{T}^{1/2}} + \frac{(\alpha + 1)\eta_1 G^2}{(2\alpha+1){T}^{1/2}}, &0 \leq \alpha <1/2   ,   \\
&\frac{(\alpha + 1)D^2}{2\eta_1 {T}^{1/2}} + \frac{(\alpha + 1)\eta_1 G^2}{{T}^{1/2}}, &\alpha \geq 1/2   .
\end{aligned}
\right.
\end{split}
\end{equation*}
\label{convex_optimization}
\end{theorem}

\noindent
\textbf{Remark.} The convergence rate for different value of $\alpha$ is in the same order of  $O(1/\sqrt{T})$, which is optimal~\cite{polyak1992acceleration,zinkevich2003online}. One may notice that a larger $\alpha$ would yield worse convergence bound for $H_T(\alpha) - F_{\mathcal{S}} (x_*)$.
It, however, does not necessarily indicate a worse optimization error in light of Lemma~\ref{lem:1}. In practice, there will be a trade-off by using different values of $\alpha$.  

\subsubsection{Generalization Error}
In this subsection, we analyze the uniform stability of SGD-PIWA.
Our analysis closely follows the route in \cite{hardt2015train}.
By showing $\sup_{z} E_{\mathcal{A}}[f(\bar{x}_T; z) - f(\bar{x}_T'; z)] \leq \epsilon$, we show the generalization error is bounded by $\epsilon$, where 
$\bar{x}_T$ is learned by SGD-PIWA on a dataset $\mathcal{S}$ and $\bar{x}_T'$ is learned  by SGD-PIWA on a dataset $\mathcal{S}'$ with $\mathcal{S}$ differing  from $\mathcal{S}'$ at most one example.

Similar to Theorem 3.8 in \cite{hardt2015train}, Lemma \ref{lemma_sequence_deviation} provides a bound of the deviation between the two sequences $x_t$ and $x_t'$ of SGD-PIWA that runs on $\mathcal S$ and $\mathcal S'$ separately. 

\begin{lemma}
Assume that the loss function $f(x; z)$ is convex, $G$-Lipchitz, $L$-smooth for every $z \in \mathcal{Z}$. 
Suppose we run SGD-PIWA with step sizes $\eta_t \leq 2/L$ for $T$ steps.
Then, 
\begin{equation}
\begin{split}
E[\|x_T - x_T'\|]\leq \frac{2G}{n} \sum\limits_{t=1}^{T-1} \eta_t.
\end{split}
\end{equation}
\label{lemma_sequence_deviation}
\end{lemma}

\begin{theorem}
Suppose Assumption 1 holds, and we further assume $f(x; z)$ is a $G$-Lipschitz and $L$-smooth function.
Set $\eta_1 \leq 2/L$.
Then the {Algorithm \ref{alpha}} has the uniform stability of
\begin{equation}
\begin{split}
&\epsilon_{stab}
 \leq \frac{4\eta_1 G^2 (\alpha + 1) {(T + 1)}^{\alpha + 1.5} }{n (\alpha + 1.5){T}^{\alpha + 1}}
\end{split}
\end{equation}
\label{convex_generalization}
\end{theorem}

\noindent
\textbf{Remark.} 
When $T$ is large, $(T+1)^{\alpha+1.5}/T^{\alpha + 1} \approx T^{0.5}$.
Then we can see that the bound of generalization error is increasing in $\alpha$.
Taking $\alpha = 0$, which reduces the PIWA to uniform averaging, has the smallest generalization error.
Even if $\alpha > 0$, generalization error of PIWA is still smaller than that of last solution \cite{hardt2015train},
since $\frac{\alpha + 1}{\alpha + 1.5}$ is bounded by $1$.
It is therefore smaller than the last solution even when $\alpha$ is very large.

\subsection{Strongly Convex Case}
In this subsection, we are going to analyze our algorithm for the strongly convex objective function.
In this case, we need the following assumptions:

\begin{assumption}
~\\
(i) $f(x; z)$ is a $\lambda$-strongly convex function; \\
(ii) $E[\|\nabla f(x; z)\|^2] \leq G^2$,  for any $x \in \Omega$.\\
\end{assumption}
\subsubsection{Optimization Error}
\begin{theorem}
Suppose Assumption 2 holds,
by setting $\eta_t = \frac{2(\alpha+1)}{\lambda t}$, we have
\begin{equation*}
\begin{split}
\frac{1}{\sum\limits_{t=1}^{T}t^{\alpha}}E\left[{\sum\limits_{t=1}^{T}t^{\alpha}(F_{\mathcal{S}}(x_t)} - F_{\mathcal{S}} (x_*)) \right]
\leq \left\{
\begin{aligned}
&\frac{G^2}{\lambda T} (1 + \log (T)), &\alpha = 0\\
&\frac{{(\alpha + 1)}^2 G^2}{\alpha \lambda T}, &0 < \alpha < 1\\
&\frac{{(\alpha + 1)}^2G^2  {(T+1)}^{\alpha}}{\alpha\lambda {T}^{\alpha+1}}, &\alpha \geq 1
\end{aligned}
\right.
\end{split}
\end{equation*}
\label{strongly_optimization}
\end{theorem}

\noindent
\textbf{Remark.} When $\alpha = 0$, the algorithm degenerates to a uniform average with a $O(\log T/T)$ convergence.
By taking $\alpha>0$, the order is improved to $O(1/T)$.
The algorithm in \cite{lacoste2012simpler} is special case with $\alpha=1$, while we have generated it to any $\alpha>0$.
Note that when $\alpha$ is close to 0, although the order is $O(1/T)$, the convergence is not significantly better compared with $\alpha=0$ because of the $\alpha$ on the denominator.

\subsubsection{Generalization Error}


Then we can get the bound of the generalization error in the following theorem.
\begin{theorem}
Suppose {Assumption 2} holds, and we further assume that $f(x; z) \in [0, 1]$, 
and $f(x; z)$ is $L$-smooth.
Then by taking $\eta_t = \frac{2(\alpha + 1)}{\lambda t}$ and $t_0 = \max\{ \frac{2(\alpha+1)G}{\lambda}, 1\}$, the Algorithm 1 has uniform stability of 
\begin{equation}
\epsilon_{stab}
 \leq \frac{t_0}{n}  + \frac{4G^2(\alpha+1)}{\lambda n}
\end{equation}
\end{theorem}

\noindent
\textbf{Remark.} This result is similar as the Theorem 3.10 in \cite{hardt2015train}.
Differently, this result depends on $\alpha$.
The bound of generalization error is polynomial to $\alpha + 1$.
Again, we see that the generalization error of PIWA is worse than uniform averaging.
Thus, we cannot take $\alpha$ to be very large in experiments.

\subsection{Non-Convex Case}
In this section, we are going to analyze the optimization and generalization error of SGD-PIWA for the non-convex case.
We need the following assumptions:
\begin{assumption}
~\\
(i) For any $x\in \Omega$ and $z \in \mathcal{Z}, E[\|\nabla f(x; z)\|^2] \leq G^2$. \\
(ii) $f(x; z)$ is $\rho$-weakly convex for any $z \in \mathcal{Z}$;  \\
(iii) $f(x; z)$ is $L$-smooth;\\
(iv) For an initial solution $x_1$, there exists $\epsilon_0 > 0$ such that $F_{\mathcal{S}}(x_1) - F_{\mathcal{S}}(x_*) \leq \epsilon_0$;\\
(v) $F_{\mathcal{S}}(x)$ satisfies the PL condition, i.e., there exists $\mu>0$
\begin{equation}
\begin{split}
2\mu(F_{\mathcal{S}}(x) - F_{\mathcal{S}}(x_*)) \leq \|\nabla F_{\mathcal{S}}(x)\|^2, \forall x \in \Omega.
\end{split}
\end{equation}
\end{assumption}
These conditions have been used in~\cite{karimi2016linear,lei2017non,bassily2018exponential,reddi2016stochastic}.  The algorithm we use for the non-convex objective function is shown in \textbf{Algorithm \ref{non-convex}}, where $ \mathcal{B}(x_{k-1}, D_k)) = \{x\in\Omega: \|x - x_{k-1}\|\leq D_k\}$. 
The algorithm uses  a stagewise training strategy.
The objective function for each stage is a sum of the loss function and an $L_2$ regularization term referring to the output of previous stage.
At each stage, the objective function is optimized by SGD-PIWA. 
The step size within a stage is set to be a constant, and is decreased after a stage.

\begin{algorithm}
\caption {Stagewise Algotithm with SGD-PIWA}
\begin{algorithmic}[H]
\STATE{\textbf{Initialize:} a sequence of decreasing step size parameters 
$\{\eta_k\}, x_1 \in \Omega, \gamma \leq {\rho}^{-1}$}.
\FOR{k = 1, ..., K}
     \STATE{Let $F_k(\cdot) = F_{\mathcal{S}}(\cdot) + \frac{1}{2\gamma}\|\cdot - x_{k - 1}\|^2$}
     \STATE{$x_k = \text{SGD-PIWA}(F_k, x_{k-1}, \eta_k, T_k,  \mathcal{B}(x_{k-1}, D_k))$}
     \STATE{$D_{k+1} = D_k / 2$}
\ENDFOR
\STATE{\textbf{Output:} $x_{K}$.}

\end{algorithmic}
\label{non-convex}
\end{algorithm}

\subsubsection{Optimization Error}
We need the following lemmas.

\begin{lemma}
If $F_{\mathcal{S}}(x)$ satisfies the PL condition, then for any $x \in \Omega$ we have
\begin{equation}
\begin{split}
\|x - x_*\|^2 \leq \frac{1}{2\mu} (F_{\mathcal{S}}(x) - F_{\mathcal{S}}(x_*)).
\end{split}
\end{equation}
\label{pl_imply}
\end{lemma}

\noindent
\textbf{Remark.} 
Please refer to \cite{bolte2017error,karimi2016linear} for a proof.

We first have the following lemma for each stage of the algorithm. To this end, we let $x^k_t$ denote the $t$-th solution at the $k$-th stage. 

\begin{lemma}
Suppose Assumption 3 holds.
By applying SGD to $F_k (x) = F_{\mathcal{S}}(x) + \frac{1}{2\gamma} \|x - x_{k-1}\|^2$ with $\gamma \leq 1/\rho$, $\eta \leq 1/L$.
And we further assume that $F_{\mathcal{S}}(x_{k-1}) - F_{\mathcal{S}}(x_*) \leq \epsilon_{k-1}$.
Moreover the solution of each iteration in $k$-th stage is projected to the Euclidean ball $B(x_{k-1}, D_k)$, where $D_k = \sqrt{\frac{\epsilon_{k-1}}{\mu}}$.
Then with a probability at least $1 - \delta$, we have
\begin{equation*}
\begin{split}
&F_{k}(x_k) - F_{k}(x_*)\leq \frac{1}{\sum\limits_{t=1}^{T_k}t^{\alpha}}E\left[{\sum\limits_{t=1}^{T_k}t^{\alpha}(F_{k}(x^k_t)} - F_{k} (x_*)) \right]\\
&\leq\frac{2(\alpha + 1)\epsilon_{k-1}}{\eta_k \mu T} + \frac{\eta_k \hat{G}^2}{2}  + \frac{2(\alpha + 1) \hat{G}D_k \sqrt{2 \ln (1/\delta)}}{\sqrt{T}},
\end{split}
\end{equation*}
where $x_t^k$ is the solution of $t$-thm iteration at stage $k$, $\hat{G}^2$ is the upper bound of $E[\|\nabla f(x; z) + \gamma^{-1}(x - x_{k-1})\|^2]$, which exists and can be set to $2G^2 + 2\gamma^{-2}D_k^2$.
\label{each_stage}
\end{lemma}

\begin{figure*}[htb]
\centering
\includegraphics[scale=0.2]{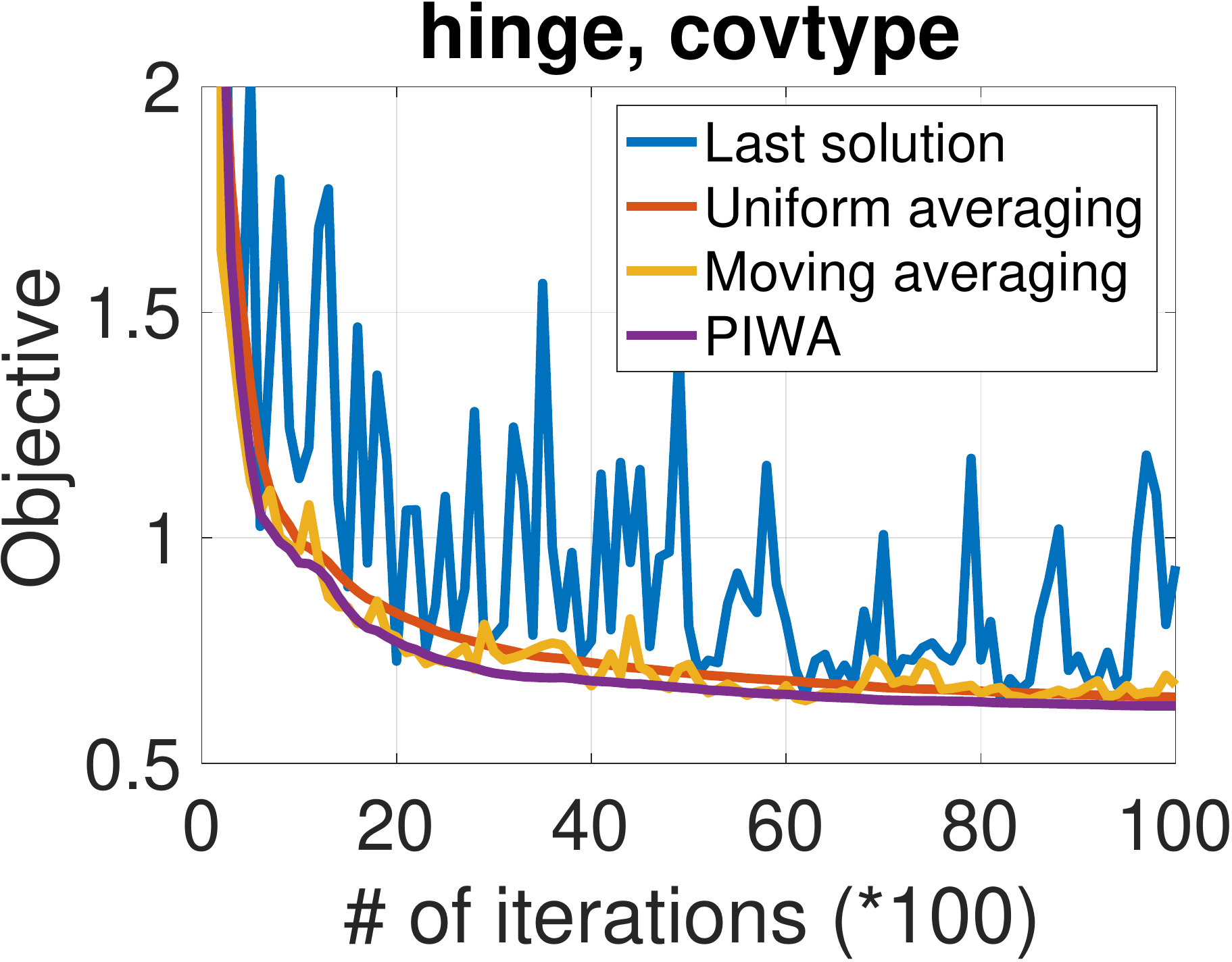}
\includegraphics[scale=0.2]{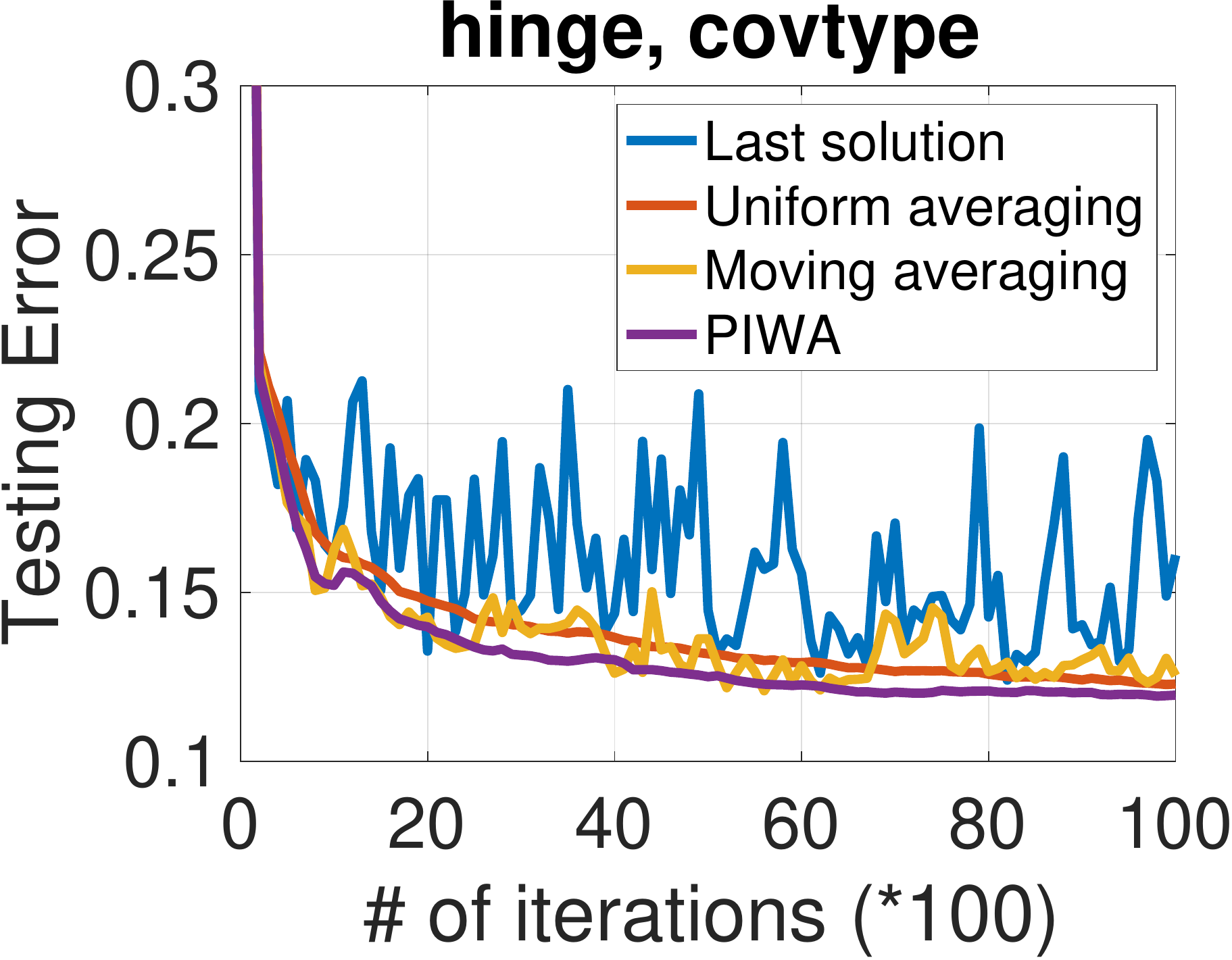}
\includegraphics[scale=0.2]{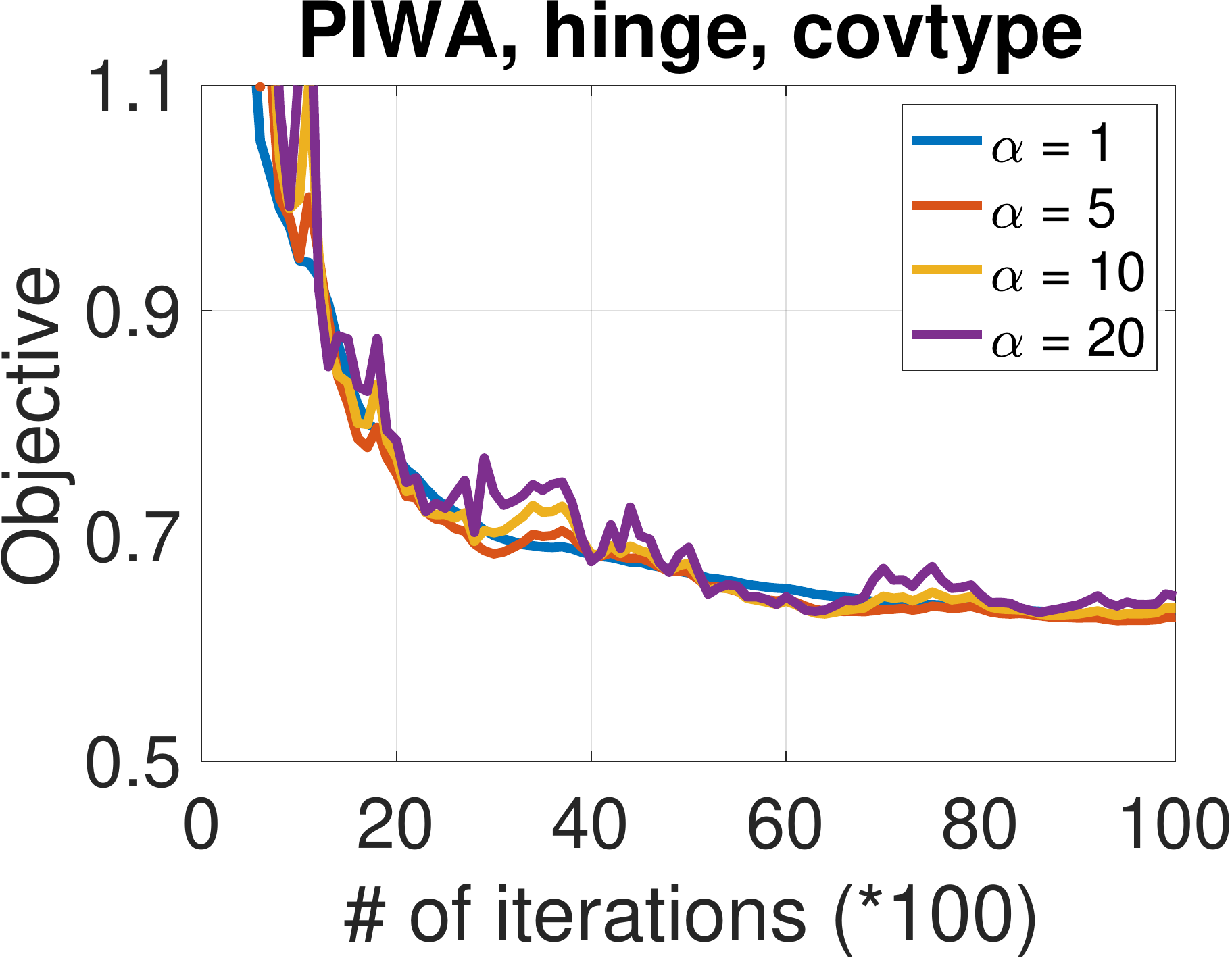}
\includegraphics[scale=0.2]{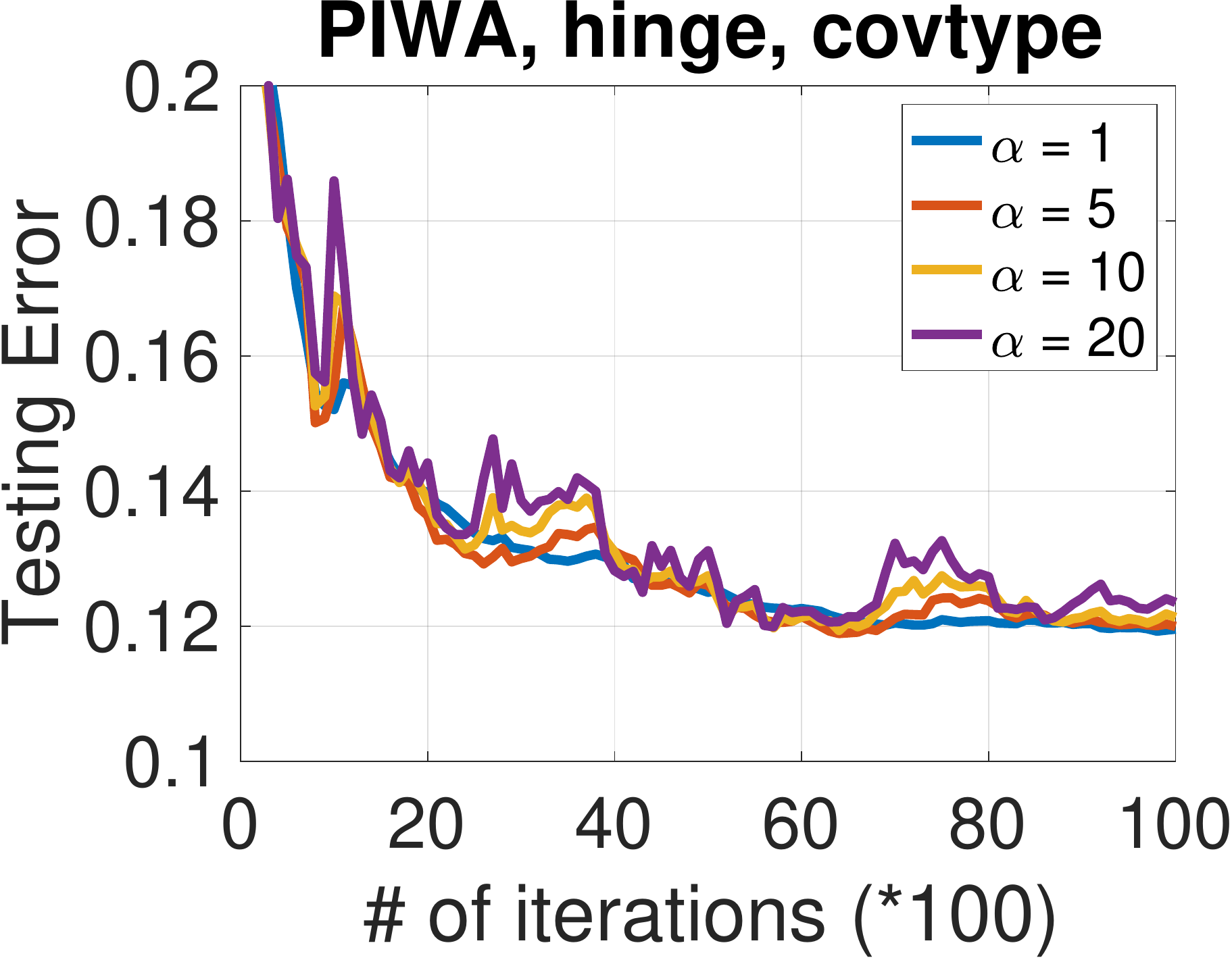}\\
\includegraphics[scale=0.2]{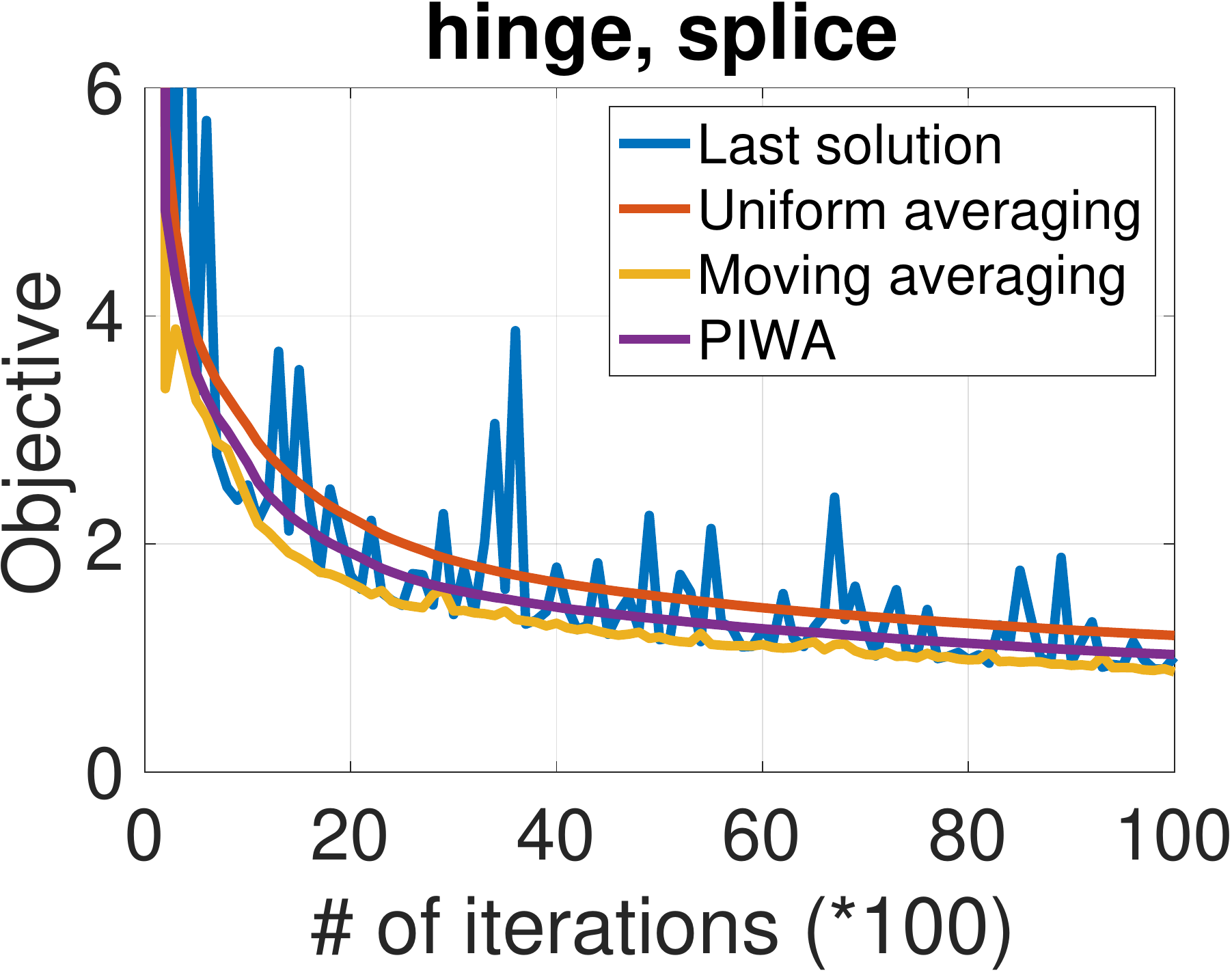}
\includegraphics[scale=0.2]{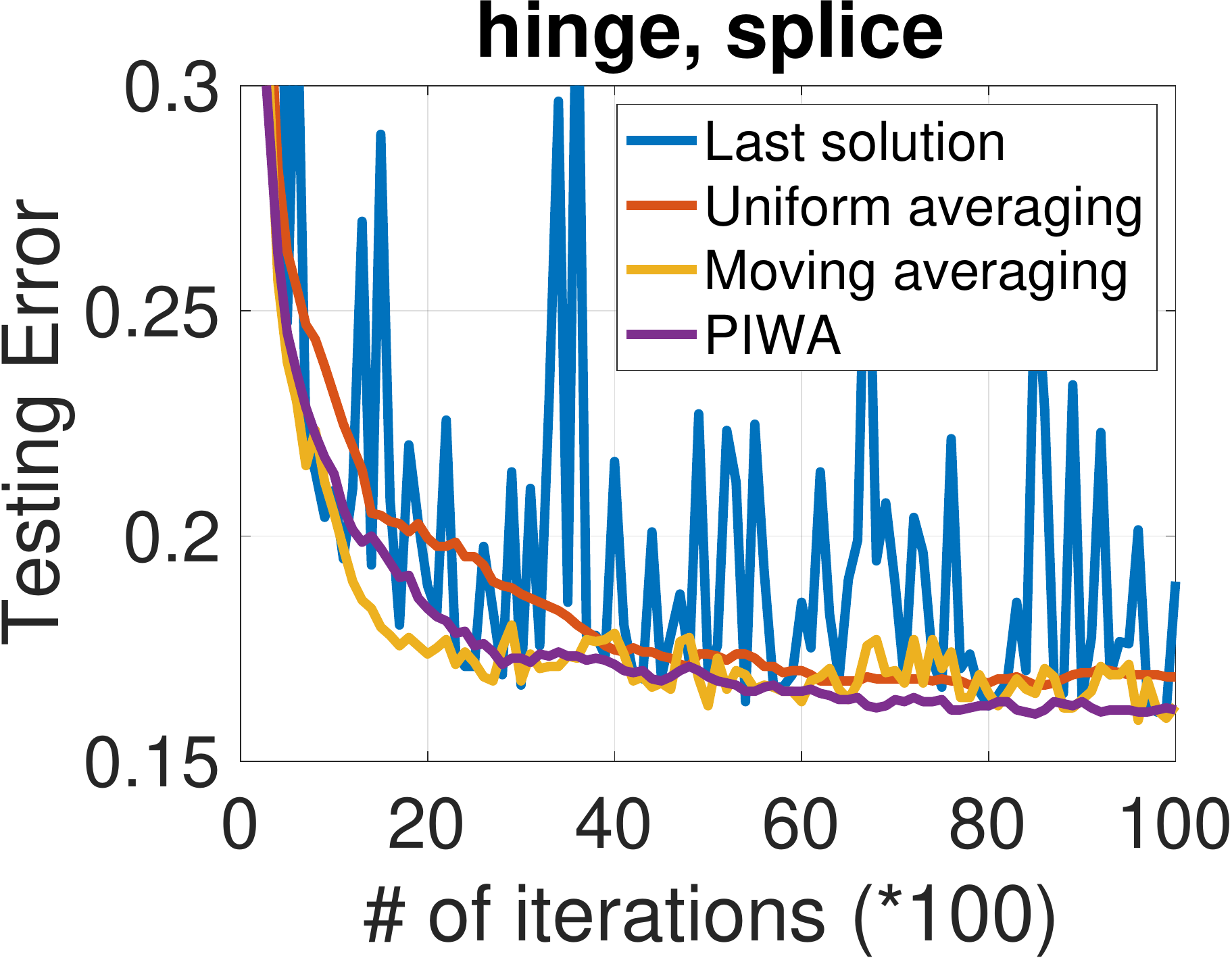}
\includegraphics[scale=0.2]{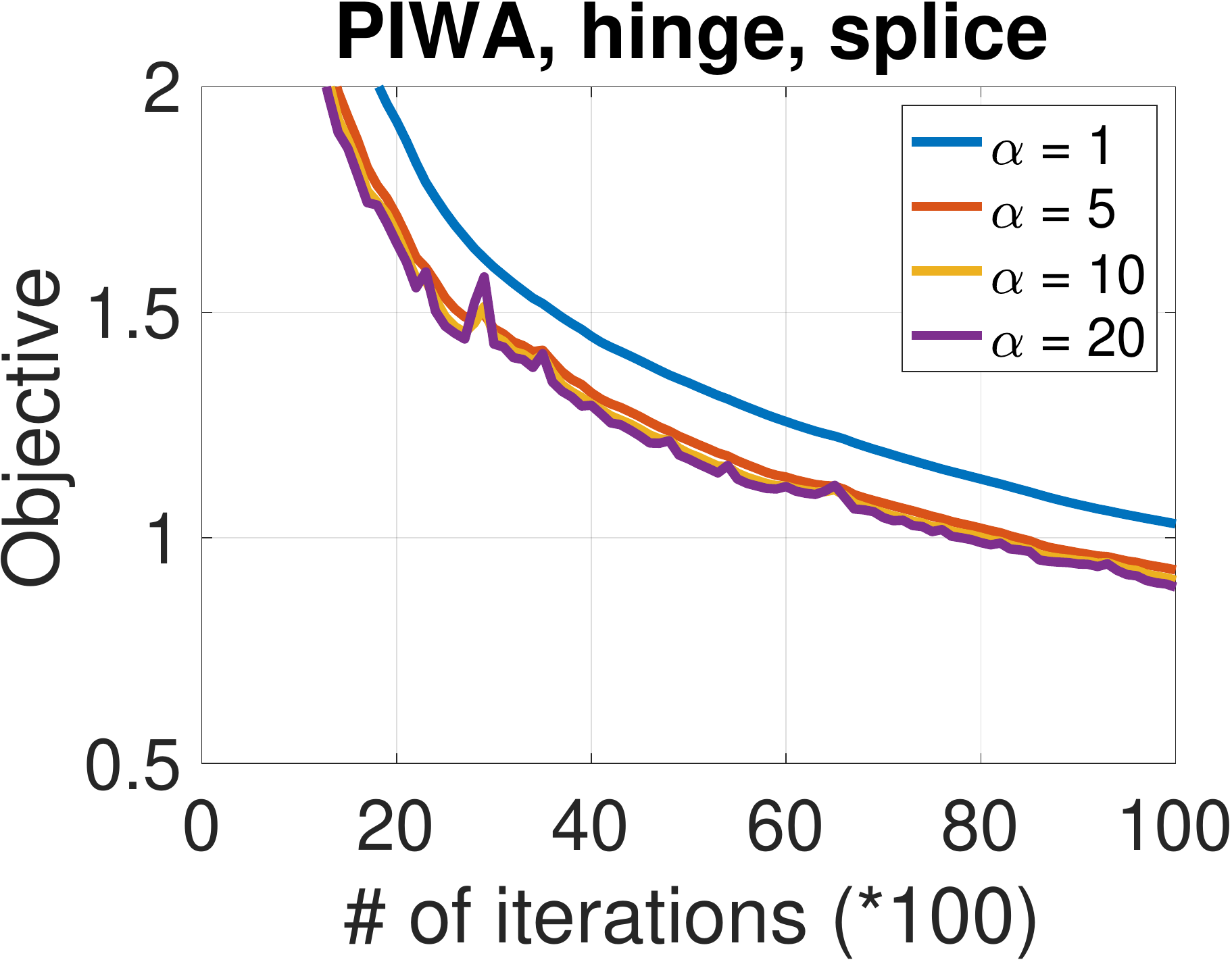}
\includegraphics[scale=0.2]{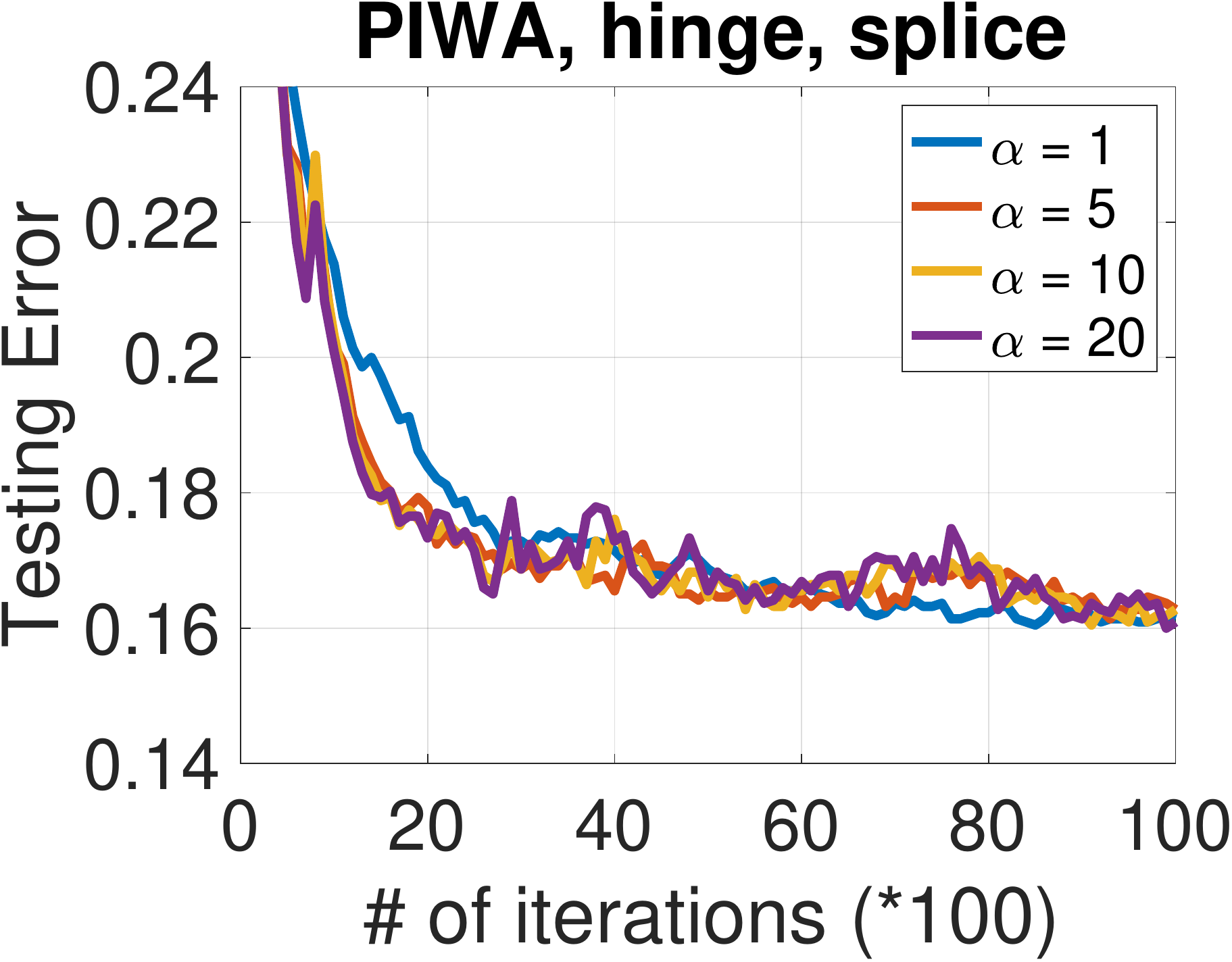}\\
\includegraphics[scale=0.2]{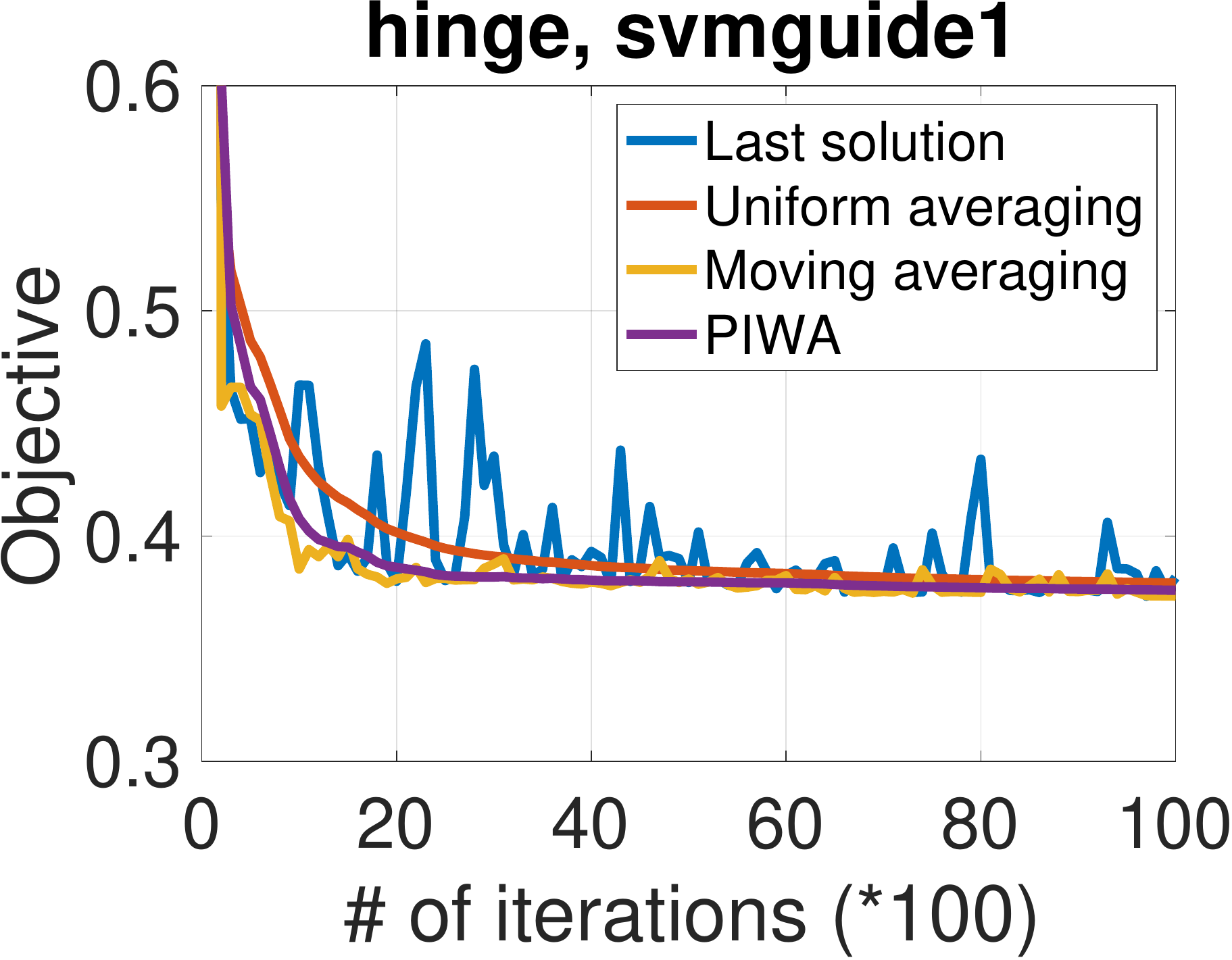}
\includegraphics[scale=0.2]{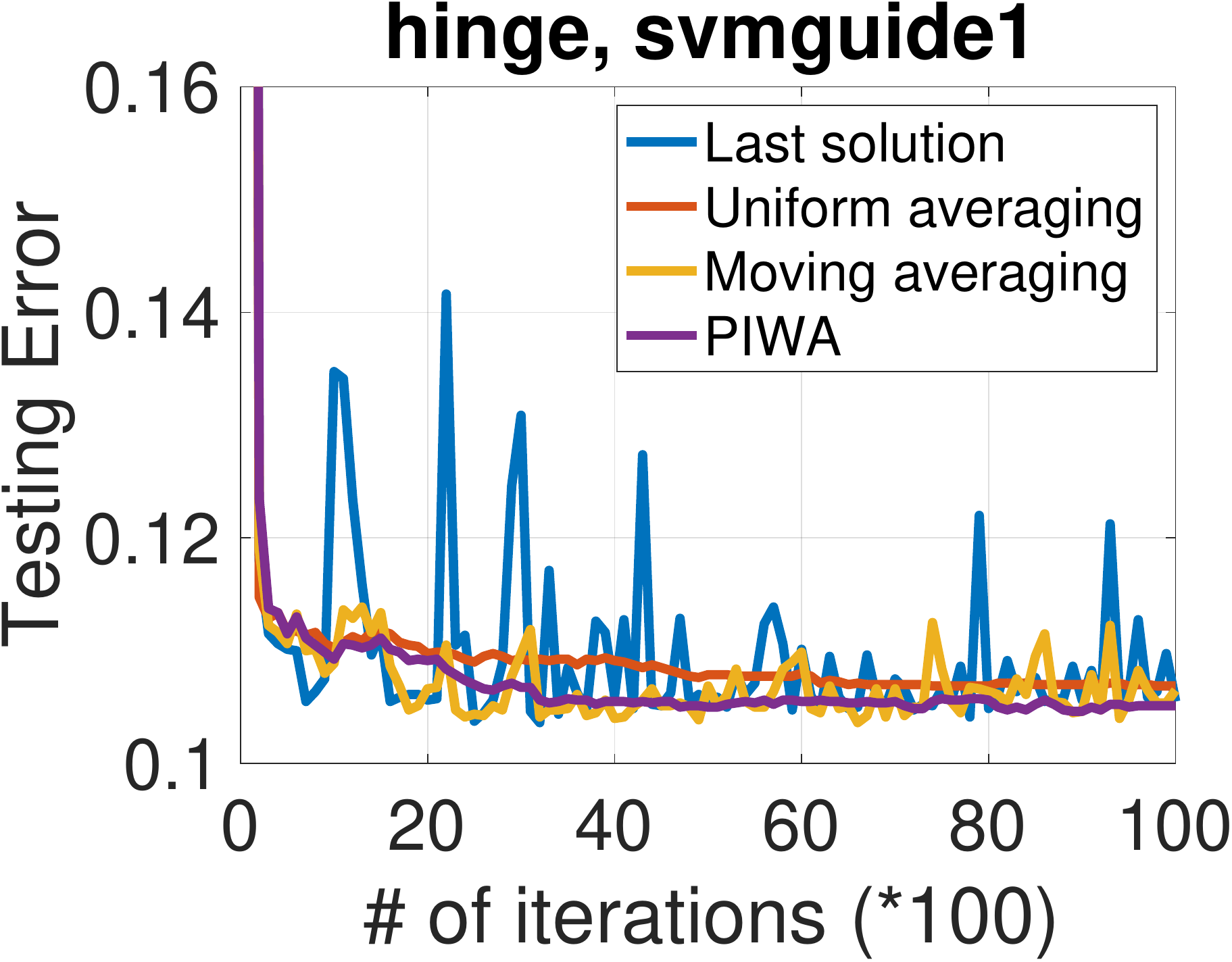}
\includegraphics[scale=0.2]{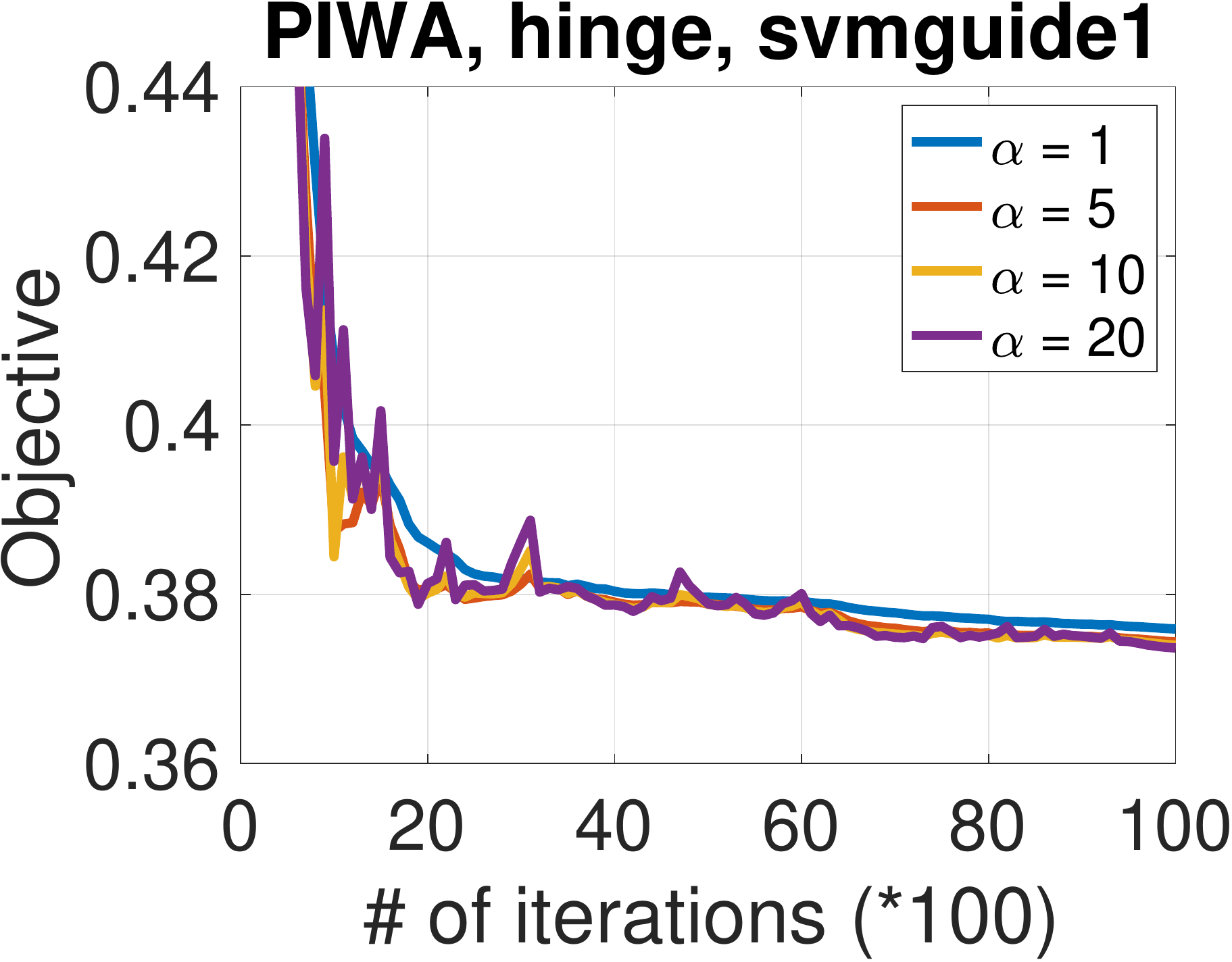}
\includegraphics[scale=0.2]{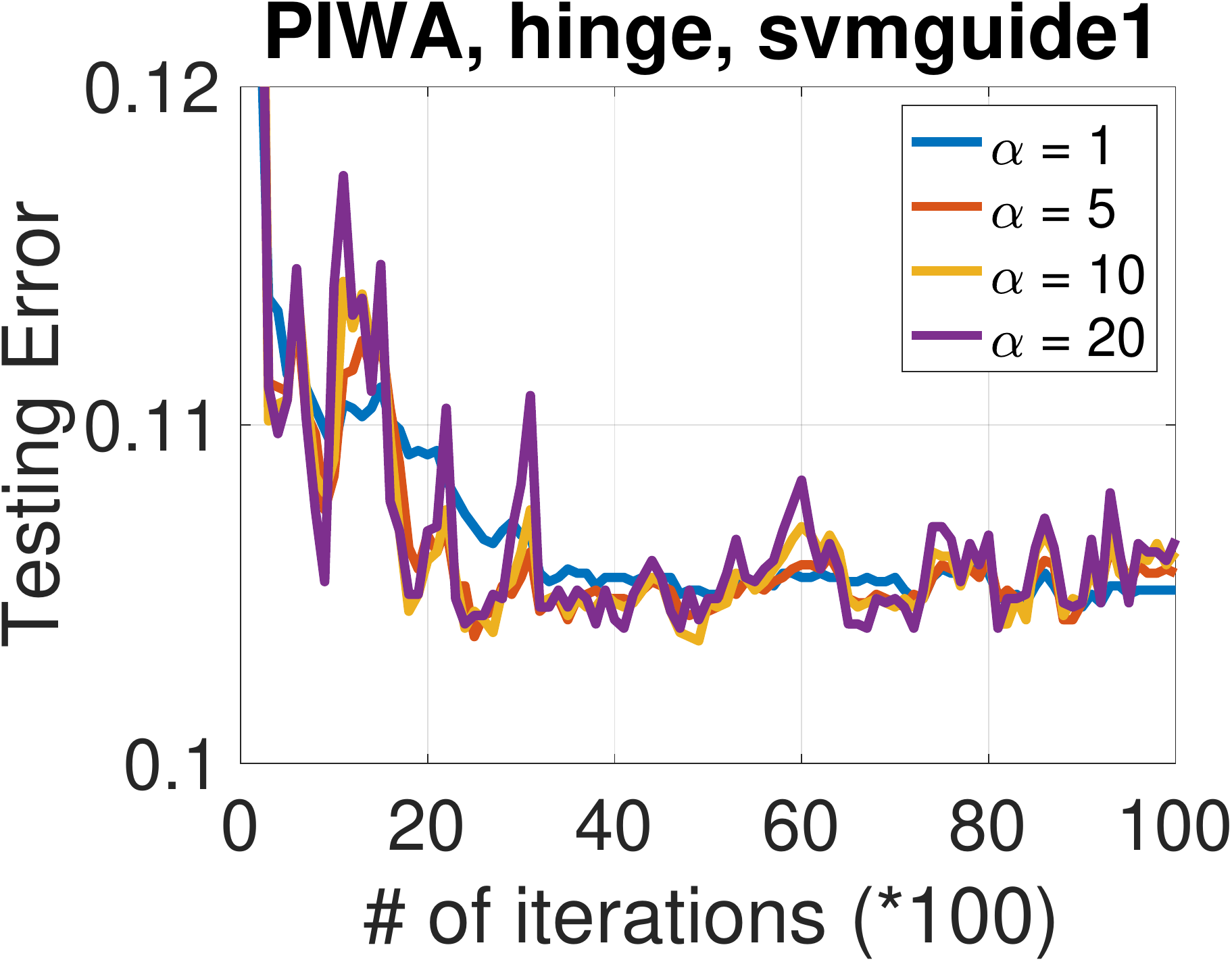}\\
\includegraphics[scale=0.2]{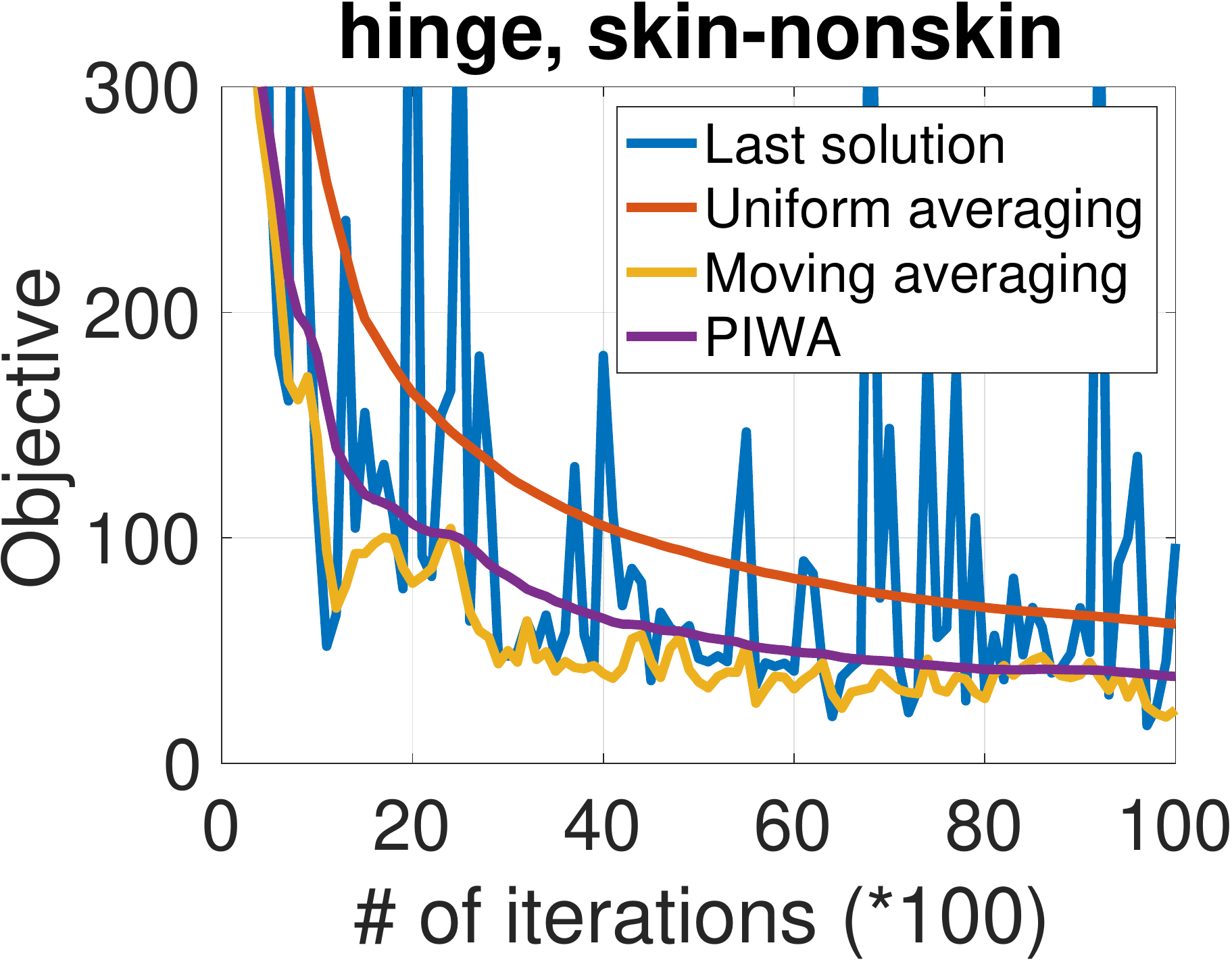}
\includegraphics[scale=0.2]{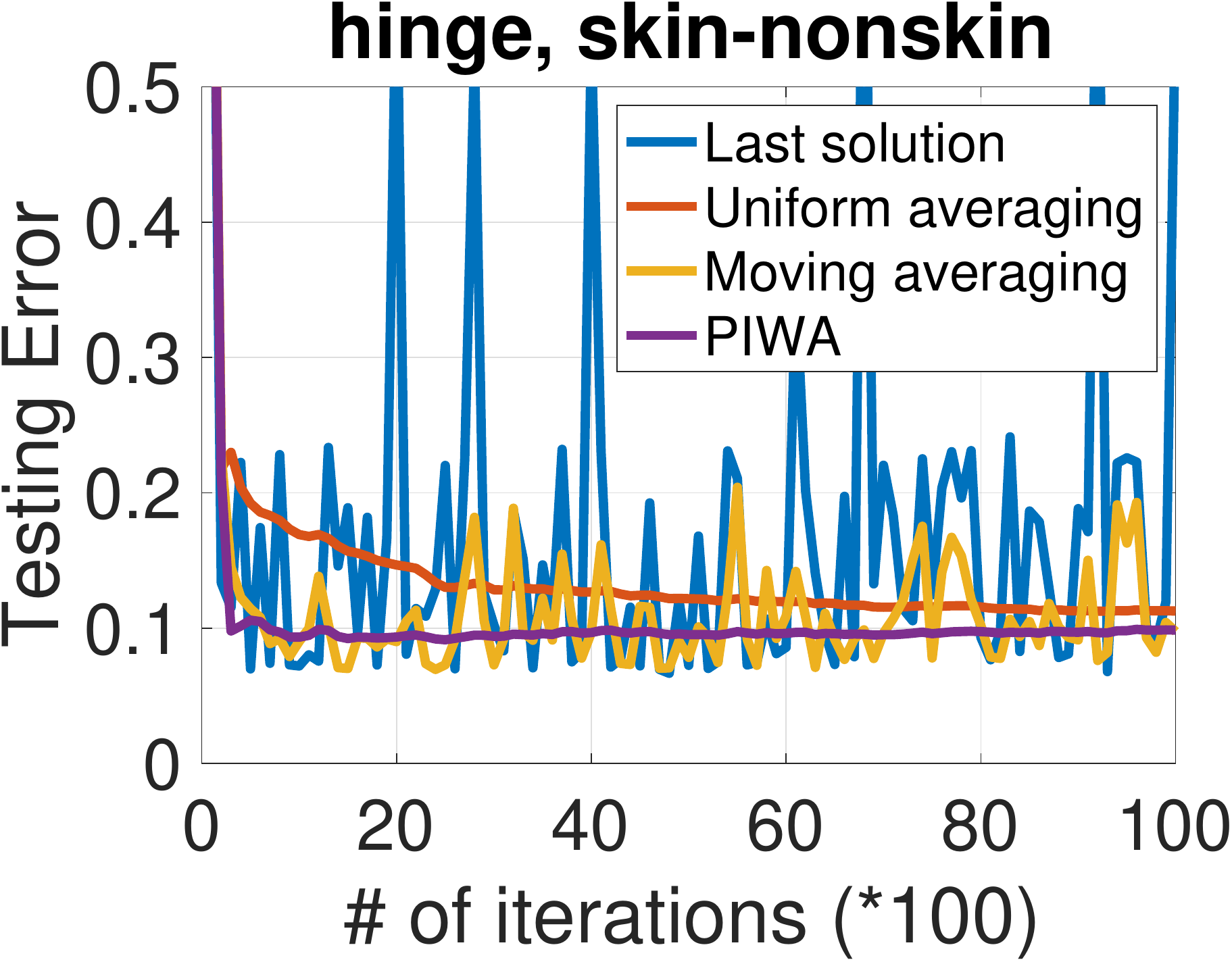}
\includegraphics[scale=0.2]{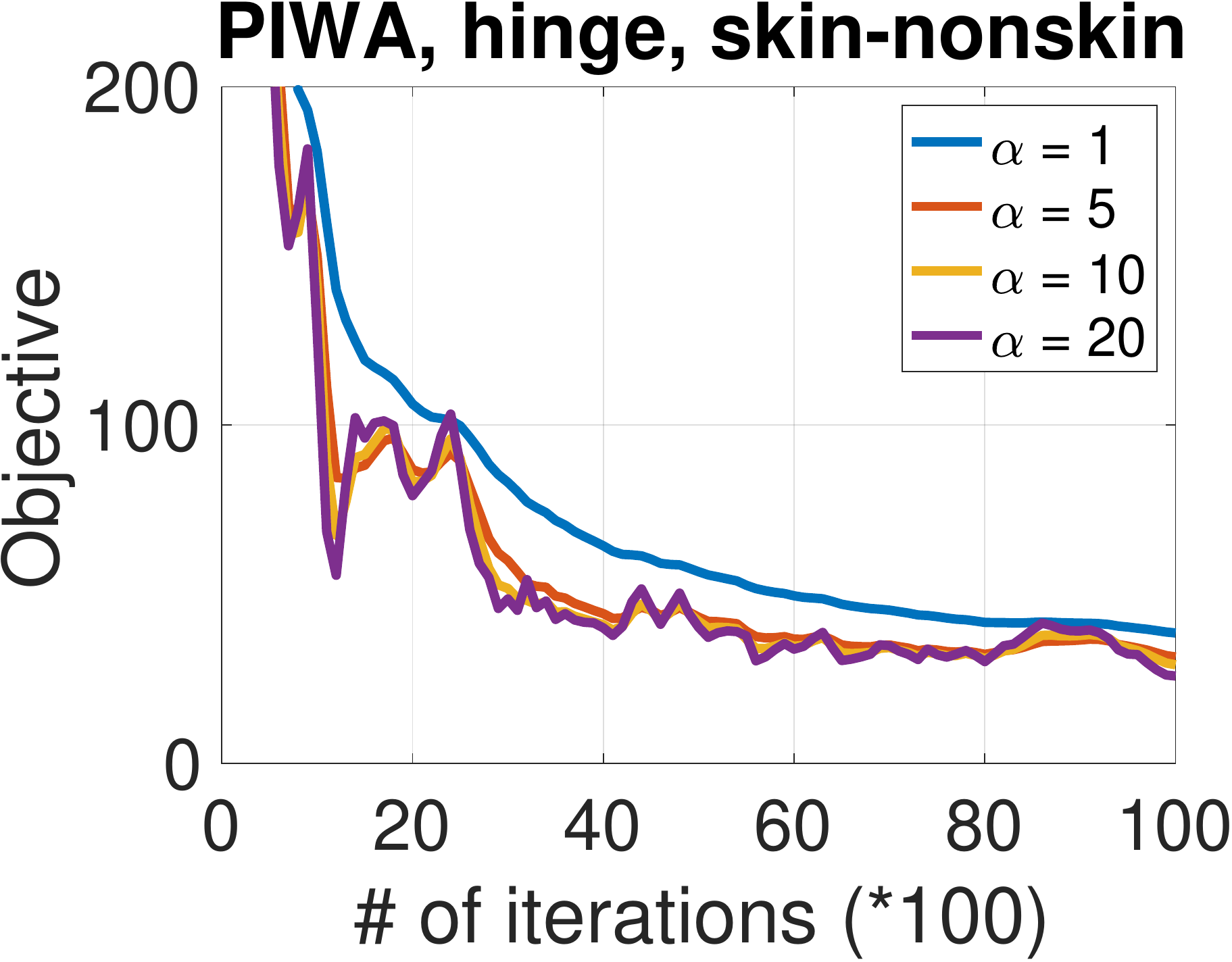}
\includegraphics[scale=0.2]{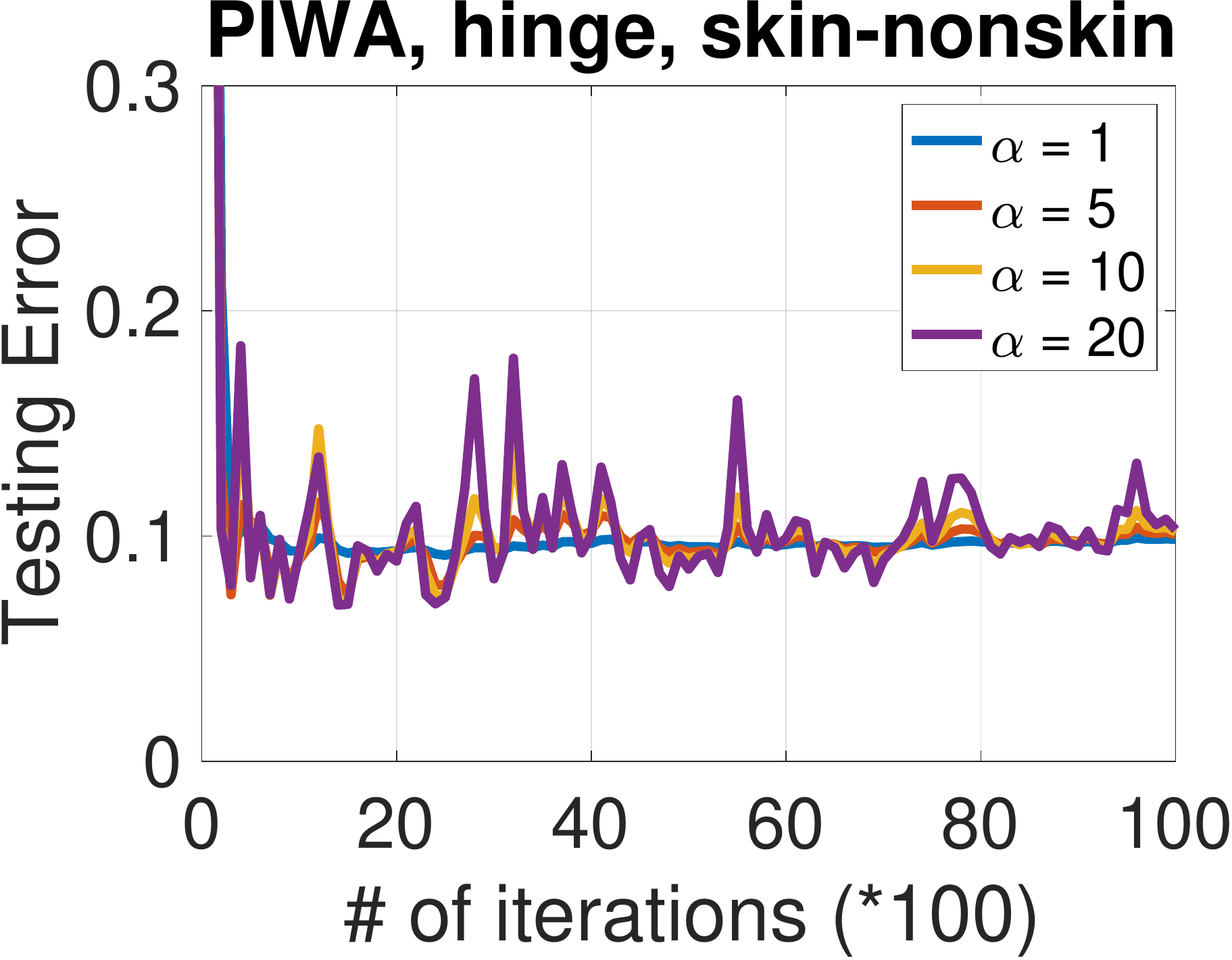}
\vspace*{-0.1in}
\caption{Experiments of convex functions: hinge loss.
         }
\label{figure:convex_results}
\end{figure*}

\noindent
\textbf{Remark.} 
Combining the above result with  Lemma \ref{pl_imply}, we can see that $E[\|x_k - x_*\|]$ is also bounded, which indicates that  if the optimal point is within the constraint ball of the initial solution, then after this stage, the optimal point is still in a constraint ball of the averaged solution with high probability.
In the following theorem, we are going to make the bounded ball smaller after each stage.

\begin{theorem}
Suppose \textbf{Assumption 3} holds, and $f(x; z)$ is $\rho$-weakly convex with $\rho \leq \mu/4$.
Then by setting $\eta_k = \frac{\epsilon_k c}{2\hat{G}^2} \leq 1/L$ 
and $T_k = \frac{d}{\mu\epsilon_k}, $ and $\gamma = 4/\mu$, 
where $c \leq \min (1, \frac{2\hat{G}^2}{L\epsilon_0})$, $d = \max\{\frac{32(\alpha+1)\hat{G}^2}{c} , 512(\alpha+1)^2\hat{G}^2 \ln(1/\delta)\}$, and after $K = \log(\epsilon_0/\epsilon)$ stages, with probability $(1 - \delta)^K$ we have
\begin{equation*}
\begin{split}
&F_{\mathcal S}(x_K) -F_{\mathcal{S}} (x_*))\leq \epsilon. 
\end{split}
\end{equation*}
\label{non_optimization}
\end{theorem}
\noindent
\textbf{Remark.} It is easy to see that the total iteration complexity is in the order $O((\alpha+1)/(\mu\epsilon))$, which is similar as $O(1/(\mu\epsilon))$ in ~\cite{yang2018does} that uses a uniform averaging.

\subsubsection{Generalization Error}

We now establish the uniform stability in the following theorem.
\begin{theorem}
Let $S_{K-1} = \sum\limits_{k=1}^{K-1} T_k$ and $\eta_K \leq c/(\mu T_K)$.
By the same assumptions and setting as Theorem \ref{non_optimization},
we have
\begin{equation*}
\begin{split}
&\epsilon_{stab}\leq \frac{S_{K-1}}{n}+\frac{1+\frac{1}{L c}}{n-1}\left(2(\alpha+1)cL^2\right)^{\frac{1}{1+Lc}} T^{\frac{Lc}{1+Lc}}
\end{split}
\end{equation*}

\end{theorem}

\noindent
\textbf{Remark.} We apply the similar conditional analysis for the non-convex objective function as in \cite{hardt2015train}.
In particular, we condition on $x_{K-1} = x_{K-1}'$, i.e., the different example will be used within the last stage, and prove the bound for $\|x_K - x_K'\|$.
We can see the bound is increasing in $\alpha$.

\vspace*{-0.1in}
\section{Experiments}

\begin{figure*}[htbp]
\centering
\includegraphics[scale=0.2]{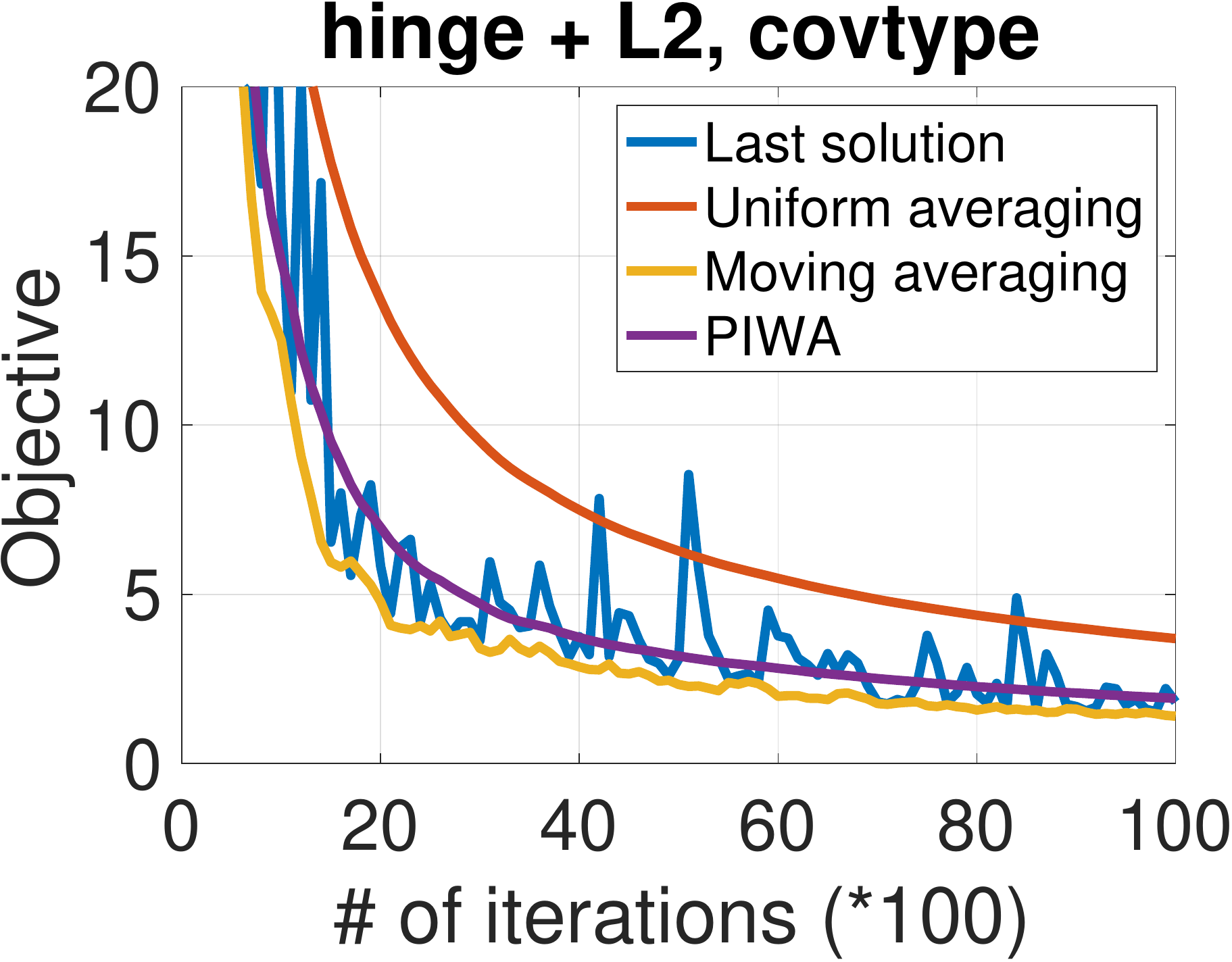}
\includegraphics[scale=0.2]{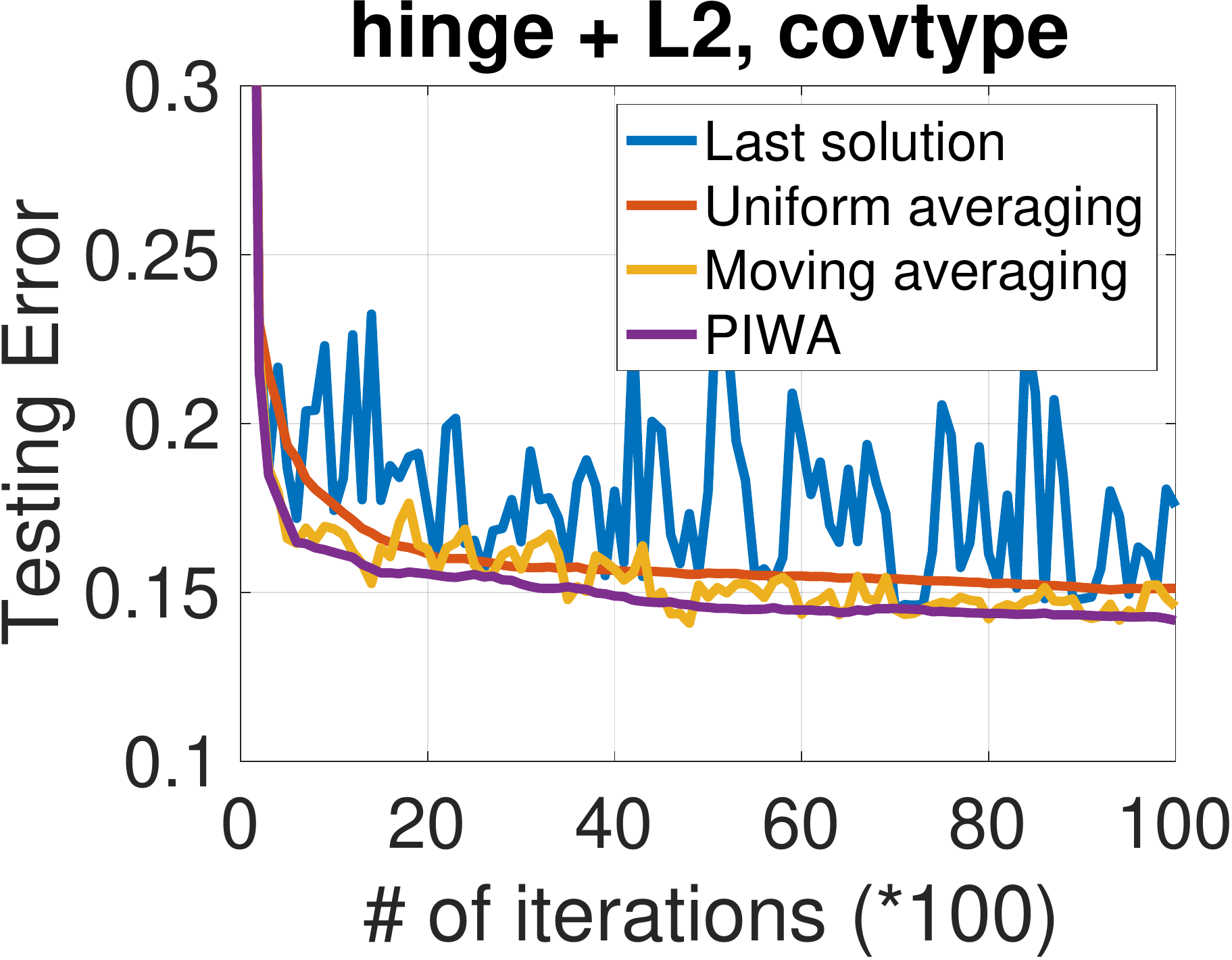}
\includegraphics[scale=0.2]{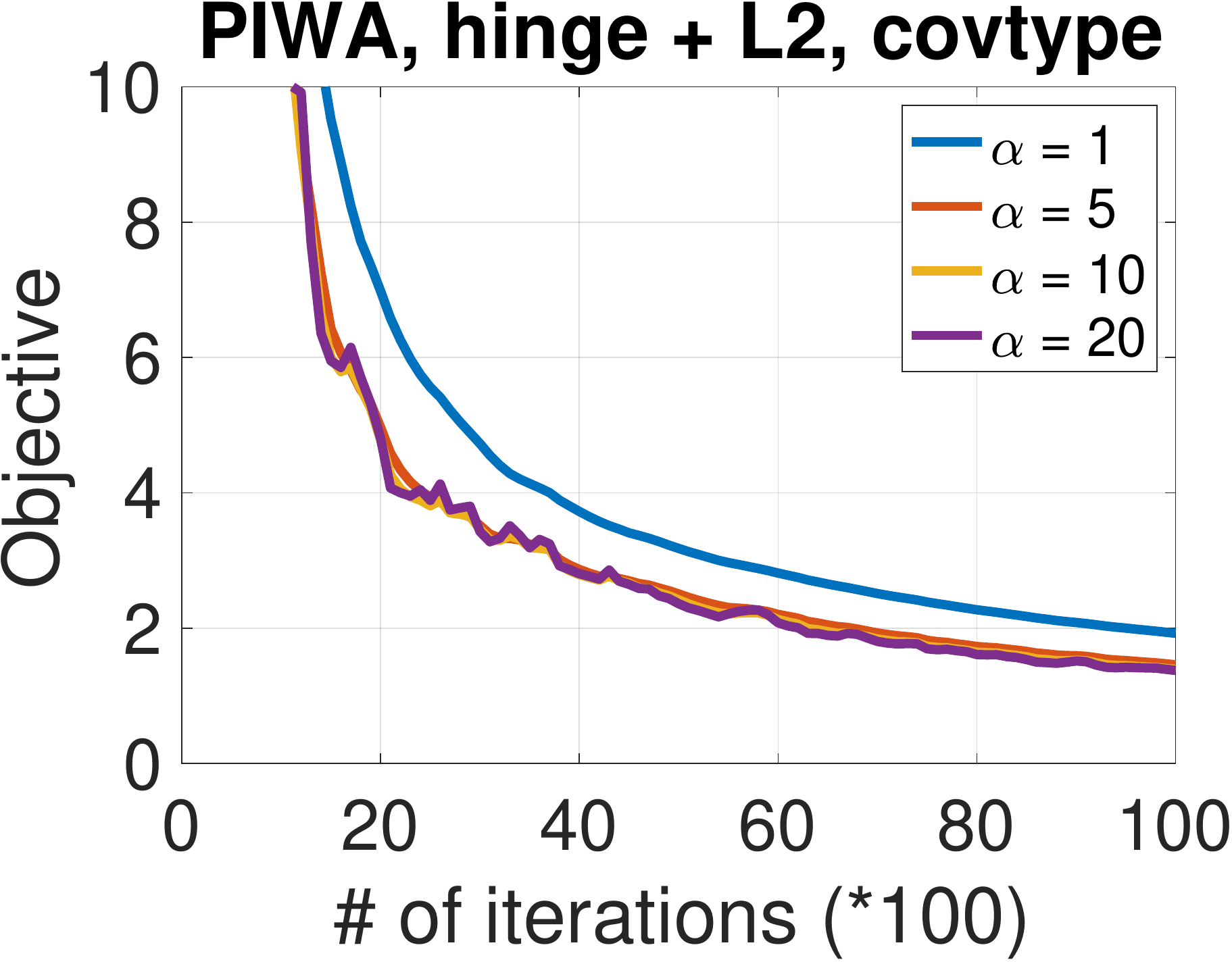}
\includegraphics[scale=0.2]{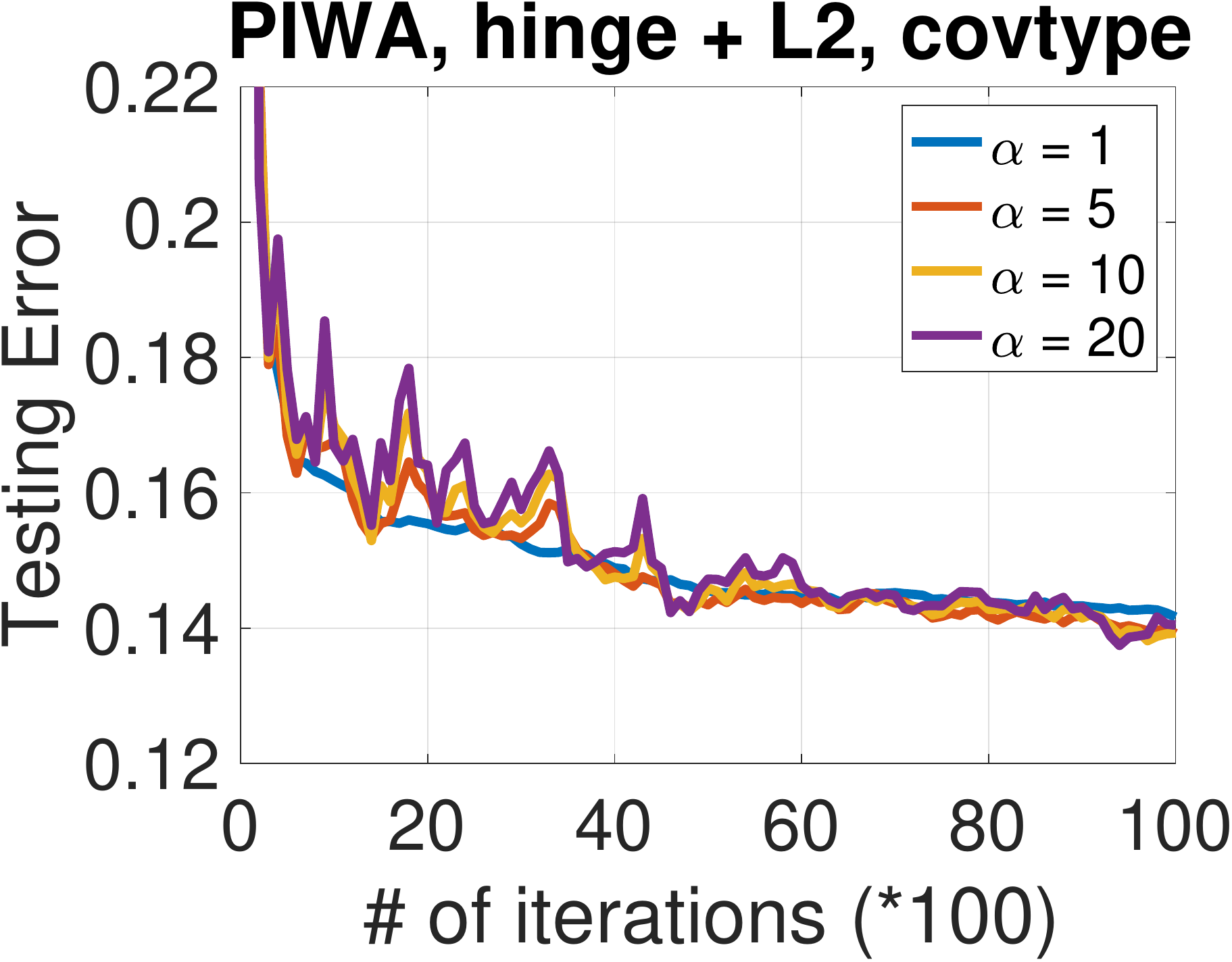}\\
\includegraphics[scale=0.2]{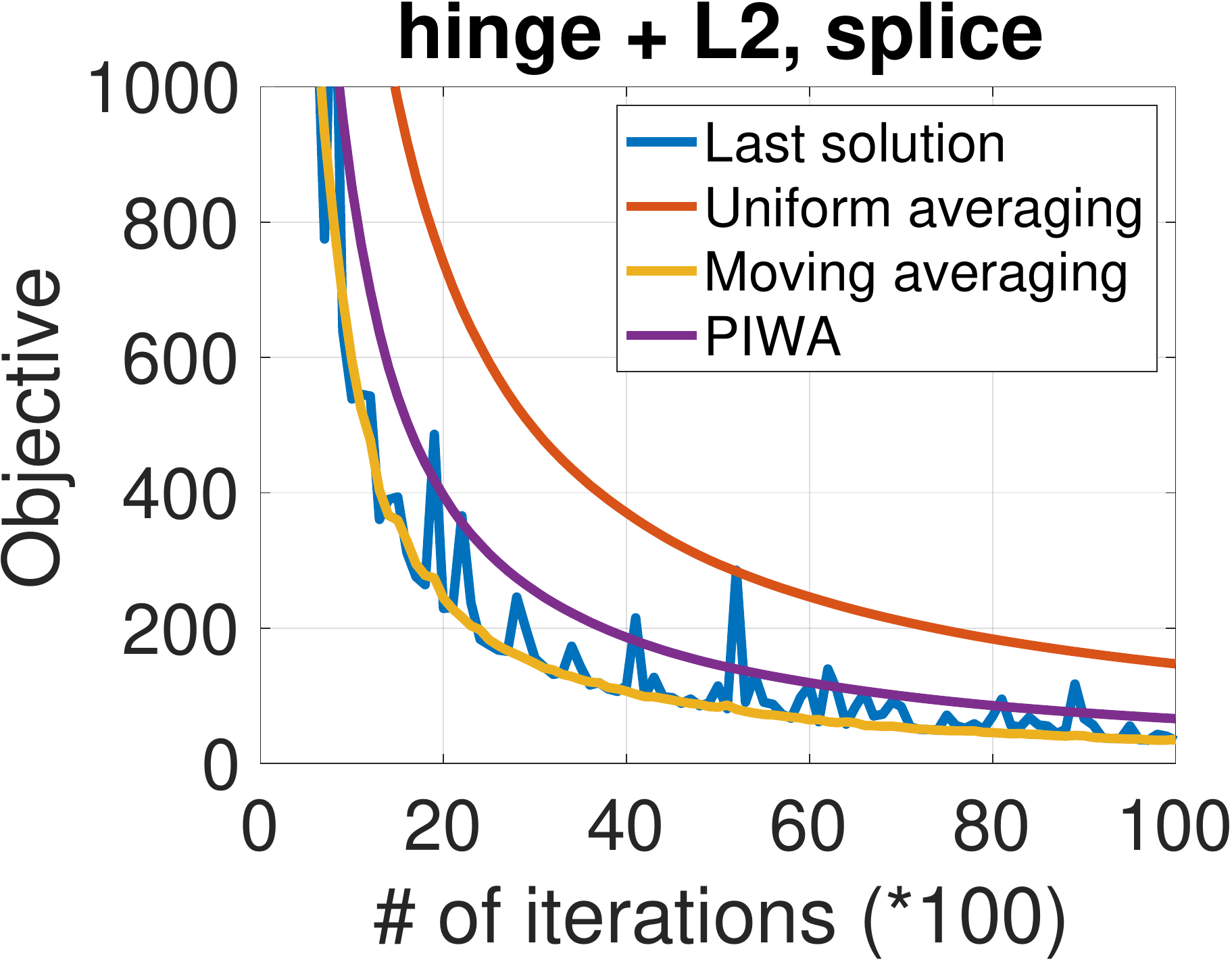}
\includegraphics[scale=0.2]{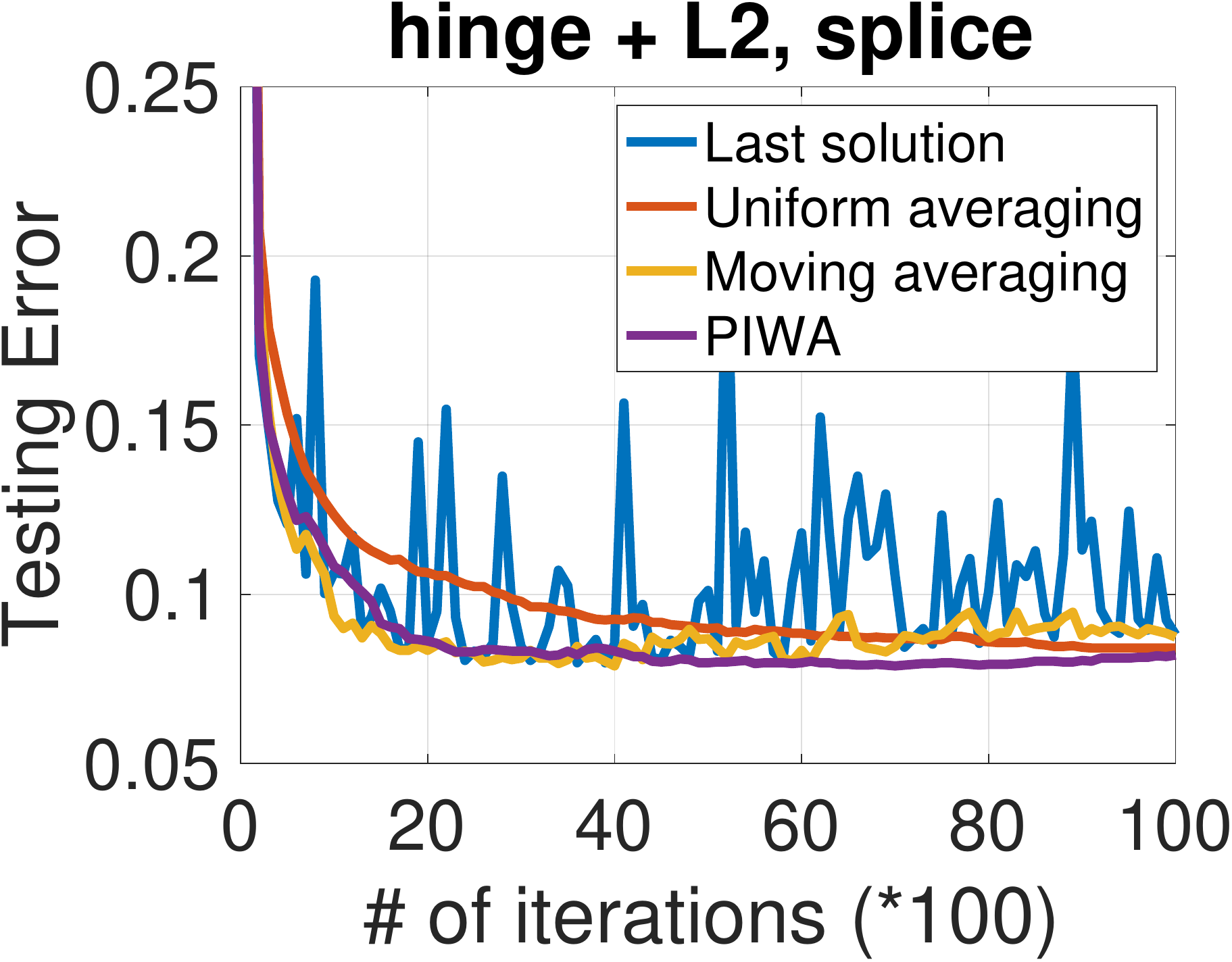}
\includegraphics[scale=0.2]{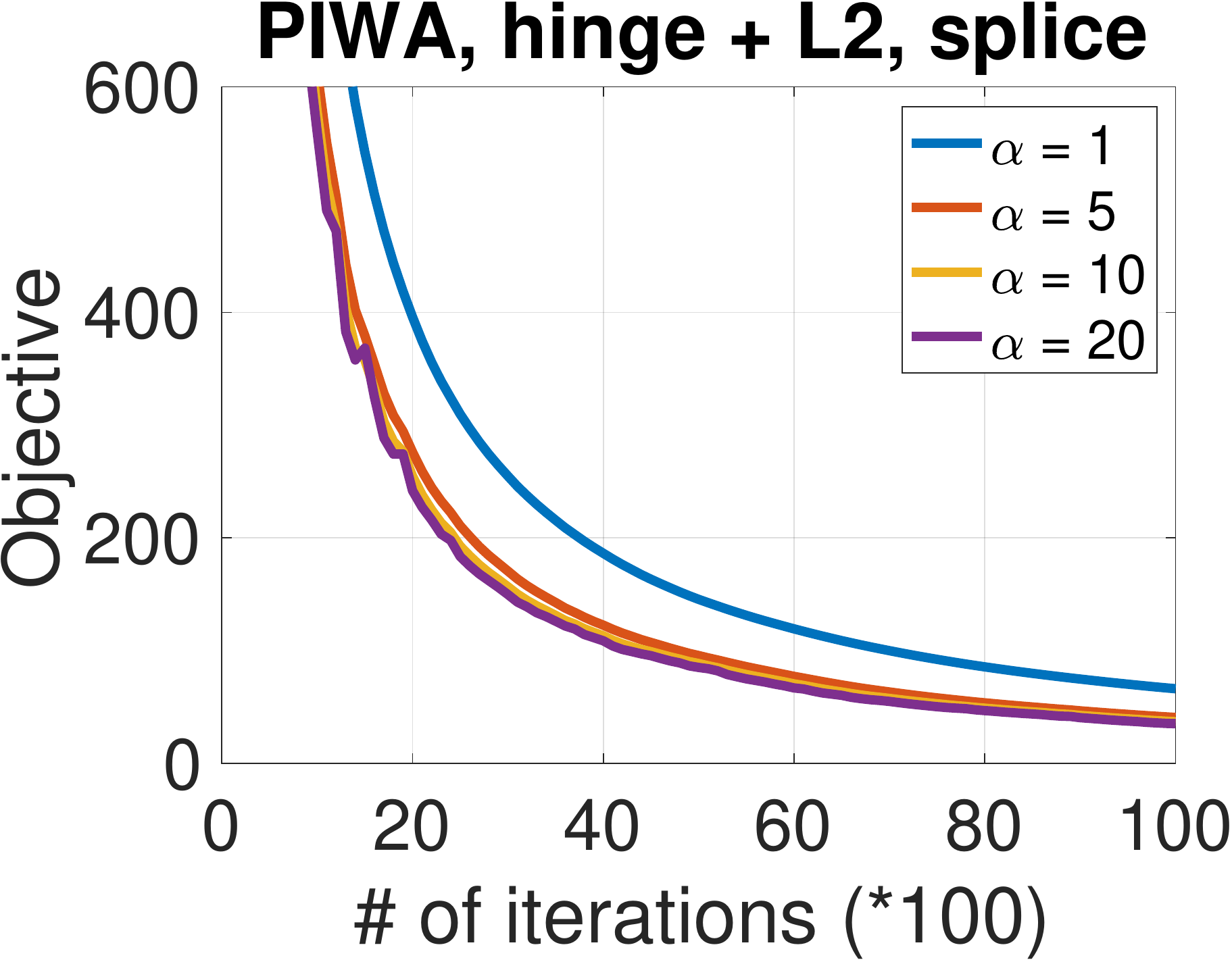}
\includegraphics[scale=0.2]{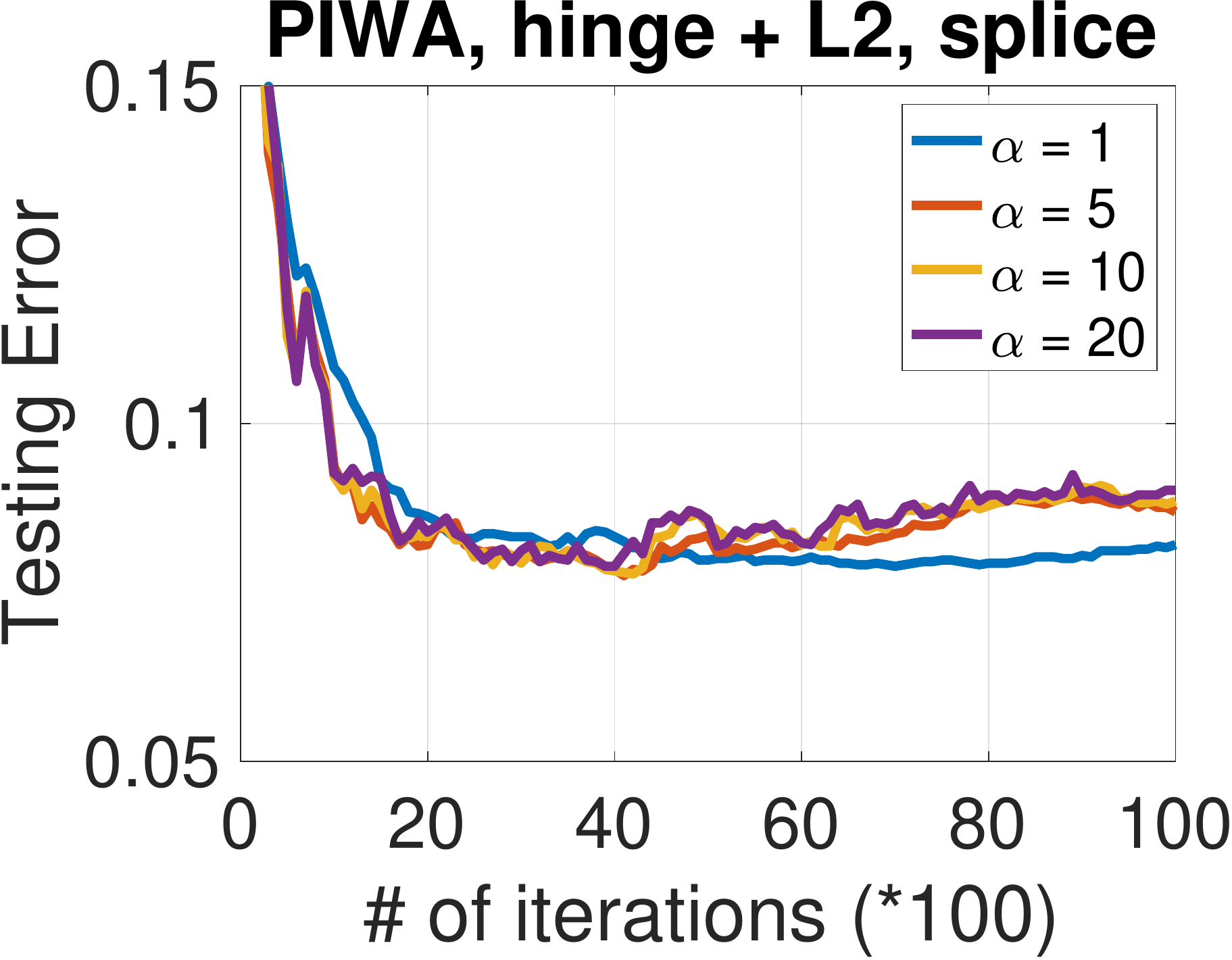}\\
\includegraphics[scale=0.2]{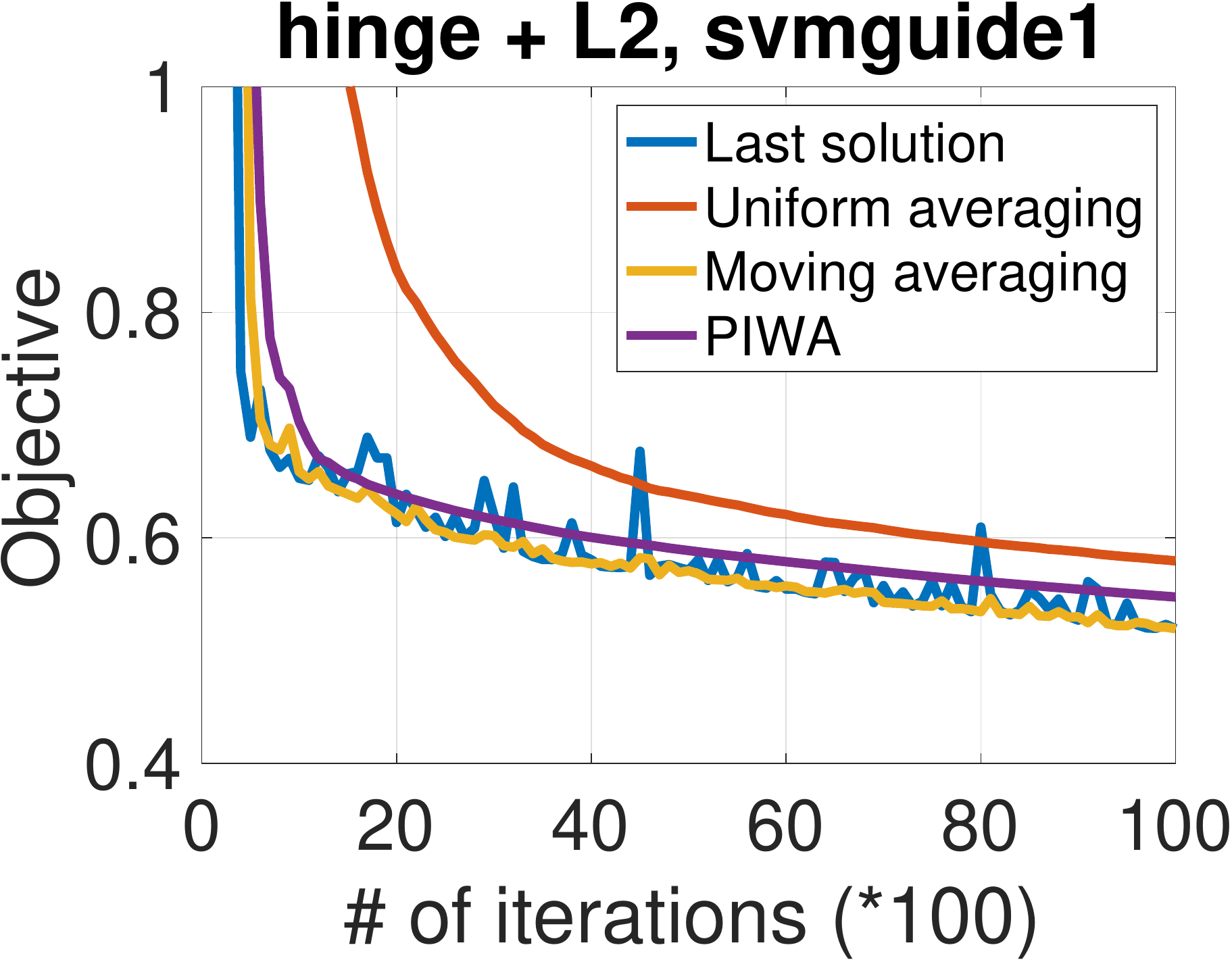}
\includegraphics[scale=0.2]{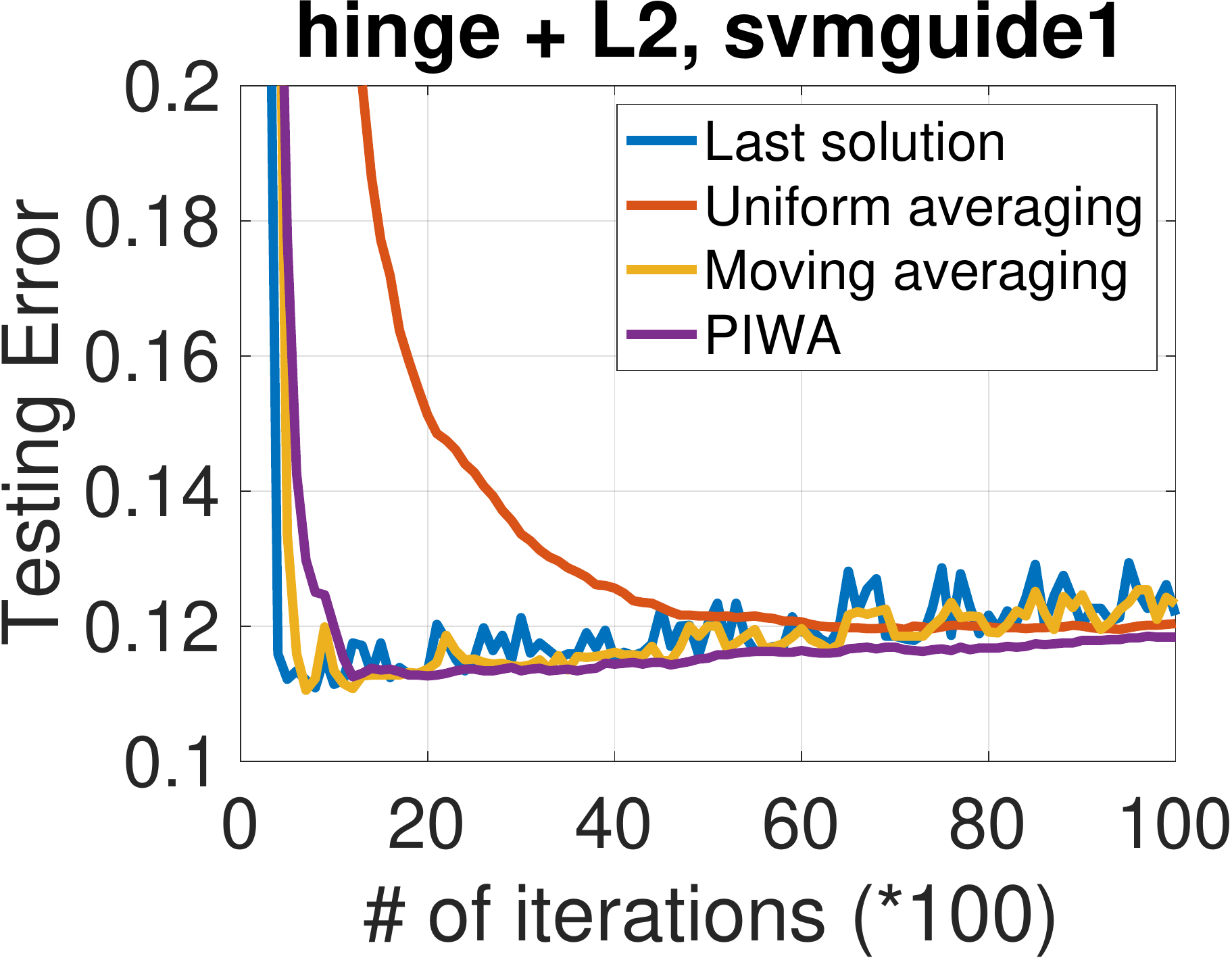}
\includegraphics[scale=0.2]{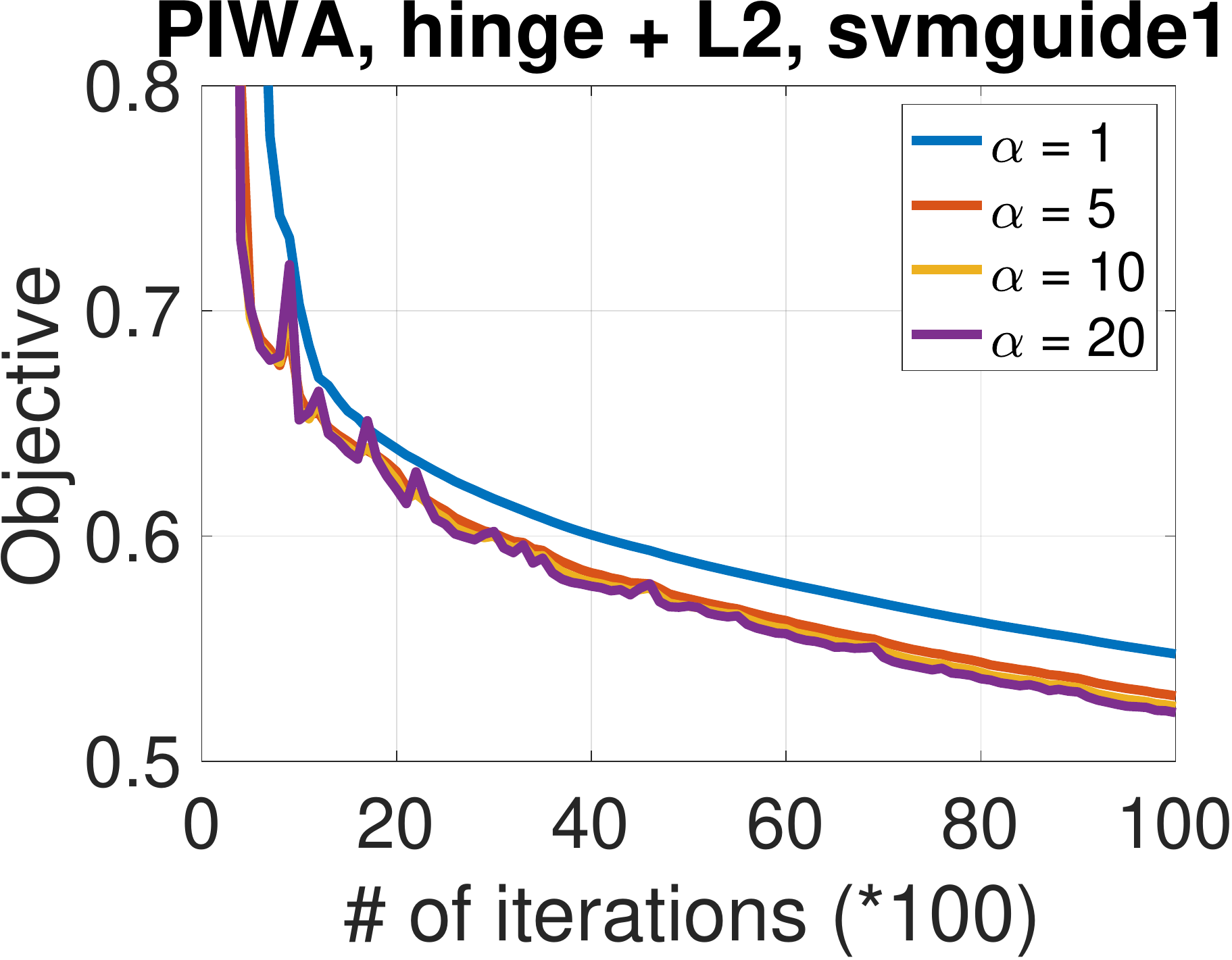}
\includegraphics[scale=0.2]{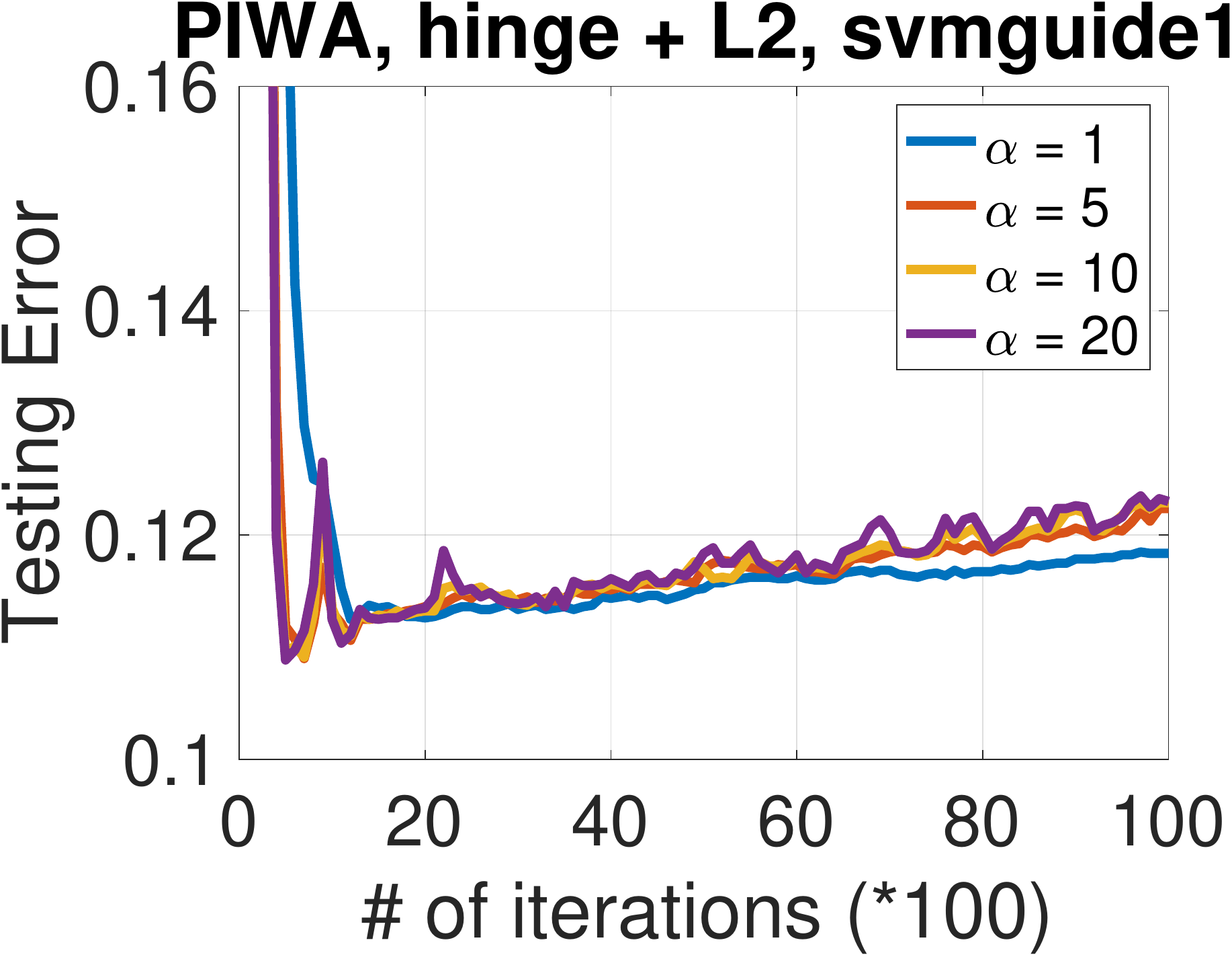}\\
\includegraphics[scale=0.197]{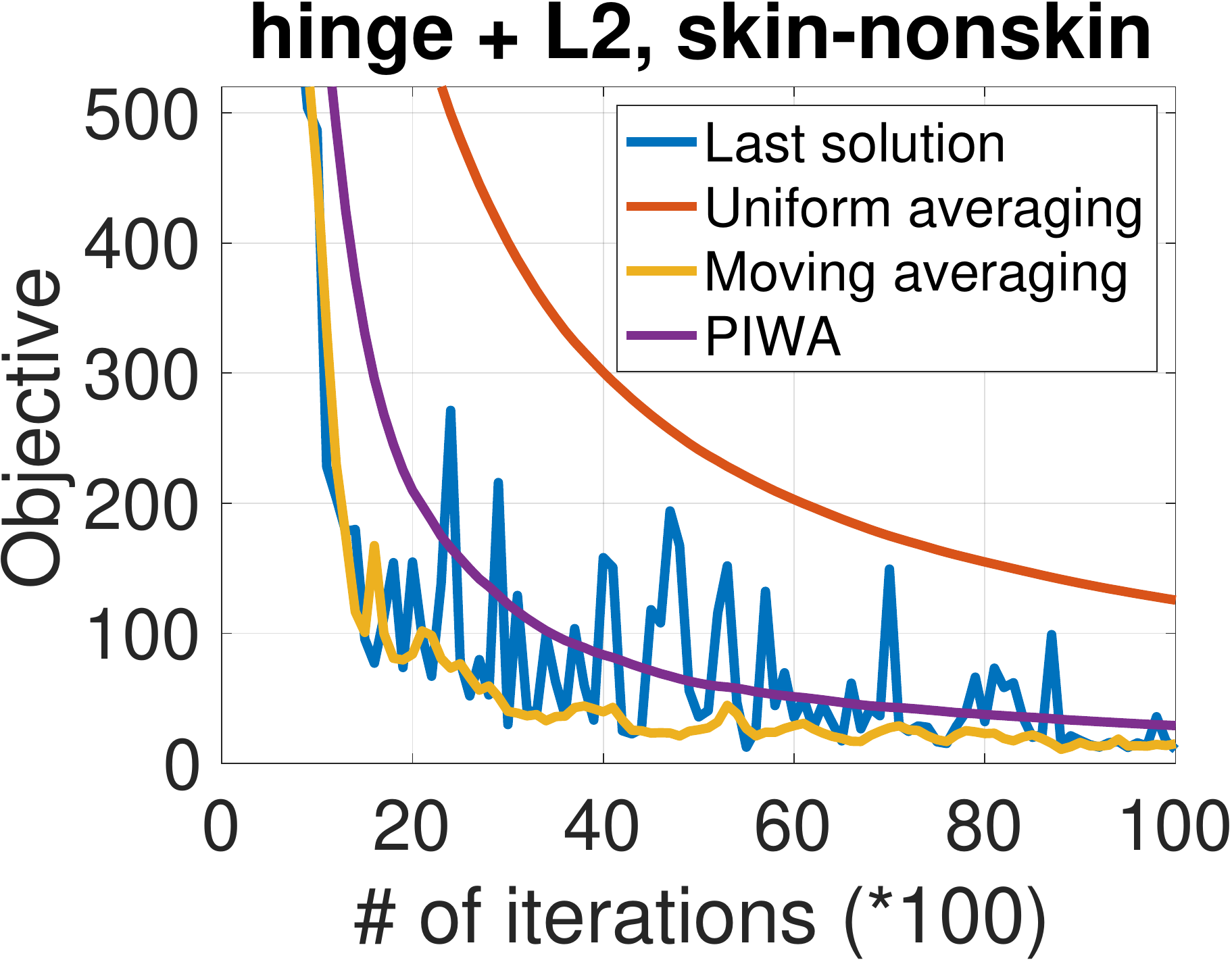}
\includegraphics[scale=0.197]{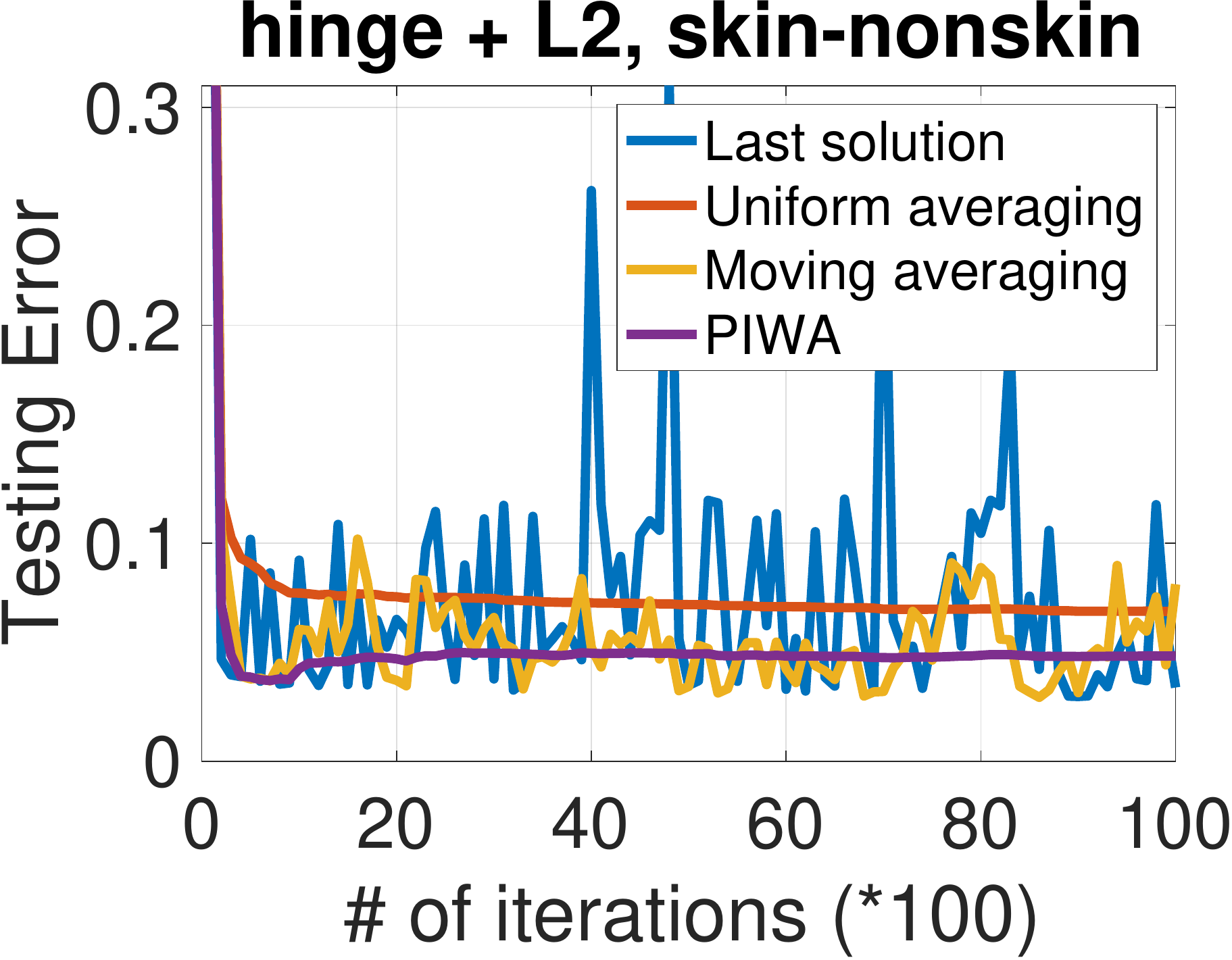}
\includegraphics[scale=0.197]{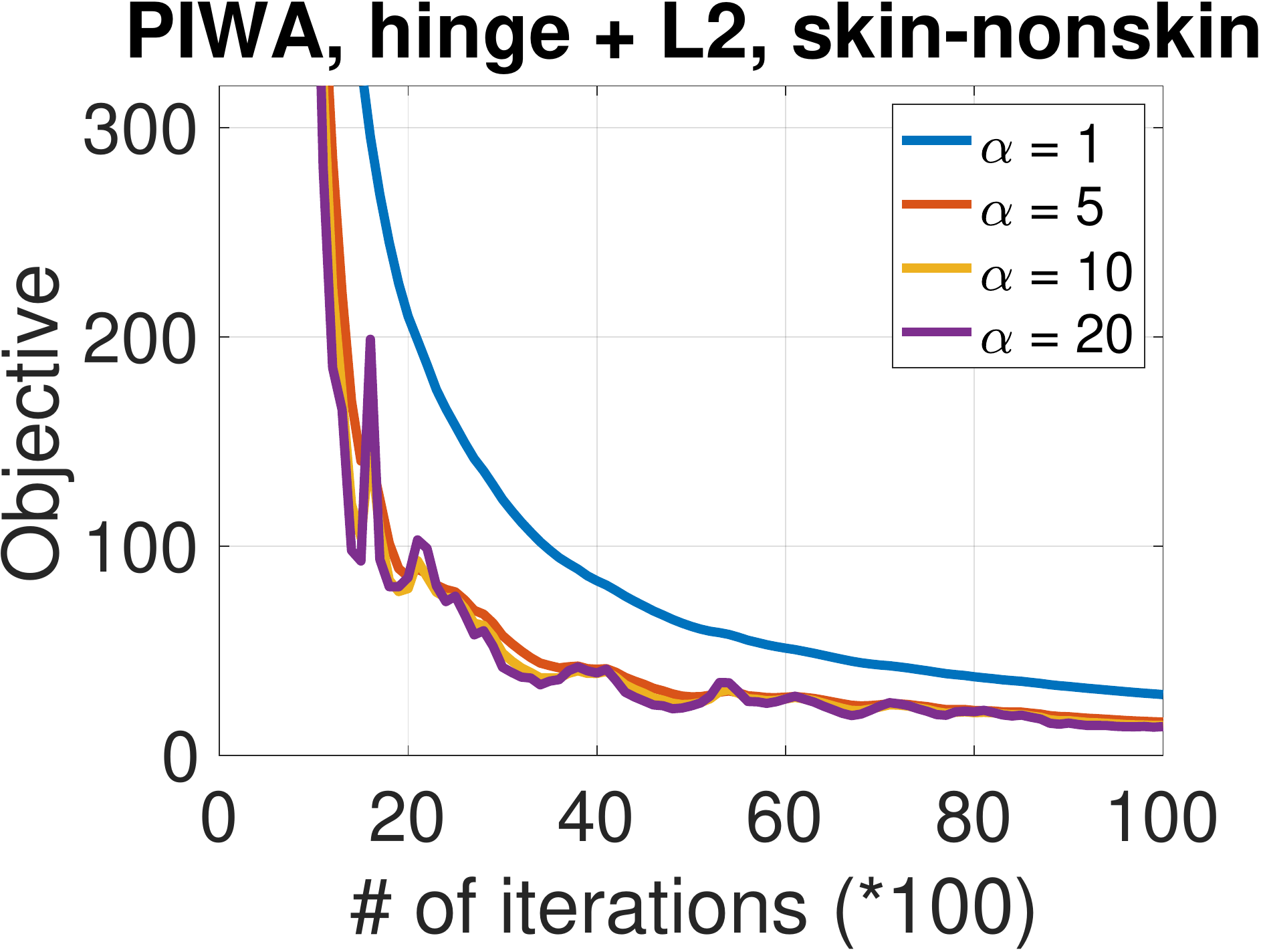}
\includegraphics[scale=0.197]{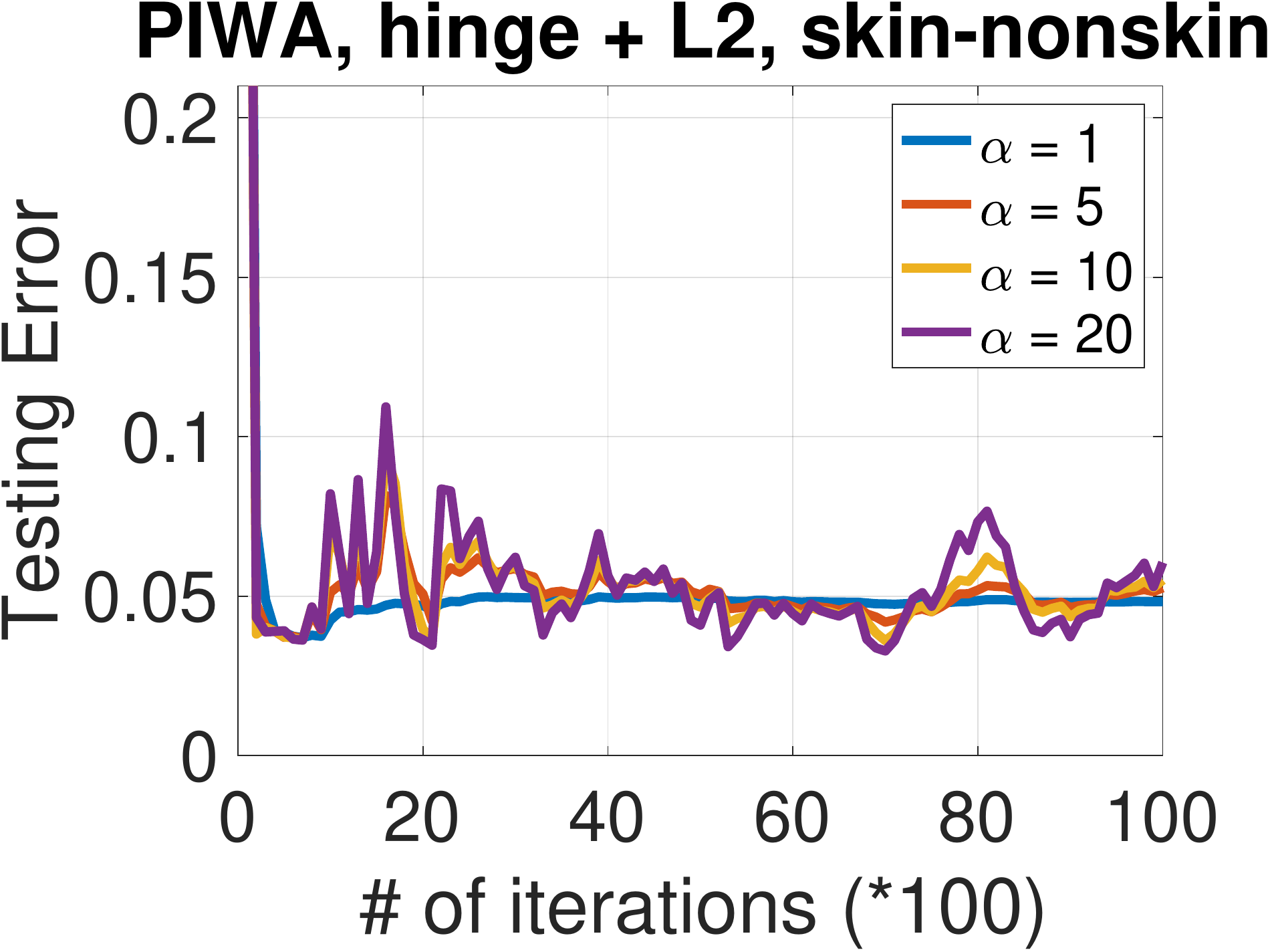}
\caption{Experiments of strongly convex objective functions: hinge loss + L2 regularization.}
\label{figure:strongly_convex_results}
\vspace*{0.1in}
\includegraphics[scale=0.2]{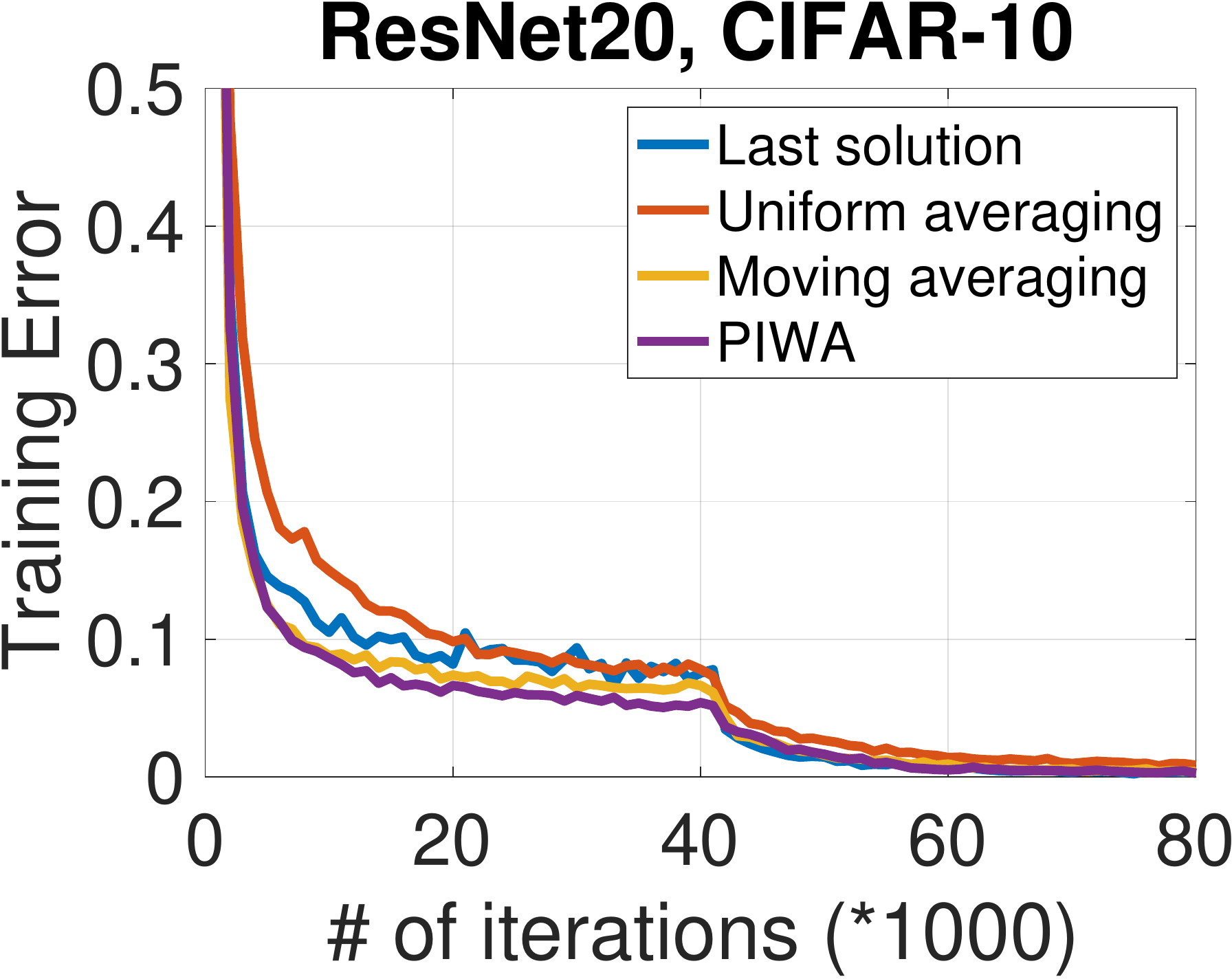}
\includegraphics[scale=0.2]{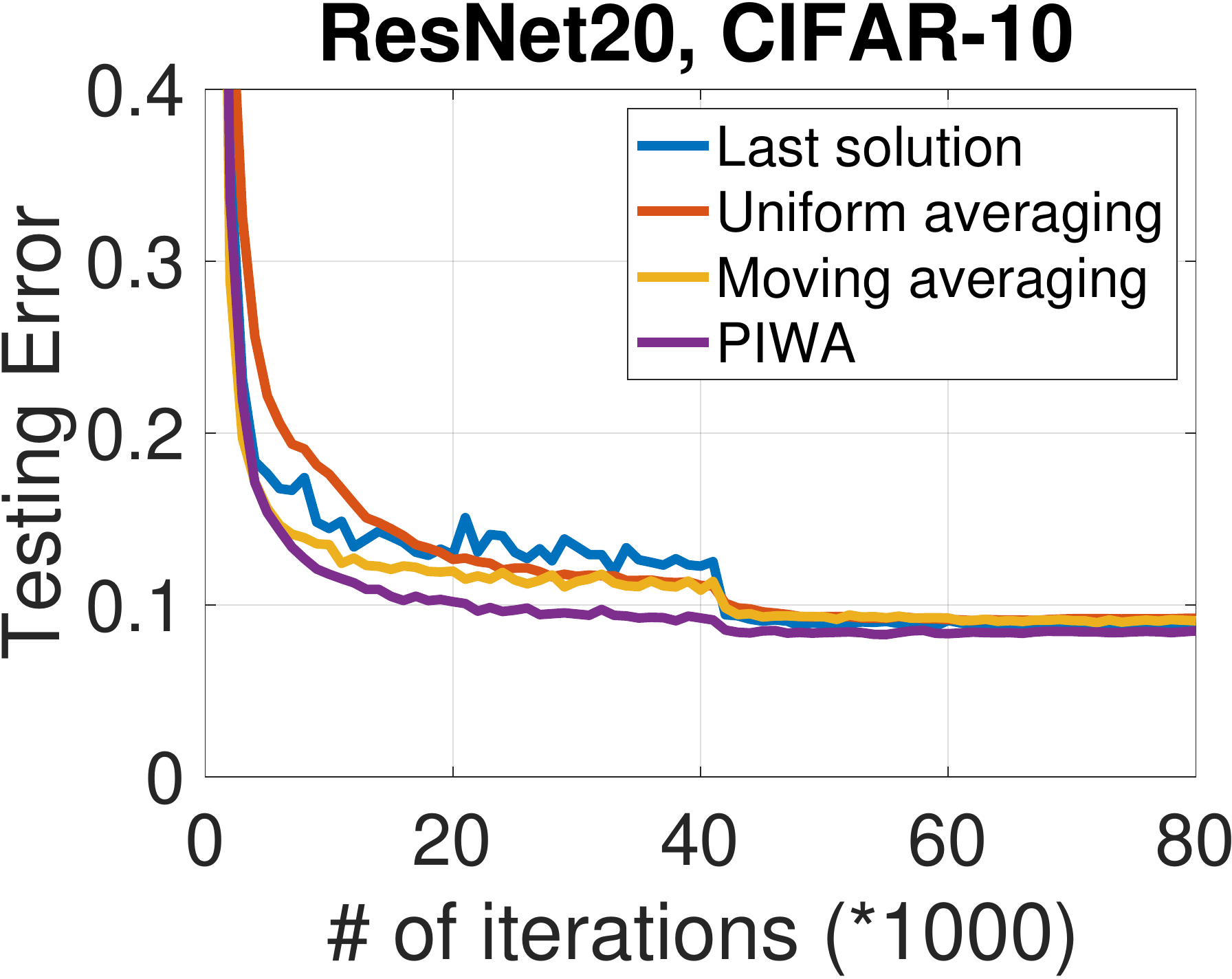}
\includegraphics[scale=0.2]{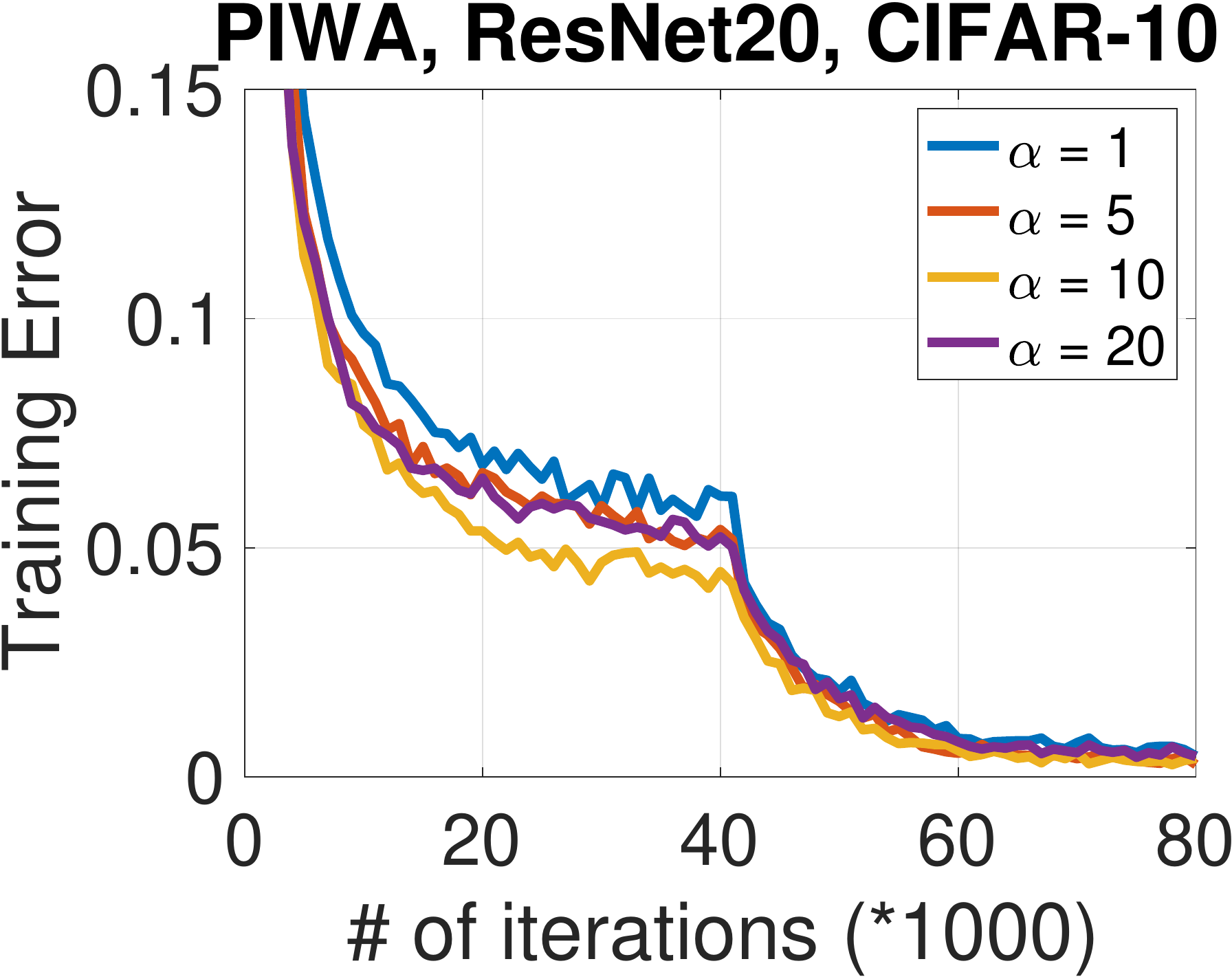}
\includegraphics[scale=0.2]{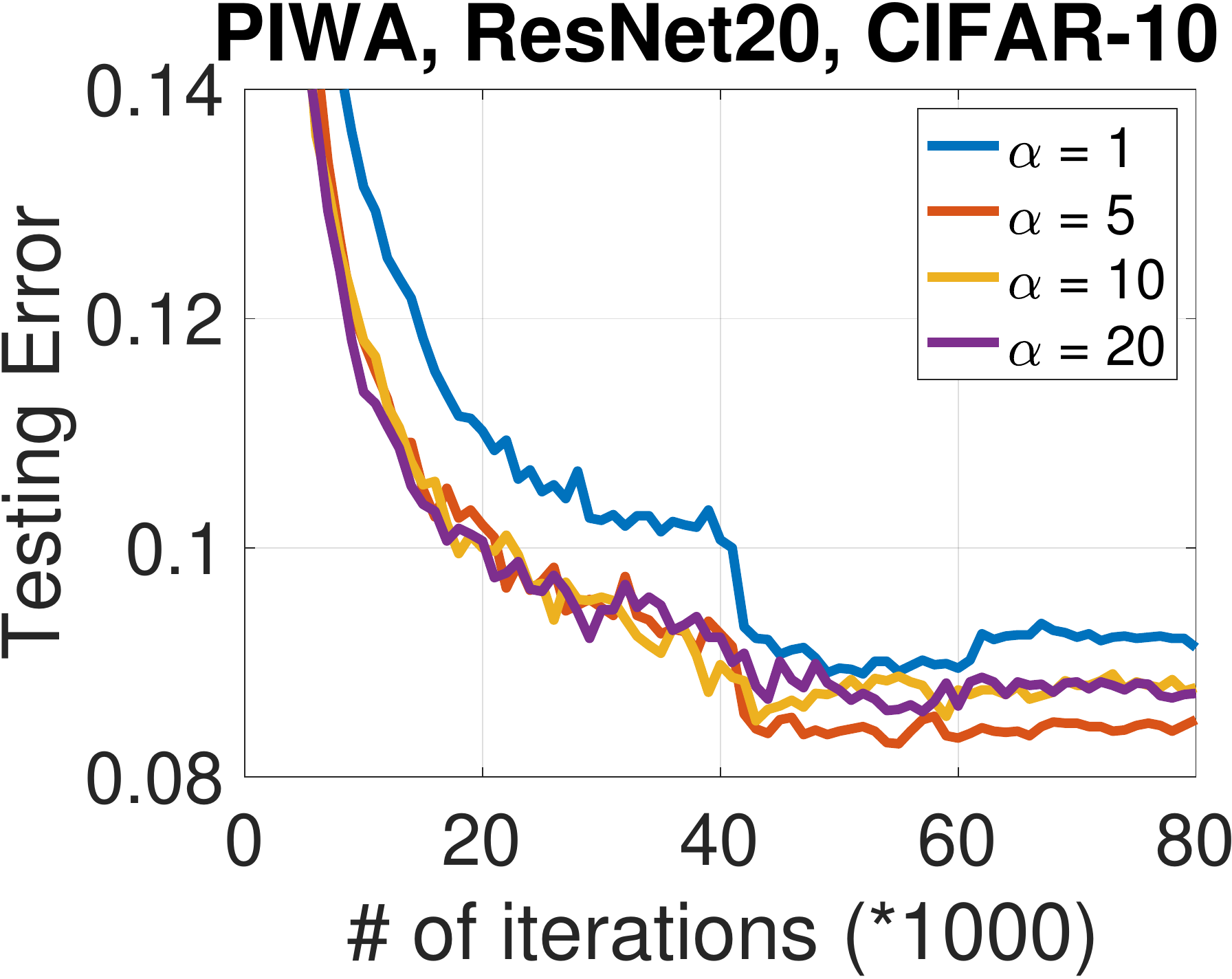}\\
\includegraphics[scale=0.2]{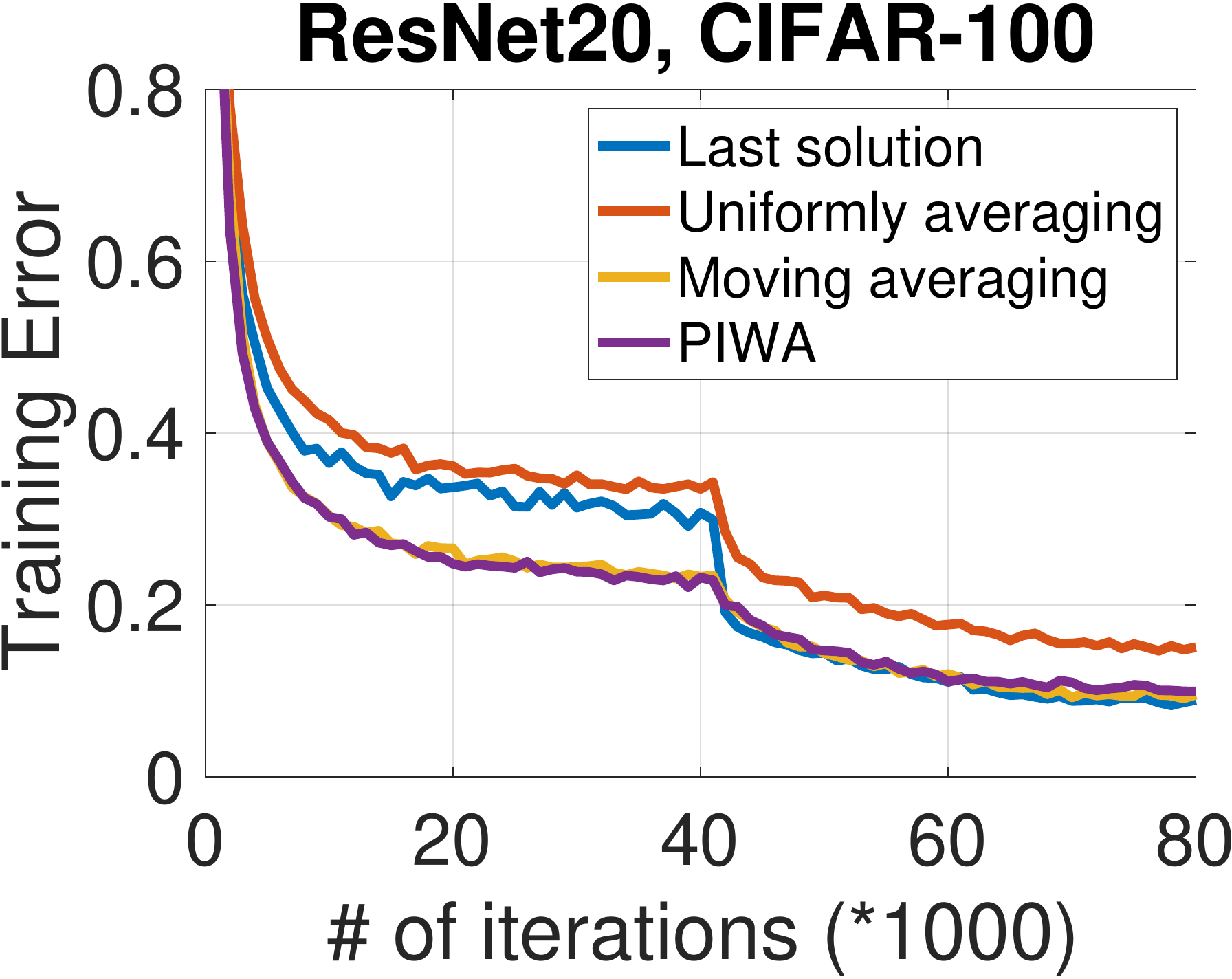}
\includegraphics[scale=0.2]{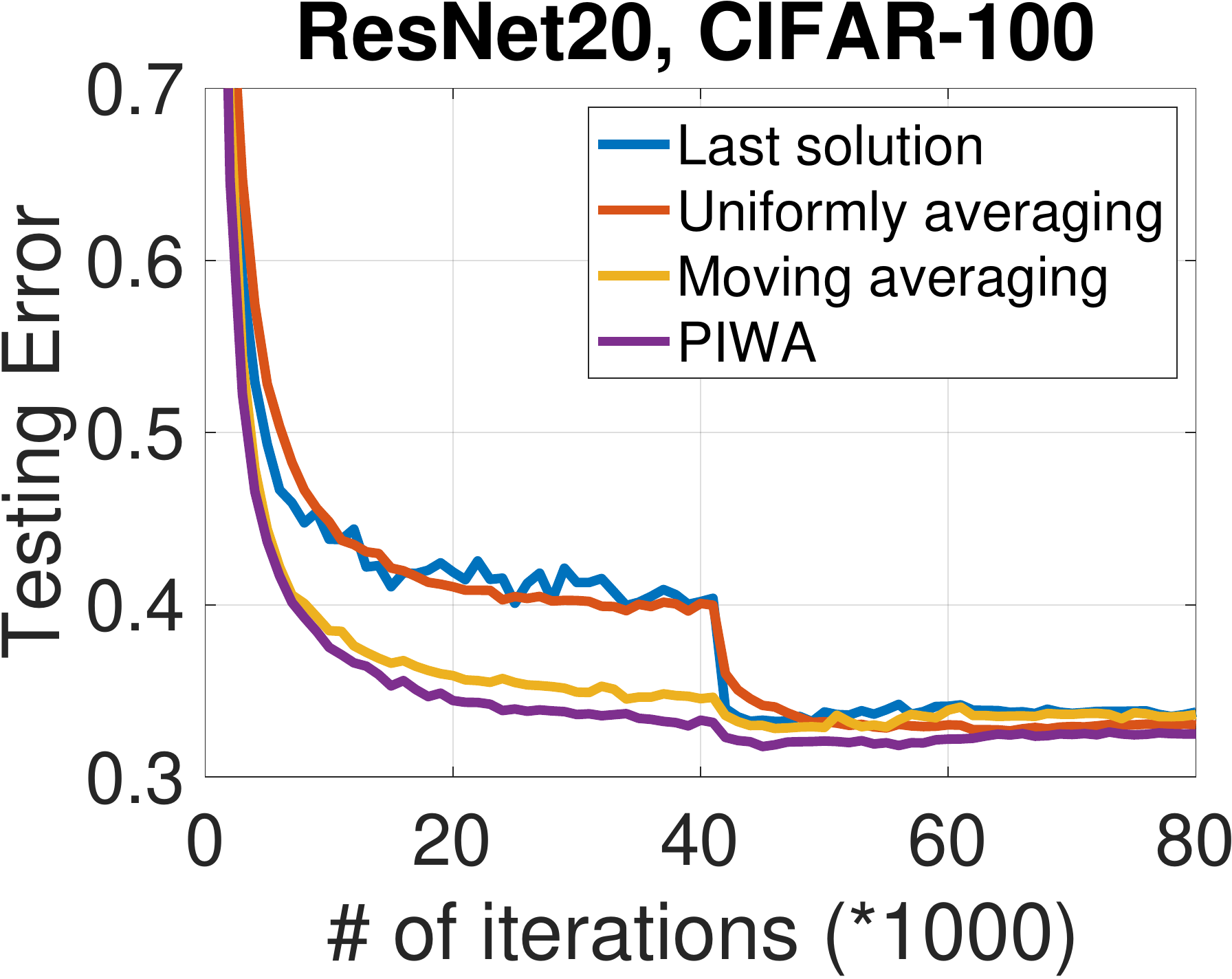}
\includegraphics[scale=0.2]{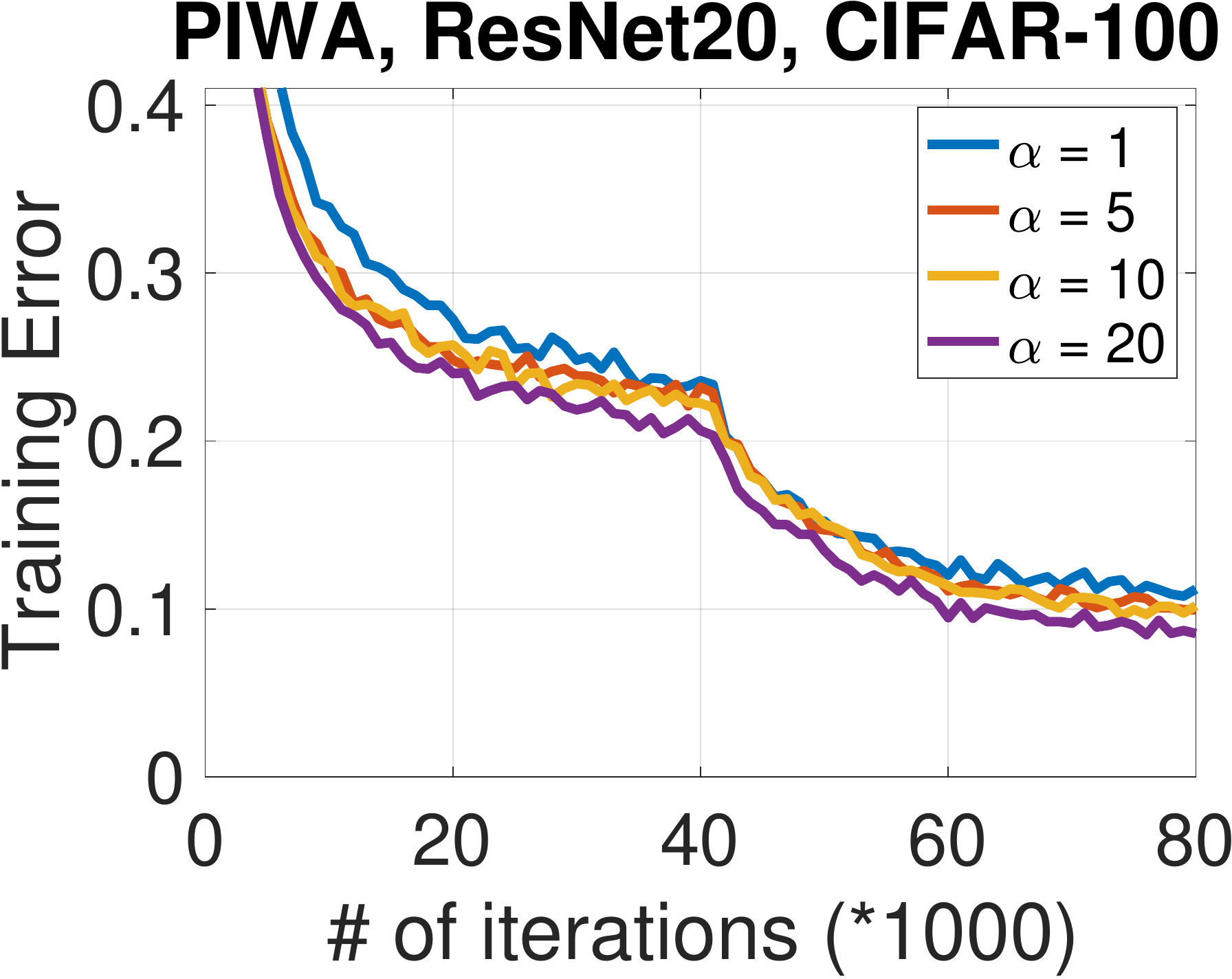}
\includegraphics[scale=0.2]{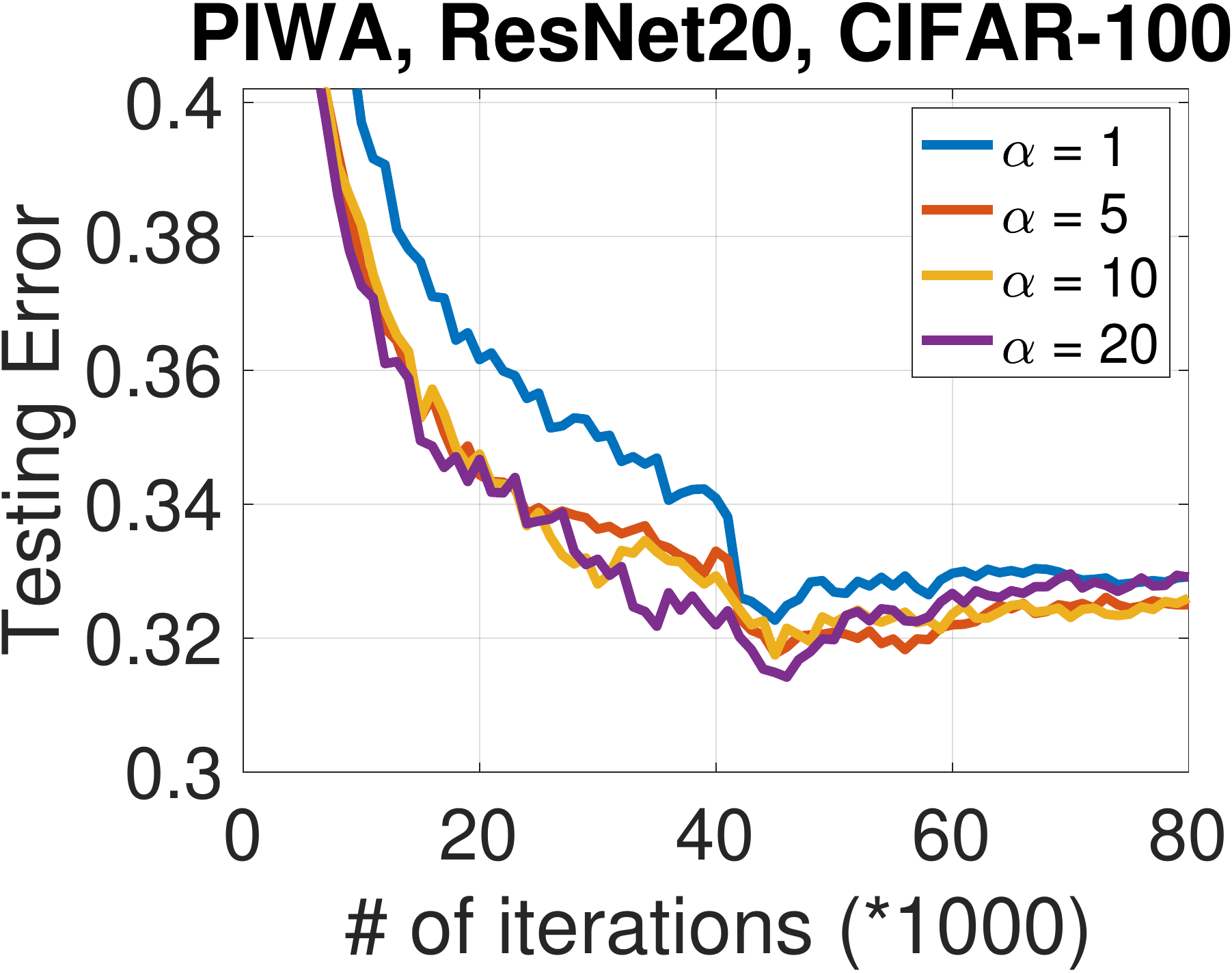}
\caption{Experiments of non-convex objective functions: ResNet20.}
\label{figure:nonconvex_results}
\end{figure*}

In this section, we demonstrate the effectiveness of PIWA.
First, we compare PIWA with different averaging baselines for convex, strongly convex and non-convex objective functions.
The baselines are the last solution, uniform averaging and moving averaging.
The goal is to show desirable generalization performance of PIWA.
Second, to reveal why PIWA achieves better generalization performance, we show that there is trade-off of $\alpha$ between optimization and generalization error by comparing different variants of PIWA.
We set various values of $\alpha$ to show how it affects optimization and generalization error for convex, strongly convex and non-convex objective functions.

{\bf Settings.}
For a dataset $\mathcal{S} = \{z_1, ..., z_n\}$, each $z_i = (a_i, b_i)$, where  $a_i\in\mathbb R^d$ is the feature vector and $b_i\in\{-1,1\}$ is the label. 
Hinge loss is used for the general convex case:
$
F(x) = \sum\limits_{i = 1}^{n} \max \{0, 1 - b_i a_i^T x\}.
$
For strongly convex case, we add  an $L_2$ norm regularization, i.e, $\lambda\|x\|^2$ into the above objective.
For non-convex case, we learn  a ResNet20 network with softmax loss \cite{he2016deep}.
We compare PIWA with the last solution, uniform averaging and exponential moving averaging.

{\bf Datasets.}
For convex and strongly convex cases, we perform experiments on covtype, splice, svmguide1 and skin-nonskin from the LIBSVM binary classification datasets \cite{chang2011libsvm}.
For covtype and skin-nonskin, as no testing datasets were provided, we randomly sampled 50,000 data points as testing data.
For other datasets used in convex/strongly convex experiments, we use the training/testing protocol provided by LIBSVM data \cite{chang2011libsvm}.
For non-convex case, we use CIFAR-10 and CIFAR-100 datasets \cite{krizhevsky2014cifar}.

{\bf Parameters.}
In the convex and strongly convex cases, the initial step size $\eta_1$ is tuned in $\{5, 1, 0.1, 0.01\}$.
In strongly convex case, the regularization weight $\lambda$ is tuned in $\{0.1, 0.01, 0.001\}$.
In convex and strongly convex cases, we sample one instance at each iteration.
In the non-convex case, the initial step size $\eta_1$ is tuned in $\{0.3, 0.5, 0.7, 0.9\}$, $\gamma$ is tuned in $\{1000, 2000, 3000\}$.
The number of iterations is set to be 40k for the first stage, and  20k for the second stage, which is the same as in \cite{he2016deep}.
The step size is decayed by a constant factor of 10 after each stage.
For exponential moving average, the decay coefficient is tuned in $\{ 0.99, 0.9, 0.8\}$ following~\cite{kingma2014adam,zhang2015deep}.
The batch size used in non-convex case is 128.
For experiments on all three cases, $\alpha$ is tuned in $\{1, 5, 10, 20\}$.

{\bf Results.}
The experimental results are shown in Figure \ref{figure:convex_results}, Figure \ref{figure:strongly_convex_results} and Figure \ref{figure:nonconvex_results}.
For convex case (Figure \ref{figure:convex_results}) and strongly convex case (Figure \ref{figure:strongly_convex_results}), we plot the curves of the values of objective functions and testing error.
Note that the testing error in the experiments refer to the rate of wrong classification.
For non-convex case (Figure \ref{figure:nonconvex_results}), we plot the curves of the training error and testing error.
For all three figures, the first two columns compare PIWA with the three baselines.
The last two columns compare variants of PIWA with different values of $\alpha$ to demonstrate the trade-off of $\alpha$.

From the first two columns of three figures, we can see that PIWA often achieves the best performance in testing error among all the averaging schemes,
even if PIWA could not always outperforms other averaging baselines in training.
In addition, PIWA tends to make the output solution stable, since the curves of PIWA rarely fluctuate dramatically.
In contrast, other baselines often return unstable solutions.
Particularly, the curves of the last solution method often encounter sharp fluctuation, specially in convex and strongly convex settings, 
where we sample one instance per iteration.
In contrast, we use a batch size of 128 in the non-convex setting, where the last solution method keeps more stable.

The last two columns of three tables show the trade-off of $\alpha$ between the optimization and generalization error.
Specifically, a larger $\alpha$ usually leads to faster training convergence, but it may make the solution more unstable.
On the other hand, a larger $\alpha$ often leads to larger testing error, even if the training performance is better.

\vspace*{-0.1in}
\section{Conclusion} 
In this paper, we have comprehensively analyzed SGD with PIWA in terms of both optimization error and generalization error for convex, strongly convex and non-convex problems.
We have shown in theory why PIWA with a proper $\alpha$ can improve the optimization error.
We have also shown that in PIWA, a larger $\alpha$ usually leads to a worse generalization error.
Thus, there is a trade-off caused by $\alpha$ between optimization error and generalization error.
Experiments on benchmark datasets have demonstrated this trade-off and effectiveness of PIWA compared with other averaging schemes.

\bibliography{reference}

\newpage
\begin{appendix}
\section{Proof of Lemma 2}
\textit{Proof.}
To show that $H_T(\alpha)$ is non-increasing in $\alpha$, it suffices to show that $\nabla H_T(\alpha) \leq 0$.
\begin{equation}
\begin{split}
&\nabla H_T (\alpha) = \frac{1}{\left (\sum\limits_{t = 1}^{T} t^{\alpha} \right )^2}
\left[
\left(\sum\limits_{t = 1}^{T} t^{\alpha} F_{\mathcal{S}}(x_t) \ln t \right) 
\left(\sum\limits_{t = 1}^{T} t^{\alpha}\right) 
- \left (\sum\limits_{t = 1}^{T}t^{\alpha} \ln t\right) \left(\sum\limits_{t = 1}^{T} t^{\alpha} F_{\mathcal{S}}(x_t)\right)
\right]
\\
&= \frac{1}{\left (\sum\limits_{t = 1}^{T} t^{\alpha} \right )^2}
\left[ \left(
\sum\limits_{i = 1}^{T}\sum\limits_{j = 1}^{T}i^{\alpha}F_{\mathcal{S}}(x_i)j^{\alpha}\ln i \right)
- \left(\sum\limits_{i = 1}^{T}\sum\limits_{j = 1}^{T}i^{\alpha} F_{\mathcal{S}}(x_i)j^{\alpha}\ln j \right)
\right]
\\
&= \frac{1}{\left (\sum\limits_{t = 1}^{T} t^{\alpha} \right )^2} {\sum\limits_{i = 1}^{T}\sum\limits_{j = 1}^{T}i^{\alpha} j^{\alpha}F_{\mathcal{S}}(x_i) (\ln i - \ln j)}\\
& = \frac{1}{2\left (\sum\limits_{t = 1}^{T} t^{\alpha} \right )^2} 
\left[\sum\limits_{i = 1}^{T}\sum\limits_{j = 1}^{T}i^{\alpha} j^{\alpha}F_{\mathcal{S}}(x_i) (\ln i - \ln j) + \sum\limits_{j = 1}^{T}\sum\limits_{i = 1}^{T}j^{\alpha} i^{\alpha}F_{\mathcal{S}}(x_j) (\ln j - \ln i)\right ]\\
& = \frac{1}{2\left (\sum\limits_{t = 1}^{T} t^{\alpha} \right )^2}\left[\sum\limits_{i = 1}^{T}\sum\limits_{j = 1}^{T}i^{\alpha} j^{\alpha}(F_{\mathcal{S}}(x_i) - F_{\mathcal{S}}(x_j))(\ln i - \ln j)\right]
\leq 0,
\end{split}
\end{equation}
where the inequality is from the assumption that $F_{\mathcal{S}}(x_T) \leq F_{\mathcal{S}}(x_{T-1}) \leq ... \leq F_{\mathcal{S}}(x_2) \leq F_{\mathcal{S}}(x_1)$.

\section{Proof of Theorem 1}
\textit{Proof.}
We have
\begin{equation}
\begin{split}
\frac{1}{\sum\limits_{t=1}^{T}t^{\alpha}} E\left[\sum_{t=1}^{T} t^{\alpha}(F_{\mathcal{S}}(x_t)-F_{\mathcal{S}}(x_*))\right]
&= \frac{1}{\sum\limits_{t=1}^{T}t^{\alpha}} \sum\limits_{t=1}^{T} t^{\alpha} 
E\left[F_{\mathcal{S}}(x_t) - F_{\mathcal{S}}(x_*)\right] \\
&= \frac{1}{\sum\limits_{t=1}^{T}t^{\alpha}} \sum\limits_{t=1}^{T} t^{\alpha} E[f(x_t; z_{i_t}) - f(x_*; z_{i_t})],
\end{split}
\end{equation}
where $i_t$ denotes the sampled data point in the $t$-th iteration.

According to standard analysis of SGD on convex functions, we have
\begin{equation}
\begin{split}
E[f(x_t; z_{i_t}) - f(x_*; z_{i_t})] \leq E[\nabla f(x_t; z_{i_t})^T (x_t - x_*)]
 \leq E\left [\frac{\|x_t - x_*\|^2}{2\eta_t} - \frac{\|x_{t+1} - x_*\|^2}{2\eta_t}+\frac{\eta_t G^2}{2} \right].\\
\end{split}
\label{gc}
\end{equation}

Multiplying  both sides by $w_t = t^{\alpha}$, and taking summation from $t=1$ to $T$, we have
\begin{equation}
\begin{split}
 & \sum\limits_{t = 1}^{T}w_t  E\left[f(x_t; z_{i_t}) - f(x_*; z_{i_t})\right]\\
&\leq \sum\limits_{t = 1}^{T}
   \left[\frac{w_t}{2\eta_t} E\left[\|x_t - x_*\|^2\right] 
   - \frac{w_{t - 1}}{2\eta_{t - 1}} E\left[\|x_t-x_*\|^2\right]+ 
   \frac{w_{t - 1}}{2\eta_{t - 1}}E\left[\|x_t - x_*\|^2\right] \right. \\
& \qquad \qquad
 \left. - \frac{w_t}{2\eta_t} E(\|x_{t+1} - x_*\|^2)
   + \frac{w_t\eta_t G^2}{2}\right]\\
& =  \sum\limits_{t = 1}^{T} 
     \left(\frac{w_t}{2\eta_t}  - \frac{w_{t - 1}}{2\eta_{t - 1}}\right)
     E\left[\|x_t - x_*\|^2\right] 
   + \frac{w_{0}}{2\eta_{0}}E\left[\|x_{1} - x_*\|^2\right]   - \frac{w_T}{2\eta_T}E\left[\|x_{T+1} - x_*\|^2\right]
   + \sum\limits_{t = 1}^{T}\frac{w_t \eta_t G^2}{2}\\
& \leq \left(\frac{w_T}{2\eta_T} - \frac{w_{0}}{2\eta_{0}} \right) D^2 + \sum\limits_{t = 1}^{T}      \frac{w_t \eta_t G^2}{2}  =
   \frac{T^{\alpha + 1/2}}{2\eta_1} D^2
   + \frac{\eta_1 G^2}{2} \sum\limits_{t = 1}^{T}{t}^{\alpha - 1/2},\\
\end{split}
\label{convexSum}
\end{equation}
where we let $w_{0} = 0$ and $\eta_{0} = +\infty$ since these variables just exist in the analysis and not used in the algorithm.

Then we can get the claimed bound by dividing both sides with $\sum\limits_{t=1}^{T} w_t$ and using the following standard calculus which is going to be used frequently in this paper:

\begin{equation}
\begin{split}
& \sum\limits_{s=1}^{S} s^{\alpha} \geq \int_{0}^{S} x^{\alpha} dx = \frac{1}{\alpha + 1} S^{\alpha + 1}, \forall \alpha > 0,~~~~~~~~~~~~~~~(5-1)\\
&\sum\limits_{s=1}^{S} s^{\alpha} \leq \int_{1}^{S+1} x^{\alpha} dx \leq \frac{(S+1)^{\alpha + 1}}{\alpha + 1}, \forall \alpha > 0,~~~~~~~~~~~~(5-2)\\
&\sum\limits_{s=1}^{S} s^{\alpha - 1} \leq S S^{\alpha - 1} = S^{\alpha}, \forall \alpha \geq 1,~~~~~~~~~~~~~~~~~~~~~~~~~~(5-3)\\
&\sum\limits_{s=1}^{S} s^{\alpha - 1} \leq \int_{0}^{S} x^{\alpha - 1} dx = \frac{S^{\alpha}}{\alpha}, \forall 0< \alpha < 1, ~~~~~~~~~~~~~~(5-4)\\
& \sum\limits_{s=1}^{S}s^{-1}\leq \ln S + 1.~~~~~~~~~~~~~~~~~~~~~~~~~~~~~~~~~~~~~~~~~~~~~~(5-5)
\end{split}
\label{calculus}
\end{equation}

Plug (5-1), (5-3) and (5-4) into (\ref{convexSum}), we get
\begin{equation}
\begin{split}
&\frac{1}{\sum\limits_{t=1}^{T}t^{\alpha}}  E\left[{\sum\limits_{t=1}^{T}t^{\alpha}(F_{\mathcal{S}}(x_t)} - F_{\mathcal{S}} (x_*)) \right] \leq \left\{
\begin{aligned}
&\frac{(\alpha + 1)D^2}{2\eta_1{T}^{1/2}} + \frac{(\alpha + 1)\eta_1 G^2}{(2\alpha+1){T}^{1/2}}, &0 \leq \alpha <1/2,\\
&\frac{(\alpha + 1)D^2}{2\eta_1 {T}^{1/2}} + \frac{(\alpha + 1)\eta_1 G^2}{{T}^{1/2}}, &\alpha \geq 1/2.
\end{aligned}
\right.
\end{split}
\end{equation}

\section{Proof of Theorem 2}

\textit{Proof.}
${x}_t$ and ${x}_t'$ are two sequences generated by SGD using two different data sets that differ only in one location, and they have the same initial point.
$\bar{x}_T$ and $\bar{x}_T'$ are the weighted average of these two sequences, respectively.
\begin{equation}
\begin{split}
\bar{x}_T = \frac{1}{\sum\limits_{t = 0}^{T} t^{\alpha}}\sum\limits_{t = 0}^{T} t^{\alpha}x_t, ~
\bar{x}_T' = \frac{1}{\sum\limits_{t = 0}^{T} t^{\alpha}} \sum\limits_{t = 0}^{T} t^{\alpha} x_t'.
\end{split}
\end{equation}

Applying Jensen's inequality, we have
\begin{equation}
\begin{split}
&E\left[\|\bar{x}_T - \bar{x}_T'\| \right] \leq \frac{1}{\sum\limits_{t = 1}^{T} w_t} 
   \sum\limits_{t = 1}^{T} {w_t}  E\left[\|x_t - x_t'\|\right].\\
\end{split}
\end{equation}

From Theorem 3.8 in (\cite{hardt2015train}), we have,
\begin{equation}
\begin{split}
E[\|x_t-x_t'\|]\leq \frac{2G}{n}\sum\limits_{i=1}^{t-1}\eta_i.
\end{split}
\end{equation}

Then,
\begin{equation}
\begin{split}
E\left[\|\bar{x}_T - \bar{x}_T'\| \right] 
 &\leq \frac{1}{\sum\limits_{t = 1}^{T} {t}^{\alpha}} 
   \sum\limits_{t = 1}^{T}\left [{t}^{\alpha}\frac{2G}{n}
   \sum\limits_{i = 1}^{t - 1} \eta_i \right] 
   \leq \frac{1}{\sum\limits_{t = 1}^{T} {t}^{\alpha}}
   \sum\limits_{t=1}^{T}\left[t^{\alpha} \frac{4G\eta_1}{n}(t-1)^{1/2}\right]\\
   &\leq \frac{1}{\sum\limits_{t = 1}^{T} {t}^{\alpha}}\sum\limits_{t=1}^{T}\left[\frac{4G\eta_1}{n}t^{1/2 + \alpha} \right] \leq \frac{4G\eta_1}{n} \frac{(\alpha+1)(T+1)^{\alpha+3/2}}{(\alpha+3/2)T^{\alpha + 1}},
\end{split}
\end{equation}
where the last inequality is from (5-1) and (5-2).

Then by the $G$-Lipschitz of $f(x; z)$, it follows that for any fixed $z$,
\begin{equation}
\begin{split}
& E[|f(\bar{x}_T; z) - f(\bar{x}_T'; z)|]
 \leq \frac{4\eta_1 G^2 (\alpha + 1) {(T + 1)}^{\alpha + 1.5} }{n (\alpha + 1.5){T}^{\alpha + 1}}.
\end{split}
\end{equation}
Since this bound holds for all $\mathcal{S}, \mathcal{S}'$ and $z$, we get the bound of the uniform stability $\epsilon_{stab}$.

\section{Proof of Theorem 3}
\textit{Proof.}
We have
\begin{equation}
\begin{split}
\frac{1}{\sum\limits_{t=1}^{T}t^{\alpha}} E\left[\sum_{t=1}^{T} t^{\alpha}(F_{\mathcal{S}}(x_t)-F_{\mathcal{S}}(x_*))\right]
&= \frac{1}{\sum\limits_{t=1}^{T}t^{\alpha}} \sum\limits_{t=1}^{T} t^{\alpha} 
E\left[F_{\mathcal{S}}(x_t) - F_{\mathcal{S}}(x_*)\right] \\
&= \frac{1}{\sum\limits_{t=1}^{T}t^{\alpha}} \sum\limits_{t=1}^{T} t^{\alpha} E[f(x_t; z_{i_t}) - f(x_*; z_{i_t})].
\end{split}
\label{eq:strong0}
\end{equation}

By standard analysis of SGD on strongly convex function, we have
\begin{equation}
\begin{split}
 &E\left[f(x_{t}; z_{i_t}) - f(x_*; z_{i_t})\right] \leq  E\left[\frac{\|x_{t} - x_*\|^2}{2\eta_{t}}\right] - E\left[\frac{\|x_{t+1} - x_*\|^2}{2\eta_{t}}\right] 
    - \frac{\lambda}{2}E\left[\|x_{t} - x_*\|^2 \right]
     + \frac{\eta_{t} G^2}{2}.\\
\end{split}
\end{equation}

Multiplying $w_t=t^{\alpha}$ to both sides and taking summation from $1$ to $T$, we have
\begin{equation}
\begin{split}
& \sum\limits_{t = 1}^{T}w_t  E\left[f(x_t; z_{i_t}) - f(x_*; z_{i_t})\right]\\
& \leq \sum\limits_{t = 1}^{T} 
   \left [
   w_{t} E\left[ \frac{\|x_{t} - x_*\|^2}{2\eta_{t}}\right] -w_{t - 1}  E \left[\frac{\|x_{t} - x_*\|^2}{2\eta_{t - 1}} \right ]
   \right.   + w_{t - 1} E \left[ \frac{\|x_{t} - x_*\|^2}{2\eta_{t - 1}} \right]
   -  w_{t}  E \left[ \frac{\|x_{t+1} - x_*\|^2}{2\eta_{t}} \right]\\
& ~~~~~
   \left.   
    - \frac{\lambda w_t}{2} E\left[\|x_{t} - x_*\|^2 \right]
    + w_{t} \frac{\eta_{t} G^2}{2}
    \right] \\
& = \sum\limits_{t = 1}^{T} 
    \left(\frac{w_{t - 1}}{2\eta_{t - 1}}E(\|x_{t} - x_*\|^2)
    -  \frac{w_{t}}{2\eta_{t}}E[\|x_{t+1} - x_*\|^2]
    \right)+\sum\limits_{t = 1}^{T} 
    \left (\frac{w_{t} }{2\eta_{t}} 
   - \frac{w_{t - 1} }{2\eta_{t - 1}} 
   - \frac{\lambda w_t}{2} \right)E[\|x_{t} - x_*\|^2]
   \\&~~~~~+ \sum\limits_{t = 1}^{T} w_{t} \frac{\eta_{t} G^2}{2} \\
 &=
   \frac{w_{0}}{\eta_{0}} E[\|x_1 - x_*\|^2]
   - \frac{w_{T}}{2\eta_{T}}E[\|x_{T+1} - x_*\|^2]+ \sum\limits_{t = 1}^{T} {\left(\frac{w_{t}}{2\eta_{t}} 
   -  \frac{w_{t - 1}}{2\eta_{t - 1}} 
   - \frac{\lambda w_t}{2}\right)
   E\left(\|x_{t} - x_*\|^2 \right)}\\
   &~~~~~+\sum\limits_{t = 1}^{T} \frac{w_{t}\eta_{t} G^2}{2} \\
&\leq \sum\limits_{t = 1}^{T} \frac{w_{t}\eta_{t} G^2}{2}, 
\end{split}
\label{eq:strong}
\end{equation}
\noindent
where the last inequality holds because we define $w_{0} = 0$ and $\eta_{0}  =+\infty$, and $\frac{w_{t}}{2\eta_{t}} -  \frac{w_{t - 1}}{2\eta_{t - 1}} - w_t\frac{\lambda}{2} \leq  0$.
In the following, we are going to prove that $\frac{w_{t}}{2\eta_{t}} -  \frac{w_{t - 1}}{2\eta_{t - 1}} - w_t\frac{\lambda}{2} \leq  0$.
It holds obviously if $t = 1$.
If $t \geq 2$, 
\begin{equation*}
\begin{split}
&\frac{w_{t}}{2\eta_{t}} -  \frac{w_{t - 1}}{2\eta_{t - 1}} = \frac{1}{4} \left ( \frac{\lambda t^{\alpha + 1}}{\alpha + 1} - \frac{\lambda (t-1)^{\alpha + 1}}{\alpha + 1} \right ) = \frac{1}{4}\lambda \int_{t}^{t+1} (x - 1)^{\alpha} dx \leq \frac{1}{4}\lambda t^{\alpha} = \frac{\lambda w_t}{4}.
\end{split}
\end{equation*}

So ${\left (\frac{w_{t}}{2\eta_{t}} -  \frac{w_{t - 1}}{2\eta_{t - 1}} - w_t\frac{\lambda}{2}\right )} \leq -\frac{\lambda w_t}{4} \leq 0$ for any
$t$.

We divide both sides of (\ref{eq:strong}) by $\sum\limits_{t=1}^{T} w_t$ and plug it back into (\ref{eq:strong0}):
\begin{equation}
\begin{split}
&\frac{1}{\sum\limits_{t=1}^{T}t^{\alpha}}E\left[{\sum\limits_{t=1}^{T}t^{\alpha}(F_{\mathcal{S}}(x_t)} - F_{\mathcal{S}} (x_*)) \right] \leq \frac{(\alpha+1)G^2\sum\limits_{t=1}^{T}t^{\alpha-1}}{\lambda \sum\limits_{t=1}^{T} t^{\alpha}}.
\end{split}
\label{eq:strong1}
\end{equation}

Then, we consider three cases of $\alpha$ as follows:

(i) $\alpha=0$: we plug (5-5) into (\ref{eq:strong1}) and get
\begin{equation}
\frac{1}{\sum\limits_{t=1}^{T}t^{\alpha}}E\left[{\sum\limits_{t=1}^{T}t^{\alpha}(F_{\mathcal{S}}(x_t)} - F_{\mathcal{S}} (x_*)) \right] \leq \frac{G^2}{\lambda T} (1 + \log (T)).
\end{equation}

(ii) $0<\alpha<1$: we plug (5-1) and (5-4) into (\ref{eq:strong1}) and get
\begin{equation}
\frac{1}{\sum\limits_{t=1}^{T}t^{\alpha}}E\left[{\sum\limits_{t=1}^{T}t^{\alpha}(F_{\mathcal{S}}(x_t)} - F_{\mathcal{S}} (x_*)) \right] \leq \frac{{(\alpha + 1)}^2 G^2}{\alpha \lambda T}.
\end{equation}.

(iii) $\alpha \geq 1$: we plug (5-1) and (5-2) into (\ref{eq:strong1}) and get
\begin{equation}
\frac{1}{\sum\limits_{t=1}^{T}t^{\alpha}}E\left[{\sum\limits_{t=1}^{T}t^{\alpha}(F_{\mathcal{S}}(x_t)} - F_{\mathcal{S}} (x_*)) \right] \leq \frac{{(\alpha + 1)}^2G^2  {(T+1)}^{\alpha}}{\alpha\lambda {T}^{\alpha+1}}.
\end{equation}


\section{Proof of Theorem 4}
\textit{Proof.} 
According to Theorem 3.11 in \cite{hardt2015train}, we have
\begin{equation}
\begin{split}
E[f(\bar{x}_T; \xi) - f(\bar{x}_T'; \xi)] \leq \frac{t_0}{n} \sup\limits_{x, \xi} f(x; \xi) + G E[\delta_T | \delta_{t_0} = 0],
\end{split}
\end{equation}
where $t_0$ is any number in $\{0, 1, ..., n\}$.

Let $\delta_t = \|x_t - x_t'\|$ and $t_0 = \max\{ \frac{2(\alpha+1)G}{\lambda}, 1\}$. 
Then as Theorem 3.10 in (\cite{hardt2015train}),  we have 
\begin{equation}
\begin{split}
    E[f(\bar{x}_T; \xi) - f(\bar{x}_T'; \xi)] \leq \frac{t_0}{n} + G E\left[ \left \|x_T - x_T'\right \| \bigg | \delta_{t_0} = 0 \right]. \\
\end{split}
\label{eq:thm4hardt}
\end{equation}

From Theorem 3.9 in (\cite{hardt2015train}), for $t>t_0\geq \frac{2(\alpha+1)G}{\lambda}$, we have
\begin{equation}
\begin{split}
E\left[\delta_{t}|\delta_{t_0}=0\right]&\leq (1-\frac{1}{n})\left(1 - \frac{2(\alpha+1)}{\lambda t} \lambda \right) E[\delta_t|\delta_{t_0}=0]+\frac{1}{n}\left(1 - \frac{2(\alpha+1)}{\lambda t} \lambda \right) E[\delta_t|\delta_{t_0}=0] + \frac{4(\alpha+1)G}{n\lambda t}\\
&=\left(1 - \frac{2(\alpha+1)}{t}\right)E[\delta_t|\delta_{t_0}=0] + \frac{4(\alpha+1)G}{n\lambda t}\\
&\leq (1 - \frac{1}{t}) E[\delta_t|\delta_{t_0}=0] + \frac{4(\alpha+1)G}{n\lambda t}\\
\end{split}
\end{equation}
Expanding the recursion, we have,
\begin{equation}
E[\delta_T] \leq \sum\limits_{t=t_0}^{T-1}\left[\prod\limits_{s=t+1}^{T-1} (1-  \frac{1}{s})\right] \frac{4G(\alpha+1)}{\lambda n t} \leq \frac{T-t_0}{T-1} \frac{4G(\alpha+1)}{\lambda n}.
\end{equation}


Then we have
\begin{equation}
\begin{split}
&E\left[\|\bar{x} - \bar{x}_T'\| | \delta_{t_0} = 0 \right] \leq  \frac{1}{\sum\limits_{t = 1}^{T} {t}^{\alpha}} \sum\limits_{t = 1}^{T} {t}^{\alpha} E\left[ \left \|x_t - x_t'\right \| \bigg | \delta_{t_0} = 0 \right]\\
&\leq \frac{1}{\sum\limits_{t = 1}^{T} {t}^{\alpha}} \sum\limits_{t = t_0 + 1}^{T} {t}^{\alpha} \frac{t - t_0}{t - 1} \frac{4G(\alpha+1)}{\lambda n}
\leq \frac{1}{\sum\limits_{t = 1}^{T} {t}^{\alpha}} 
   \frac{4G(\alpha+1)}{\lambda n}
   \left(\sum\limits_{t = 1}^{T} {t}^{\alpha}  
   \right) = \frac{4G(\alpha+1)}{\lambda n}.
\end{split}
\end{equation}

Then we plug the above inquality into (\ref{eq:thm4hardt}) and get 
\begin{equation}
\begin{split}
E[f(\bar{x}_T; \xi) - f(\bar{x}_T'; \xi)] \leq \frac{t_0}{n}  + \frac{4G^2(\alpha+1)}{\lambda n}.
\end{split}
\end{equation}

\section{Proof of Lemma 5}
\textit{Proof.} 
We need Azuma's inequality for the bound.

\textbf{Azuma's inequality:}
Let $X_1, ..., X_T$ be a martingale difference sequence.
Suppose that $|X_t| \leq b$.
Then, for $\delta > 0$, we have
\begin{equation}
\begin{split}
Pr \left[\sum\limits_{t = 1}^{T} X_t \geq b \sqrt{2T\ln (1/\delta)} \right] \leq \delta.
\end{split}
\end{equation}

Using standard techniques as in the convex case (but note here step size is a constant within a stage), we get
\begin{equation}
\begin{split}
&\frac{1}{\sum\limits_{t=0}^{T}w_t}\sum\limits_{t = 0}^{T} w_t \nabla f_k(x_t; z_{i_t})^T (x_t - x_*)
\leq \frac{2(\alpha + 1)D^2}{\eta (T+1)} + \frac{\eta \hat{G}^2}{2},
\end{split}
\label{lemma6_1}
\end{equation}
where $\hat{G}$ is the Lipchitz constant of $F_k(x)$, and thus can be set as $2G^2+2\lambda^{-2}D^2$.

Let $F_k = \sum\limits_{i=1}^{n} f_k(x; z)$.
Since $E[\nabla f_k(x_t; z_{i_t})^T (x_t - x_*)] = \nabla F_k(x_t)^T (x_t - x_*)$, the following defines as a martingale difference sequence:
\begin{equation}
\begin{split}
&X_t = t^{\alpha} \left(\nabla F_k(x_t)^T(x_t - x_*)-\nabla f_k(x_t;z_{i_t})^T(x_t- x_*)\right)\\
\end{split}
\end{equation}
for any $t$.

We can bound $|X_t|$ for any $t$ as follows:
\begin{equation}
\begin{split}
|X_t| \leq T^{\alpha} (\|\nabla F_k(x_t)^T(x_t - x_*)\| + \|\nabla f_k(x_t;z_{i_t})^T(x_t- x_*)\|) 
\leq 4{T}^{\alpha}\hat{G}D_k,
\end{split}
\end{equation}
which uses the $\hat{G}$-Lipchitz of $F_k(x)$ and the assumption this stage has a constraint ball and triangle inequality, i.e., $\|x_t-x_*\| \leq \|x_t-x_{k-1}\| + \|x_*-x_{k-1}\| \leq 2D_k$.

By Azuma's inequality, with probability at least $1 - \delta$, the following holds:
\begin{equation}
\begin{split}
&\frac{1}{\sum\limits_{t = 1}^{T} w_t} \left(\sum\limits_{t = 1}^{T} t^{\alpha}\nabla F_k(x_t)^T(x_t - x_*) \right. \left. - \sum\limits_{t = 1}^{T} t^{\alpha} \nabla f_k(x_t; z_{i_t})^T(x_t - x_*) \right)\\
&\leq \frac{1}{\sum\limits_{t = 1}^{T} w_t}2\hat{G}D T^{\alpha}\sqrt{2T\ln (1/\delta)}
\leq \frac{4(\alpha + 1) \hat{G}D_k \sqrt{2 \ln (1/\delta)}}{\sqrt{T}}
\end{split}
\label{lemma6_2}
\end{equation}

By Jensen's inequality and the convexity of $F_{\mathcal{S}}$, we have
\begin{equation}
\begin{split}
F_k(x_t) - F_k(x_*)\leq \sum\limits_{t = 1}^{T} t^{\alpha} F_k (x_t)^T(x_t-x_*).
\end{split}
\label{lemma6_3}
\end{equation}

Add (\ref{lemma6_1}) and (\ref{lemma6_2}) and plug it into (\ref{lemma6_3}), we get
\begin{equation}
\begin{split}
&F_{k}(x_k) - F_{k}(x_*)\leq \frac{1}{\sum\limits_{t=1}^{T_k}t^{\alpha}}{\sum\limits_{t=1}^{T_k}t^{\alpha}(F_{k}(x^k_t)} - F_{k} (x_*))\\
&\leq\frac{2(\alpha + 1)\epsilon_{k-1}}{\eta_k \mu T} + \frac{\eta_k \hat{G}^2}{2}  + \frac{4(\alpha + 1) \hat{G}D_k \sqrt{2 \ln (1/\delta)}}{\sqrt{T}},
\end{split}
\end{equation}
with a probability at least $1 - \delta$.

\section{Proof of Theorem 5}
\textit{Proof.} We will prove by induction that $E[F_{\mathcal{S}}(x_k) - F_{\mathcal{S}}(x_*)] \leq \epsilon_k$, where $\epsilon_k = \epsilon_0/2^k$.
This is true for $k = 0$ by assumption.
Suppose this true for $(k-1)$-th stage, which means,
$F_{\mathcal{S}}(x_{k-1}) - F_{\mathcal{S}}(x_*) \leq \epsilon_{k-1} = \epsilon_0/2^{k-1}$.
Applying Lemma 4, we see that $\|x_{k-1}-x_*\|$ is bounded by $\frac{\epsilon_{k-1}}{\mu}$.

By applying Lemma 5 to the $k$-th stage and plug in $F_k(x)=F_{\mathcal{S}(x)}+\frac{1}{2\gamma}\|x-x_{k-1}\|^2$, with probability $1-\delta$,
\begin{equation}
\begin{split}
&E[F_{\mathcal{S}}(x_k) - F_{\mathcal{S}}(x_*)]
\leq \frac{\|x_{k-1} - x\|^2}{2\gamma} + 
\frac{2(\alpha + 1)\epsilon_{k-1}}{\eta_k \mu T_k} + \frac{\eta_k \hat{G}^2}{2} + \frac{2(\alpha + 1) \hat{G}D_k \sqrt{2 \ln (1/\delta)}}{\sqrt{T}}\\
&\leq  \frac{\epsilon_{k-1}}{2\gamma\mu} + 
\frac{2(\alpha + 1)\epsilon_{k-1}}{\eta_k \mu T_k} + \frac{\eta_k \hat{G}^2}{2}
+ \frac{2(\alpha + 1) \hat{G}D_k \sqrt{2 \ln (1/\delta)}}{\sqrt{T}}.
\\
\end{split}
\end{equation}

By setting $\eta_k = \frac{c \epsilon_k}{2\hat{G}^2}$,
$ T_k = \frac{d}{\mu\epsilon_k}$ and $\gamma = 4/\mu$, 
we get
\begin{equation}
\begin{split}
E[F_{\mathcal{S}}(x_k) - F_{\mathcal{S}}(x_*)] \leq \frac{\epsilon_{k-1}}{2} = \epsilon_k.
\end{split}
\end{equation}

By induction, after $K = \lceil{\log (\epsilon_0/\epsilon)} \rceil$ stages, we have 
\begin{equation}
\begin{split}
E[F_{\mathcal{S}}(x_K)-F_{\mathcal{S}}(x_*)] \leq \epsilon.
\end{split}
\end{equation}

The total iteration complexity is $\sum\limits_{k = 1}^{K} T_k = O((\alpha + 1) / (\mu\epsilon))$.

\section{Proof of Theorem 6}
\textit{Proof. }
We first need to establish the bound of stability for a single stage.
Let $\delta_t = \|x_t - x_t'\|$.
We need to consider two scenarios.
The first scenario is $f=f'$, then
\begin{equation}
\begin{split}
\delta_{t+1}&=\left\|\left(x_t- \eta \left(g(x_t) - \frac{1}{\gamma} (x_t - x_{1})\right)\right) - \left(x_t'- \eta \left(g(x_t') - \frac{1}{\gamma} (x_t' - x_{1}')\right)\right)\right\| \\
&\leq \frac{\eta}{\gamma}\|x_1-x_1'\| + \frac{\gamma-\eta}{\gamma}\|x_t - x_t'\| + \eta \|g(x_t) - g(x_t')\|\\
&\leq \frac{\eta}{\gamma}\|x_1 - x_1'\| + \left(\frac{\gamma-\eta}{\gamma} + \eta L\right)\|x_t - x_t'\|.
\end{split}
\end{equation}

The second scenario is $f\neq f'$, then
\begin{equation}
\begin{split}
\delta_{t+1} &= \left\|\left(x_t- \eta \left(g(x_t) - \frac{1}{\gamma} (x_t - x_{1})\right)\right) - \left(x_t'- \eta \left(g(x_t') - \frac{1}{\gamma} (x_t' - x_{1}')\right)\right) \right\|\\
& = \frac{\gamma-\eta}{\gamma}\left\|x_t - x_t'\right\| + \frac{\eta}{\gamma}\|x_1-x_1'\| + \eta \|g(x_t) - g'(x_t)\| \\
& \frac{\gamma - \eta}{\gamma} \delta_t + \frac{\eta}{\gamma} \delta_1 + 2\eta G.
\end{split}
\end{equation}

Then,
\begin{equation}
\begin{split}
\delta_{t+1} \leq \left\{
\begin{aligned}
\frac{\eta}{\gamma} \delta_1 +\left( \frac{\gamma - \eta}{\gamma} + \eta L \right) \delta_t, ~ f_t = f_t' \\
\frac{\eta}{\gamma} \delta_1+ \frac{\gamma - \eta}{\gamma} \delta_t+ 2\eta G, ~ otherwise.
\end{aligned}
\right.
\end{split}
\end{equation}

We will condition on $x_{K-1}=x_{K-1}'$, i.e., the different sample will only be used in the last stage.

If $x_{K-1}=x_{K-1}'$, for the $K$-th stage,  we have (Theorem 3.10 in (\cite{hardt2015train})):
\begin{equation}
\begin{split}
    E[f(\bar{x}_T; z) - f(\bar{x}_T'; z)] \leq \frac{t_0}{n} + G E\left[ \left \|\bar{x}_T - \bar{x}_T'\right \| \bigg | \delta_{t_0} = 0 \right], \\
\end{split}
\end{equation}
for any $t_0\in \{0,1,...,n\}$.

Following techniques in Theorem 3.12 in (\cite{hardt2015train}), since $\eta_K \leq c/(\mu T) \leq c/t$, we have
\begin{equation}
\begin{split}
E[\|x_t - x_t'\| | \delta_{t_0}=0] \leq \frac{2G}{L(n-1)}\left(\frac{t}{t_0}\right)^{Lc/\mu}.
\end{split}
\label{eq:finalsingle}
\end{equation}

Then using Jensen's inequality
\begin{equation}
\begin{split}
&E[\|\bar{x}_T-\bar{x}_T'\||\delta_{t_0}=0]\leq
\frac{1}{\sum\limits_{t=1}^{T}w_t} \sum\limits_{t=1}^{t}w_t E[\|x_t-x_t'\|| \delta_{t_0}=0]\\
&\leq \frac{2G(\alpha+1)}{L(n-1)T^{\alpha}}\sum\limits_{t=1}^{T}t^{\alpha}\left(\frac{t}{t_0}\right)^{Lc}
\leq \frac{2G (\alpha+1)T^{\alpha+Lc+1}}{L(n-1)T^{\alpha+1}t_0^{L c}} = \frac{2G (\alpha+1)T^{Lc}}{L(n-1)t_0^{Lc}},
\end{split}
\end{equation}
the second inequality is due to (\ref{eq:finalsingle}).

Then we get,
\begin{equation}
E[\|f(\bar{x}_T; z)-f(\bar{x}_T'; z)\|] \leq \frac{t_0}{n}+\frac{2G^2(\alpha+1)T^{Lc}}{L(n-1)t_0^{Lc}}.
\label{eq:41}
\end{equation}
The right hand side is approximately minimized when
\begin{equation}
t_0=(2(\alpha+1)cL^2)^{\frac{1}{Lc+1}} T^{\frac{Lc}{1+Lc}}.
\label{t_0}
\end{equation}
Pluging (\ref{t_0}) into (\ref{eq:41}) and recalling we have conditioned on $x_{K-1} = x_{K-1}'$, we get
\begin{equation}
E[|f(x_K; z)-f(x_K'; z)|]\leq \frac{S_{K-1}}{n} + \frac{1+\frac{1}{Lc}}{n-1}(2(\alpha+1)cL^2)^{\frac{1}{1+Lc}}T^{\frac{Lc}{1+Lc}}.
\end{equation}

\end{appendix}
\end{document}